# Learning to Answer Multilingual and Code-Mixed Questions

*Submitted in partial fulfillment of the requirements
of the degree of*

## Doctor of Philosophy

*by*

### Deepak Kumar Gupta
(Roll No: 1621CS08)

*Under the supervision of*

## Dr. Asif Ekbal & Prof. Pushpak Bhattacharyya

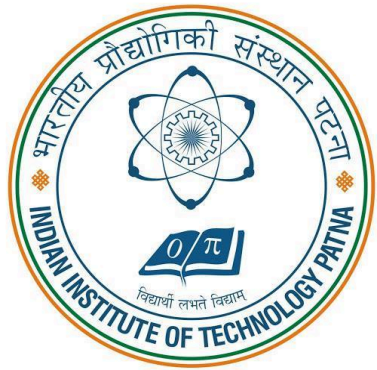

**DEPARTMENT OF COMPUTER SCIENCE & ENGINEERING
INDIAN INSTITUTE OF TECHNOLOGY PATNA
PATNA - 801 103, INDIA
November 2020**

Dedicated to the memory of my dearest grandparents

# Abstract

Question-answering (QA) that comes naturally to humans is a critical component in seamless human-computer interaction. It has emerged as one of the most convenient and natural methods to interact with the web and is especially desirable in voice-controlled environments. Despite being one of the oldest research areas, the current QA system faces the critical challenge of handling multilingual queries. To build an Artificial Intelligent (AI) agent that can serve multilingual end users, a QA system is required to be language versatile and tailored to suit the multilingual environment.

Recent advances in QA models have enabled surpassing human performance primarily due to the availability of a sizable amount of high-quality datasets. However, the majority of such annotated datasets are expensive to create and are only confined to the English language, making it challenging to acknowledge progress in foreign languages. Therefore, to measure a similar improvement in the multilingual QA system, it is necessary to invest in high-quality multilingual evaluation benchmarks. In this dissertation, we focus on advancing QA techniques for handling end-user queries in multilingual environments. To achieve this goal, we create multilingual benchmark setups, and develop computational models using the concept of deep neural network techniques to extract the answer for multilingual questions from the text or image data sources.

This dissertation consists of two parts. In the first part, we explore multilingualism, and a new dimension of multilingualism referred to as code-mixing. Towards this, we create the high-quality benchmark dataset required for multilingual question answering (MQA) systems in English, Hindi, and code-mixed English-Hindi languages. Additionally, we develop a unified deep neural network framework to retrieve the multilingual answers by exploring the attention-based recurrent neural network to generate the effective representation of multilingual questions and documents. We also extend the study of multilingualism to visual question answering (VQA) by exploring and analyzing the weakness of the existing VQA systems. We create a VQA dataset supporting the English, Hindi, and English-Hindi code-mixed languages. We propose a unified architecture to better encode the multilingual and code-mixed questions by introducing a hierarchy of shared layers. We control the behavior of these shared layers by an attention-based soft layer sharing mechanism. Our proposed approach can answer the language variant end-users' query in VQA.

In the second part of this dissertation, we investigate how we can further enhance the performance of the current QA systems by solving multiple related QA subtasks. To attain this, first, we focus on advancing the *semantic question matching* by introducing and augmenting the taxonomy knowledge into the deep neural network framework. Second, we propose the technique to solve the task of *multi-hop question generation* by exploiting multiple documents. We employ multitask learning with the auxiliary task of *answer-aware* supporting fact prediction to guide the question generator. The generated questions can be further used as additional data to improve the QA pipeline. Finally, to tackle the issue of data scarcity in code-mixing, we propose an approach for automatic *code-mixed text generation* in eight different code-mixed language pairs *viz.* English-Hindi, English-Bengali, English-Malayalam, English-Tamil, English-Telugu, English-French, English-German and English-Spanish using English corpus.

This dissertation demonstrates the effectiveness of developing multilingual QA systems for handling language versatile user queries. Experiments show our models achieve state-of-the-art performance on answer extraction, ranking, and generation tasks on multiple domains of MQA, VQA, and language generation. The proposed techniques are generic and can be widely used in various domains and languages to advance the QA systems.

***Keywords:*** Natural Language Processing, Question Answering, Multilingual Question Answering, Machine Comprehension, Code-Mixing, Code-Switching, Taxonomy, Answer Extraction, Question Classification, Question Generation, Machine Learning, Deep Learning, Neural Network,

# List of Tables

# List of Figures

# Abbreviations

| | |
|---|---|
| NLP | Natural Language Processing |
| QA | Question Answering |
| MQA | Multilingual Question Answering |
| TF-IDF | Term Frequency–Inverse Document Frequency |
| CV | Computer Vision |
| Trans | Translation |
| Q | Question |
| Ans | Answer |
| VQA | Visual Question Answering |
| MCVQA | Multilingual and Code-Mixed Visual Question Answering |
| MFB | Multi-modal Factorized Bilinear Pooling |
| MFH | Multi-modal Factorized High-order Pooling |
| CM | Code-Mixing |
| QQ | Question-Question |
| QG | Question Generation |
| SQC | Semantic Question Classification |
| SQR | Semantic Question Ranking |
| MQG | Multihop Question Generation |
| RCNN | Recurrent Convolutional Neural Network |
| CNN | Convolution Neural Network |
| LSTM | Long Short Term Memory |
| GRU | Gated Recurrent Unit |
| MAP | Mean Average Precision |
| MRR | Mean Reciprocal Rank |
| RNN | Recurrent Neural Network |
| RL | Reinforcement Learning |
| MLE | Maximum Likelihood Estimation |
| EM | Exact Match |
| BLEU | Bilingual Evaluation Understudy |
| ROUGE | Recall-Oriented Understudy for Gisting Evaluation |
| METEOR | Metric for Evaluation of Translation with Explicit Ordering |
| P | Precision |
| R | Recall |
| F | F1-Score |
| R@k | Recall at k |
| P@k | Precision at k |
| PoS | Parts-of-Speech |
| NER | Named Entity Recognition |
| SVM | Support Vector Machine |
| ML | Machine Learning |
| DL | Deep Learning |
| NN | Neural Network |
| Bi-LSTM | Bi-directional Long Short Term Memory |
| Bi-GRU | Bi-directional Gated Recurrent Unit |
| MTL | Multi-task Learning |
| STL | Single-task Learning |

# Contents

# Chapter 1

# Introduction

A question-answering (QA) system aims to automatically extract/generate an answer(s) for a given question from the data repository (e.g., web, document, passage, etc.). Web search engines can be treated as an alternative to finding a response to the users' queries/ questions. However, web search engines have to process billions of web documents on multiple topics to return a response to users' queries. It is often that, web search engines return hyperlinks to multiple web documents as per user query. The majority of these documents do not contain useful information at the demand, and only the top few documents constitute the actual information. Question answering offers a solution to address the limitation of search engines by providing specific information needs. Consequently, there is a need for specific information (e.g. *Who won the National Book Awards in 2016?*) rather than general information (e.g. *National Book Awards in 2016?*).

A search engine and a QA system differ in design, objectives, and processes. A classical search engine aims to deliver snippets or documents for a given query, whereas a QA system strives to deliver the exact answer to a natural language question. In other words, we can say that a QA system is an extension to the search engine with additional functionalities. For building a QA system, usually, three sub-processes are followed **(i)** *question classification* **(ii)** *document(s) extraction* and **(iii)** *appropriate answer(s) extraction.*

In the literature [125], there are three major paradigms in question answering.

1. **Information Retrieval (IR) based Question Answering:** This paradigm is mainly focused on retrieving the answer from the collections of documents for factoid-type questions. There are three steps of an IR-based factoid QA system:
    (a) **Question Processing**: This is the first step where *answer type* is detected

by classifying the questions into the set of entity types. The answer type can typically be the named entities (NEs), such as '*person name*', '*location name*', '*time*', etc.). This step also performs the query formulation by selecting the word having nouns, adjectives, or verbs as their parts-of-speech tag. The **query** specifies the keywords that are used by the IR system to search the relevant documents. For e.g.,

**Question (1):** *Who was the first Russian astronaut to do a spacewalk?*

**Query:** *first Russian astronaut, spacewalk*

**Answer Type:** *Person Name*

(b) **Passage Retrieval:** The documents against the user query are extracted and irrelevant documents are discarded that do not have the potential answer. Thereafter, the documents are ranked based on their similarities with the users' query.

(c) **Answer Processing:** This is the last step, where the specific answer is extracted from the documents. Expected answer types, together with regular expression patterns, are used to find the final answer to the question. E.g., for Question (1) the following document contains the exact answer (*'Alexey Arkhipovich Leonov'*).

**Document:** *Alexey Arkhipovich Leonov is a retired Russian cosmonaut, he becomes the first human to conduct extravehicular activity (EVA), exiting the capsule during the Voskhod 2 mission for a 12-minute spacewalk.*

Alternatively, reading comprehension algorithms can be used to read these documents/passages and extract the start & end position of the answer from the text.

2. **Knowledge-based Question Answering:** In this paradigm of QA, the answer is extracted from the structured database instead of the text documents. First, the questions are mapped to a logical form query. These logical forms can be predicate calculus or query languages like SQL or SPARQL. Based on the underlying database, the query is fired to retrieve the answer from the database. E.g.

**Question (2):** *Which is the worst affected country from COVID-19?*

**Logical Form:** `worst-affected-country (COVID-19, ?x)`

3. **Hybrid Question Answering:** In the hybrid QA paradigm, answers are extracted from various information sources rather than only text-based or knowledge-based sources. One of the successful systems built on a hybrid QA paradigm is IBM's Watson [65], which won the Jeopardy![1] challenge in 2011.

The one major limitation of current QA systems is the support interaction in a single language, specifically, English, even though the number of speakers in other languages (Spanish and Mandarin) is more than English speakers, closely followed by the Hindi speaker. It extremely inhibits the ability of a multilingual end-user to connect and communicate with the QA system. To give an idea, when a native Hindi speaker wants to know the winner of the '*Best Picture*' for the Oscar 2019, s/he is more likely to express it as:

**Question (3):** "*Kaun si picture ne Oscar 2019 ke best picture ka award jita?*"

(**Translation**: *Which movie won the best picture award in the Oscar 2019?* )

where the words '*picture*', '*best*', '*award*' are all English words, and the rest are written in the roman script (Hindi). This phenomenon of embedding the morphemes, words, phrases, etc. of one language into another is popularly termed code-mixing (CM). Therefore, to increase the horizon, impact, and efficacious of QA in multilingual societies, it is indispensable to support QA in multilingual and code-mixed languages.

## 1.1 Scope of Dissertation

The first part of this dissertation proposes novel techniques for developing code-mixed and multilingual question-answering systems. The later part of this dissertation focus on improving the question-answering module with various QA sub-tasks like semantic question matching, complex (multi-hop) question generation, and code-mixed text generation. We propose various neural QA and generation techniques by effectively modeling the linguistics phenomenon of multilingual and code-mixed texts. The proposed frameworks are generic, and can easily be adapted to any other domains and languages without much effort.

Towards our first research problem: *multilingual question answering (MQA)*, we begin by providing details about the creation and annotation scheme of the English-Hindi question answering dataset followed by extensive data analysis. We benchmark the dataset by developing an IR-based framework that utilizes the machine translation system for non-English questions and answers. We further extend the study by developing a neural approach for solving the multilingual question-answering problem. Towards this, we utilize the popular English QA dataset, SQuAD [216], to generate our synthetic English-Hindi dataset. More specifically, we propose a unified deep neural network framework to retrieve the multilingual answer by exploring the attention-based recurrent neural network to generate the adequate representation of multilingual questions and snippets. We utilize the soft alignment of words from English and Hindi questions to generate a single shared representation of questions. To generate the snippets, we propose a snippet-generation algorithm. The algorithm takes into account the semantic information with lexical similarity to rank the probable sentences by considering their relevance to the question. Along with this, we represent the sentences of documents as a graph, where each pair of sentences are linked based on their lexico-semantic similarity (obtained through word embeddings) toward the question. The proposed system's effectiveness is demonstrated to extract the answer to an English and/or Hindi question from English and/or Hindi snippets. Our experiments on a recently released multilingual QA dataset show that our proposed model achieves state-of-the-art performance.

---

[1]<https://en.wikipedia.org/wiki/Jeopardy!>

In our second research problem, we explored the associated dimension of multilingual QA– *code-mixed question answering*. We propose a robust framework for code-mixed question generation (CMQG) and code-mixed question answering (CMQA) involving English and Hindi languages. First, we develop a linguistically-motivated code-mixed question generation technique. We followed this approach, as we did not have access to any labeled data for code-mixed question generation. Thereafter, we propose an effective framework based on a deep neural network for CMQA. In our proposed CMQA technique, we use multiple attention-based recurrent units to represent the code-mixed questions and the English passages. Finally, our answer-type focused network (attentive towards the answer type of the question being asked) extracts the answer for a given code-mixed question. We provide a detailed experimental setup along with the results and analysis of the benchmark datasets.

In our third and last research problem of part I, we focus on *multilingual and code-mixed visual question answering*. In the literature, there are several techniques available for monolingual (especially English) datasets for which robust Visual Question Answering (VQA) model can be developed. However, we are interested in building a system that can answer questions from different languages (multilingual) and the language formed by mixing multiple languages (code-mixed). Toward this, we create a dataset for VQA in a multilingual and code-mixed environment. Thereafter, we utilize the dataset to propose a unified model that can handle natural language questions from multiple languages to find the answer from a given image.

We show that in a cross-lingual scenario due to language mismatch, applying a learned system directly from one language to another language results in poor performance. Thus, we propose a technique for multilingual and code-mixed VQA. Our proposed method mainly consists of three components. The first is the *multilingual question encoding*, which transforms a given question into its feature representation. This component handles multilingualism and code-mixing in questions. We use multilingual embedding coupled with a hierarchy of shared layers to encode the questions. To do so, we employ an attention mechanism on the shared layers to learn language-specific question representation. Furthermore, we utilize self-attention to obtain an improved question representation by considering the other words in the question. The second component (*image features*) is to obtain an effective image representation from object-level and pixel-level features. The last component is *multimodal fusion*, which is accountable to encode the question-image pair representation by ensuring that the learned representation is tightly coupled with both the question (language) and image (vision) feature.

In the first research problem of Part II of this dissertation, we focus on creating a taxonomy for natural language questions to improve the performance of the *neural semantic question matching* and retrieval system. Towards this, we create a semantic question-matching dataset, build a taxonomy, and propose an efficient technique to match and retrieve semantically similar questions from the pool of questions.

In the proposed semantic question-matching framework, we use attention-based neural network models to generate question vectors. We create a hierarchical taxonomy by considering different types and sub-types so that questions with similar answers belong to the same taxonomy class. We propose and train a deep learning-based question classifier network to classify the taxonomy classes. The taxonomy information is helpful in making a decision on semantic similarity between them. Taxonomy can provide very useful information when we do not have enough data for generating useful deep learning-based representations, which is generally the case with restricted domains. In such scenarios, linguistic information obtained from prior knowledge helps significantly improve the system's performance. Empirical evidence establishes that our taxonomy, when used in conjunction with deep learning representations, improves the performance of the system on semantic question (SQ) matching tasks.

In the second research problem toward improving the QA system, we exploited the task of generating complex (multi-hop) questions. We propose to tackle the *multi-hop question generation* problem in two stages. In the first stage, we learn supporting facts aware encoder representation to predict the supporting facts from the documents by jointly training with question generation and subsequently enforcing the utilization of these supporting facts. The former is achieved by sharing the encoder weights with an *answer-aware* supporting facts prediction network, trained jointly in a multi-task learning framework. The latter objective is formulated as a *question-aware* supporting facts prediction reward, which is optimized alongside supervised sequence loss. Additionally, we observe that the multi-task framework offers substantial improvements in the performance of question generation and also avoid the inclusion of noisy sentence information in generated question, and reinforcement learning (RL) brings the complete and complex question to otherwise maximum likelihood estimation (MLE) optimized QG model.

In our final research work, we investigated the novel method for *code-mixed text generation* in Indian and European languages. We model the code-mixed phenomenon using the feature-rich and pre-trained language model-assisted encoder-decoder paradigm. The feature-rich encoder helps the model capture the linguistic phenomenon of code-mixing, especially to decide when to switch between the two languages. Similarly, the pre-trained language model provides the task-agnostic feature, which helps the model encode the generic feature. We adopt the gating mechanism to fuse the features of the pre-trained language model and the encoder. In addition, we also perform transfer learning to learn the prior distribution from the pre-trained neural machine translation (NMT). The pre-trained NMT weights are used to initialize the code-mixed generation network. Our manual evaluation shows that generated code-mixed texts are almost syntactically correct and fluent across the language pairs.

## 1.2 Applications

Several applications can be drawn from this dissertation. Some of the significant applications are as follows:

- **Conversational Agents:** Conversational agents, also known as dialog agents,

interpret and responds to statements made by users in ordinary natural language. It helps users in various forms, from providing weather reports to interacting with their everyday devices. Some of the popular agents are Google Assistant[2], Siri[3], Cortona[4], Amazon Alexa[5], etc. It has gained visibility and success in recent years. The access to the agents has increased manifold because of the ease and wide access to the internet and mobile devices. Although the conversational agent has established itself in society and gradually becomes an essential part of our life, it still lacks in assisting the community with multiple speaking languages. The deep learning algorithm used to build such a system is highly dependent on the monolingual data, especially English. Due to this, they often fail to respond and assist multilingual communities with their mixed query of different languages. In this dissertation, we provide an algorithm that can address the problem of handling multilingual queries.

- **Biomedical Question Answering:** With the unprecedented growth of scientific literature in the biomedical domain, and due to the significant impact of such information on several real-world applications [311, 348, 324, 321, 313, 322, 317], there is a particularly essential and increasing demand for biomedical QA systems that can efficiently and effectively assist health-care professionals and biomedical researchers in their information search. The work done in this dissertation can help in advancing the biomedical QA system in two ways, *viz.* **(i)** by identifying the similar questions which have already been asked whose answers are readily available in the database; **(ii)** introducing the multilingual and code-mixing features in the biomedical QA system to handle the queries posed in those environments and provide a second opinion to the medical practitioner in an initial understanding of the consumer-health query [310, 320] of the multilingual end-users.

- **Online Knowledge Service:** There are many online knowledge services where students or professionals can ask their questions, and these services have the capability to process the question and give the answer in a legitimate format. One such service is WolframAlpha[6], which uses computational intelligence and knowledge bases to deduce the solution with a step-by-step explanation. As a beneficial application for the student, the online knowledge source can be further enhanced with the generation, multilingual query handling, and other features that assist the larger parts of the end-users.

---

[2] https://assistant.google.com/
[3] https://www.apple.com/in/siri/
[4] https://en.wikipedia.org/wiki/Cortana
[5] https://alexa.amazon.com/
[6] https://www.wolframalpha.com/

- **Clinical VQA:** Nowadays, medical and physiological images ("ct-scan", "x-ray", etc.) and reports for patients are easily accessible with the increase in the use of medical portals. However, the patient still needs to visit a medical expert to fully understand those reports and get answers to their queries. This process is both time-consuming and cost-sensitive. However, some existing VQA systems [54, 94] can be sought as an alternative, but they lack multilingual and code-mixing capability. VQA system inbuilt with handling multilingual questions will be the best option for such an application.

- **Multilingual Search Engine:** The limited ability of modern search engines towards the multilingual and code-mixed query is discussed in Section 3.1.2 and Section 4.1.2. Advancing this search engine is also a significant application aligned with the work presented in this dissertation. This will further improve the user experience and accuracy in retrieving the relevant document or answer. We also provide the generic approach to extend the bilingual query handling to multiple and mixed questions. Additionally, we also devise a technique to utilize the existing resource-rich dataset (English) to train and improve the performance of the current MQA system.

- **Improving Customer Experience in E-commerce:** The e-commerce site can also improve the product search capability from monolingual to multilingual and mixed script queries. We discuss the approach in this dissertation, which can either transform the multilingual query to an English query or directly process the multilingual query to retrieve the answer (here in terms of product) from the information sources. This could be an essential feature for the e-commerce site to increase its customer base and experience in bilingual countries like India, China, Singapore, etc. By enabling this feature, they can show more relevant content to the customer who searched with a mixed query. The firms can also attract the customer base with the appropriate ads, directly correlated to the customer's linguistic origin, leading to better customer satisfaction.

- **Assisting Multilingual Communities:** Apart from the various applications discussed aforementioned, there are some other applications as well, whereby properly handling the multilingual query can be used to assist the multilingual communities—for example, assisting patients with their multilingual queries towards their clinical and radiology reports, finding the appropriate responses after querying the automated help desk system, helping the persons in natural calamities by providing them safety instructions in regional languages, etc.

## 1.3 Contribution of Dissertation

In this dissertation, we solve some of the problems to advance and improve the existing question-answering system. These tasks have a wide array of applications ranging from solving other high-level Natural Language Processing (NLP) tasks (information retrieval, question answering, automatic summarization, etc.) to the development of virtual assistants. We use the data from various sources like Wikipedia, news articles, NLP and vision literature, etc. Literature shows that there have been prior works in these areas, but mainly limited to a few popular languages, especially English, and domains. Below we highlight the major contributions of the dissertation:

- In Chapter 3, we create a benchmark setup for multilingual QA focusing on English and Hindi languages. We also propose a unified end-to-end deep neural network model for multilingual QA, where questions and answers can be either in English or Hindi, or both. We propose a deep neural network framework by utilizing the soft alignment of words from English and Hindi questions to generate a single shared representation of questions. This effective shared representation is used by the pointer network to select the position of the answer in the passage. Our proposed model achieves state-of-the-art performance for multilingual QA on the benchmark MQA dataset.
- In Chapter 4, we propose a linguistically motivated unsupervised algorithm for Code-Mixed Question Generation. We also create two Code-Mixed Question Answering datasets: CM-MMQA and CM-SQuAD to explore the research on CMQA further. Finally, we provide a state-of-the-art answer-type focused neural framework utilizing bi-linear attention to extract answers from the English passages for the corresponding code-mixed questions. The answer-type information guides the network to disambiguate the answer candidate and provides the cue to correctly predict the answer from the passage.

- In Chapter 5, first, we create linguistically driven Hindi and code-mixed VQA datasets. To the best of our knowledge, this is the very first attempt toward VQA. We propose a simple, yet powerful mechanism based on the soft-sharing of shared layers to better encode the multilingual and code-mixed questions. This bridges the gap between VQA and multilingualism. Our proposed unified neural model for multilingual and code-mixed VQA can predict the answer to a multilingual or code-mixed question. The evaluation shows that our proposed multilingual model achieves state-of-the-art performance in all these settings.

- In Chapter 6, we introduce a two-layer question taxonomy to categorize the questions at coarse and fine-grained levels. We propose a framework to integrate semantically rich taxonomy classes with a neural network-based encoder to improve the performance and achieve new state-of-the-art results in semantic question ranking and classification on benchmark datasets. Finally, we release two manual annotated datasets, one for semantically similar questions and the other for question classification.

- In Chapter 7, we introduce the problem of multi-hop question generation and propose a multi-task training framework to generate the questions from multiple documents. We formulate a novel reward function, to enforce the maximum utilization of supporting facts to generate a question. We develop an automatic evaluation metric to measure the coverage of supporting facts in the generated question. Empirical results show that our proposed method outperforms the current state-of-the-art single-hop QG models over several automatic and human evaluation metrics on the HotPotQA dataset.

- In Chapter 8, we propose a robust and generic method for code-mixed text generation. Our method is tailored to generate syntactically correct, adequate, and fluent code-mixed sentences using the prior knowledge acquired by the transfer learning approach. To the best of our knowledge, this is the first step in proposing a generic method that produces the correct and fluent code-mixed sentences on multiple language pairs. We demonstrate with detailed empirical evaluations the effectiveness of our proposed approach on eight different language pairs, *viz.* English-Hindi (en-hi), English-Bengali (en-bn), English-Malayalam (en-ml), English-Tamil (en-ta), English-Telugu (en-te), English-French (en-fr), English-German (en-de) and English-Spanish (en-es).

In the next chapter, we present a survey on the various existing works in question answering, code-mixing, multilingual question answering, reading comprehension, question generation, code-mixed text generation, and metrics used to evaluate and compare the performance of the proposed approaches.

# Chapter 2

# Literature Survey

In this chapter, we provide a survey and analysis of the existing approaches for multilingual and code-mixed question answering. We also discuss the existing works in semantic question matching, question generation, and code-mixed text generation. Section 2.1 gives a brief survey of the works where multilingual and code-mixing challenges are addressed in the literature. In Section 2.2, we aim to provide readers with a detailed discussion of the existing works on reading comprehension. Section 2.3 offers a brief survey on the dataset and techniques for cross-lingual and multilingual question answering. Section 2.4 provides the survey of existing datasets and methodology in VQA. Section 2.5 provides the details about the existing approach to semantic question matching, retrieval, and answer sentence selection. In Section 2.6, we provide a review of question generation literature and the multi-hop question-answering approach. We also discussed the approaches (Section 2.7) presented in the literature to generate the code-mixed text. Furthermore, we discussed the various evaluation metrics (Section 2.8) used to assess the performance of the proposed models on benchmark datasets.

## 2.1 Multilingual and Code-Mixing in NLP

Multilingualism is the act of using, or promoting the use of, multiple languages, either by an individual speaker or by a community of speakers. It is becoming a social phenomenon governed by the need for globalization and cultural openness. Code-mixing is another dimension of multilingualism where people mixes the script of multiple languages to express themselves. Formally, code-mixing refers to the phenomenon of mixing more than one language in the same sentence. Creating resources and tools capable of handling multilingual and code-mixed languages is more challenging compared to the traditional language processing activities concerned with only one language. In recent times, researchers have started investigating methods for creating tools and resources for various NLP applications involving multilingual and code-mixed languages. In this section, we discuss the recent NLP works, such as distributional semantics, sentiment analysis, and other NLP problems focused on multilingual and code-mixed languages.

In linguistics, word embedding comes under the research area of distributional semantics. The recent advancement in word embeddings has shown great success for downstream NLP tasks. Along this line of research, [63] proposed an approach to learning bilingual word vectors using canonical correlation analysis (CCA) [11]. They showed that the combination of CCA and dimensionality reduction improved the performance of monolingual vectors on standard evaluation tasks. Their approach was further extended by [8] for multilingual word embeddings. They proposed a multiCCA and multiCluster approach to map 59 languages into a single shared space. [255] trained offline word vectors and proved that the linear transformation between the two spaces should be orthogonal to each other. This phenomenon is known as the orthogonal Procrustes problem [234]. They achieved the orthogonal transformation using the singular value decomposition. In another prominent study conducted by [36] proposed an unsupervised algorithm to train multilingual word embedding using only monolingual corpora. They proposed methods to exploit inter-dependencies between any two languages and map all the monolingual embeddings into a shared multilingual embedding space via a two-stage algorithm. [147] proposed a technique to learn cross-lingual word embeddings by building a bi-lingual dictionary using adversarial learning. Another study conducted by [6] proposed an unsupervised multilingual alignment method that maps every language into a common space while minimizing the impact on indirect word translation. Recently, [117] proposed a geometric approach for learning multilingual word embeddings by aligning the word embeddings of various languages in a common space. [211] advocated for special word embedding for code-mixed language. Their study demonstrates that the existing processes of bilingual embedding techniques are not ideal for code-mixed text processing, and there is a need for learning multilingual word embedding from the code-mixed text. In a similar work line, [34] proposed a deep recurrent belief network for learning word dependencies in the text, which uses Gaussian networks with time delays to initialize the weights of each hidden neuron.

Learning the long-term dependencies in a text helps the model make the correct decision in classification or sequence labeling tasks.

Sentiment analysis of multilingual and code-mixed data involves multiple opinion-mining applications in multilingual communities, including customer satisfaction and social campaign analysis. Innovations in this direction are obstructed by the absence of an adequate annotated dataset. [227] performed an interesting study by considering the tweets from Indian bilingual users. They try to understand users' preferred language for expressing opinions and sentiments. Firstly, they detected the opinionated tweets and then classified them into positive, negative, and neutral sentiments. The study indicates that the user uses the Hindi language to express negative sentiments. A similar work towards language dominance in Spanish-English has also been done by [262]. [121] introduced a sub-word level representation learning using the Long Short-term Memory (LSTM) for code-mixed English-Hindi sentiment analysis. They argued that sub-word level representations are effective in learning the information about the sentiment value of important morphemes instead of character-level or word-level representations. [285] proposed the approaches to perform multilingual polarity classification in monolingual, multilingual, and code-mixed environments. Their approaches utilized rich features to build the machine learning-based model focusing on English and Spanish languages. There are some more works [166, 254] where the code-mixed and multilingual challenges are handled in sentiment analysis.

Aspect-based sentiment analysis (ABSA) [97, 83] is an extension of the sentiment analysis task, and it is involved with the entities and their aspect in the text. [4] explore and propose the two-step method for ABSA by utilizing the particle swarm optimization-based feature selection technique to identify the relevant set of features for the ABSA task. [171] tackled the targeted aspect-based sentiment analysis (TABSA) and proposed a method to integrate the common-sense knowledge in the deep neural network for TABSA. In order to infuse explicit knowledge, they propose an extension of LSTM named Sentic LSTM. Sentic LSTM can utilize explicit knowledge to control the information flows from time to time.

There are some other NLP areas like parts-of-speech [259, 203, 95], question classification [260, 261], language identification [257, 106, 258, 96], entity extraction [84, 92, 20], sentiment analysis [88], hate speech detection [40, 218] etc, where code-mixing phenomena are explored and analyzed.

## 2.2 Reading Comprehension

Building a machine to understand and comprehend human language documents is one of the long-standing challenges in Artificial Intelligence. Reading comprehension (RC) is a task to answer comprehension questions by reading a passage or document. In this dissertation, since we aim to retrieve the answer for multilingual (Chapter 3) and code-mixed (Chapter 4) questions from documents, we formulated the task of QA as a reading comprehension problem. The field of building automated reading comprehension has a long history. In the 1970s, [148] proposed a computational model for question answering. Their study focused on the story understanding in which a question-answering system was used as a demonstration of comprehension. [110] introduced a reading comprehension system named 'Deep Read' by creating a corpus of 60 development and test stories of $3^{rd}$ and $6^{th}$ grade educational materials. They developed a rule-based system using bag-of-words and shallow linguistic features, such as stemming, semantic class identification, and pronoun resolution. Most of the proposed methods on the reading comprehension tasks can be broadly categorized into *machine learning* and *deep learning* approaches, discussed as follows:

### 2.2.1 Machine Learning Approaches

The human-labeled training examples (question, passage, answer) are required to build a supervised machine-learning system for reading comprehension. Towards this, two popular datasets, namely MCTest [224] and ProcesBank [18] were created. MCTest dataset consists of two sub-datasets, MC160 and MC500 with 160 and 500 fictional stories, respectively. Each story in the dataset has 4 multiple choice questions, where each question has 4 answers, and one of them is correct. ProcesBank dataset comprises of 585 questions over the 200 paragraphs. Each paragraph in the dataset describes a biological process, and the goal is to answer questions that require an understanding of the relations between the entities and events in the process.

[224] proposed rule-based baseline systems: sliding window and window with distance. Both the baseline systems measure the weighted word overlap/distance information question words and answer. Another study conducted by [229] explored MCTest dataset for machine comprehension task. They posed the problem as the entailment task, where the hypothesis was that there is a hidden structure that explains the relation between

the question, correct answer, and passage. The model uses the max-margin framework to find the hidden structure. Another machine learning-based approach was introduced by [288] with rich linguistic features, such as semantic frames, word embeddings, syntax information, etc., to address the machine comprehension problem. Another prominent work by [191] proposed a discourse-aware method that utilizes various linguistic features, such as syntactic dependency [316, 195], entity matching, entity relationship [323], and co-reference resolution for reading comprehension tasks.

### 2.2.2 Deep Learning Approaches

The availability of large-scale reading comprehension datasets has led to the origin of several deep-learning techniques for reading comprehension. The first attempt to create a large-scale reading comprehension dataset was made by [105]. They created two machine comprehension datasets by exploiting the online newspaper articles and their matching summaries from CNN[1] and Daily Mail[2] news portals. Their proposed '*Attentive Reader*,' a neural network model for machine comprehension, outperformed the machine learning-based approaches by a fair margin. Another study by [35] provided a detailed analysis of the CNN and Daily Mail datasets and extended the previous state-of-the-art '*Attentive Reader*' to further enhance the system performance by introducing the task-specific component. They also argued that the dataset is noisy and it provides very little challenge to the machine comprehension system. To address these limitations, [216] released another dataset named SQuAD. The SQuAD dataset is built on Wikipedia articles, and the questions and answers are created by the crowd workers. The answer to every question in the dataset is a span of text from the corresponding passage.

In recent years, plenty of machine reading comprehension (MRC) models have been developed. A Bi-Directional Attention Flow, in short, BiDAF network for reading comprehension, is proposed in [240]. BiDAF consists of a hierarchical architecture to encode the context representation at different levels of granularity. It encodes the words in question and context by three different levels of embeddings: character, word, and contextual. The best point of this architecture is the use of bi-directional attention flow from a query (question) to a paragraph and vice-versa, which provides complementary information to each other. With the help of bi-directional attention, they computed the query-aware context (paragraph) representation. The attention operation is performed at each time step to obtain an attended vector. The obtained attended vector and representations from the previous layers are passed to the next layer in the architecture.

---

[1] http://www.cnn.com/
[2] http://www.dailymail.co.uk/

Match-LSTM model [294] proposed a neural-based solution for machine comprehension tasks. The proposed framework is based on the match-LSTM and Pointer Net [286] to point the answer in the given input context or passage. The model provides two different ways to obtain the answer: *sequence* and *boundary*. In the *sequence* model, the proposed architecture predicts the sequence of answer tokens. In the *boundary* model, it only predicts the start and end indices of the answer in the original passage. The words present between the start and end indices are considered to be the answer sequence. The *boundary* model performs better compared to the *sequence* model. Recently, [113] introduced the reinforced mnemonic reader for the MRC tasks. The proposed model improves the attention mechanism by introducing a re-attention mechanism to re-compute the current attention. In addition to this, the authors also introduced dynamic-critical reinforcement learning, which dynamically decides the reward needs to be maximized.

The QANet model [333] is different from the other neural-based approaches for reading comprehension. The majority of the approaches exploit the RNNs (LSTM or GRU) and attention mechanisms. Unlike the other approaches, QANet focused on convolution and self-attention techniques. A pointer network with an attention mechanism was introduced by [127] for machine comprehension. [246] proposed a method capable of iteratively inferring the answer with a dynamic number of reasoning steps and is trained with reinforcement learning. A joint model of question generation and answering based on a sequence-to-sequence neural network model is proposed in [295], where a generative machine comprehension model learns jointly to ask and answer questions based on the given documents.

Some of the other notable works using the advancement of deep neural networks are [293, 39, 246, 296, 282]. Machine comprehension problems [24, 143, 107] have also been studied with a purview of leveraging the memory networks [271].

Recently, Transformer [284] based approaches have gained visibility and achieved state-of-the-art performance on the machine comprehension task. [52] proposed the pre-trained network named Bidirectional Encoder Representations from Transformers (BERT) that outperformed the previous deep learning-based approaches on several NLP tasks. They also demonstrated the effectiveness of the BERT model on reading comprehension tasks.

Thereafter, various other pre-trained networks [164, 329, 123] based on Transformer architecture have pushed the state-of-the-art performance close to the human-level performance.

## 2.3 Multilingual Question Answering

Multilingual question-answering is an important component of an accessible natural language interface. In the recent past, there has been a remarkable progress in question answering and reading comprehension. However, the majority of these works have focused only on the English language. There have been few studies conducted on multilingual QA that have explored machine-learning and deep-learning based techniques which we present below.

### 2.3.1 Machine Learning Approaches

In the literature, there have been few attempts towards developing multilingual QA systems [75, 68, 196]. The majority of these works made use of machine translation, where questions and/or documents in less-resourced languages were translated into resource-rich language(s) like English. In these works, the motivation has been to utilize the resources and tools available in resource-rich languages. [72] described the main characteristics of multilingual QA systems. Further, they analyzed the quality of the output produced by the machine translation systems (Google Translator[3], Promt[4] and Worldlingo[5]). The obtained results showed the potential in the context of multilingual question answering.

There are a few existing works related to QA in English and Indian languages. [144] implemented the Hindi search engine. The task of the search engine is to retrieve relevant passages from the collection of the passages. In the proposed architecture, various modules were introduced. The automatic Entity Generator module was developed to identify the domain-related entities from which users can ask questions. It also has question classification, answer extraction, and answer selection modules. The question classification module deals with several categories of questions. The answer extraction module extracts the answer. By using ranking, the answer selection module selects the answer among the candidate answers. [231] discussed an approach for the question-answering system focused on the Hindi language. This work deals with four types of questions: '*when*', '*where*', '*how many*', and '*what time*'. For a given question, the answer was retrieved from the Hindi text. Each sentence in the text was analyzed to understand its meaning. In this work, they represented the questions using query logic language (QLL), which is a subset of Prolog[6]. For the identification of nouns, verbs, and question words in Hindi, the shallow parser was used. [269] implemented the web-based Hindi question-answering system. In this work, the question and answer deal with only the Hindi language; if the answer was not presented in the Hindi document, it was retrieved from Google[7] search engine. A question-answering system for Hindi and English was developed by [236] in which the questions and answers were in Hindi. These question-answer pairs were retrieved from the Hindi newspaper. [220] proposed a question-answering system in the Telugu language. The system was dialogue-based and railway-specific domain. The architecture was based on the keyword-based approach. The query analyzer generates the tokens and keywords. From tokens, SQL statements were generated. Using SQL query, the answer was retrieved from the database. [98] develop the question-answer system in English and Punjabi language. In this work, a pattern and a matching algorithm were introduced to retrieve the most relevant and appropriate answer from multiple sets of answers for a given question.

---

[3] https://translate.google.com/
[4] https://www.online-translator.com/
[5] http://www.worldlingo.com/microsoft/computer_translation.html

In another line of work, a language-independent passage retrieval system for multilingual question answering was proposed by [266]. It has the advantage of processing the question and the passages without using any knowledge about the lexicon and the syntax of the corresponding language. They demonstrated the performance of the system on Spanish language query and passage documents. Similarly, [355] presented the QALL-ME Framework to build multilingual and cross-lingual question answering that answer questions from the structured data source. The framework is based on a series of steps, such as language identification, entity annotation, term annotation, and query generation. They demonstrated the framework for English, German, and Spanish language query.

[172] developed DIOGENE, a multilingual question-answering system based on standard IR-based pipeline architecture of question answering. Multilingual wordnet is used to bring multilingualism in the proposed DIOGENE system. [278] build a multilingual question-answering system that can retrieve the answer from the knowledge base for a

---

[6]http://www.gprolog.org/
[7]https://www.google.com/

question asked in different languages. The proposed system uses the grammatical analyzer and template-based matching to find the answer from the Wikidata knowledge base. [353] introduces the Multilingual Grammatical Question Answering (MuG-QA) system for answering questions in the English, German, Italian and French languages over DBpedia. The architecture exploited the Grammatical Framework [219] to support multilingualism.

Developing a QA system in a code-mixed scenario in itself is very novel in the sense that there have not been very significant attempts in this direction. [31] conducted the very first study and proposed an end-end web-based factoid QA system for code-mixed languages. The system is based on the standard IR technique for QA. They demonstrate the system's performance on two code-mixed language pairs, Hindi-English and Telugu-English.

### 2.3.2 Deep Learning Approaches

Recently, large-scale cross-lingual and multilingual question answering are being created, which enables the use of recent advancements of deep learning in multilingual question answering. Towards this, [48] built an answer sentence selection dataset and proposed the various approaches to retrieve the answer to questions asked in different languages. In this dataset, the answer sentences are only in English; however, questions are in 11 different South-African languages. They provide the experimental setup and results on naive Bayes, exact and approximate k-nearest neighbors on cross-lingual word embeddings, as well as long-short term memory networks trained end-to-end. Towards this, [163] created a dataset named XQA to investigate the cross-lingual open-domain question answering (OpenQA). This dataset contains training sets in English and test and development sets in eight other languages. They also provide several baseline systems for cross-lingual OpenQA, focusing on machine translation-based methods and zero-shot cross-lingual methods. [150] presented MLQA, a highly-parallel multilingual QA benchmark in seven languages. MLQA dataset consists of $12K$ instances in English and $5K$ in each other language, with each instance parallel between 4 languages on average. Similar to the [163], MLQA also provides a strong baseline and state-of-the-art on the developed cross-lingual MLQA dataset. [44] released TyDi QA, a QA dataset covering typologically diverse languages with $204K$ question-answer pairs. They also provided a detailed analysis of the dataset at language-level and example levels. The dataset offers the bench-mark for two sub-problems of question answering: passage selection and answer span identification tasks.

Dialogue systems are another kind of QA system which can handle multiple turns from the users. A turn can consist of a sentence, although it might be as short as a single word or as long as multiple sentences. [308] proposed a generative dialogue model for task-oriented dialogue systems. The proposed model is end-to-end trainable under unsupervised, semi-supervised, or reinforcement learning frameworks and augmented with discrete latent intention inferred from the user's utterance. Their experimental studies illustrate the benefit of the discrete latent variable model compared to a fully deterministic [265] and continuous latent variables [242]. [331] investigate to exploit the commonsense knowledge to improve end-to-end dialogue systems by selecting the appropriate response. They encode the commonsense using the LSTM block and use it as external memory to the system. While most of the work on dialogue systems focused on text-based conversation, recently [332] explores the impact of incorporating the audio features of the user message into generative dialogue systems. Firstly, they learn the appropriate audio representation by training a binary classifier, distinguishing between the real and randomly assembled audio context and response. Finally, they augment the audio representation as the word-level modality of the context in an encoder-decoder framework to generate the response.

## 2.4 Multilingual and Code-Mixing in VQA

### 2.4.1 VQA Datasets

Quite a few VQA datasets have been created to encourage multi-disciplinary research involving NLP and Computer Vision (CV). To this end, [251] released a dataset named as NYU-Depth V2. It is comprised of video sequences from a variety of indoor scenes as recorded by both the RGB and Depth cameras. It consists of 1449 RGBD images together with annotated semantic segmentation, where every pixel is labeled into some object class. This dataset is collected from commercial and residential buildings in three different US cities. [174] collected question-answer pairs by considering synthetic and human annotations. The synthetic question-answer pairs are automatically generated question-answer pairs by using the basic templates. They also created the question-answer by human annotations. The answer in this dataset is either basic colors, numbers, or objects or sets of those.

[71] started with 158,392 images from the MS COCO dataset [161]. The annotations are collected using Baidu's online crowd-sourcing[8]. The annotators were asked to provide question-answer pairs by looking at the image. The questions are formulated in such a way that they should be answered by visual content and commonsense. Another VQA dataset, Visual7W dataset [351] contains 1.7 million QA pairs of the 7 different *wh* question types. Moreover, this dataset also includes extra annotations such as object groundings, and multiple choices, and human experiments, making it a clean and complete benchmark for evaluation and analysis. [12] provides a VQA dataset containing over 250K images, 760K questions, and around 10M answers. They use real and abstract images to formulate the questions which require common-sense as well as visual understanding to answer the questions. [79] extended the dataset released by [12] to counter language priors for the task of VQA. Specifically, they balanced the VQA dataset by collecting complementary images such that every question in the balanced dataset is associated with not just a single image, but rather a pair of similar images that result in two different answers to the question.

[119] proposed a diagnostic dataset (named CLEVR) to assess the ability of VQA systems to perform visual reasoning. The dataset contains images and questions that test the various reasoning abilities (counting, comparing, logical reasoning) of a system. CLEVR provides detailed annotations which facilitate in-depth analyses of reasoning abilities that were not possible with the other VQA datasets. [99] introduced a VQA dataset named "VizWiz". This dataset contains the questions asked by blind people who were seeking answers to their daily visual questions. This dataset also promotes the problem of predicting whether a visual question is answerable. *Vizwiz* dataset is constructed by explicitly asking annotators whether a visual question is answerable when collecting answers to our visual questions. [142] released the Visual Genome dataset that facilitated a multi-layered understanding of an image. This dataset allows a system to understand the multiple perspectives from pixel-level information, like objects, to relationships that require further inference. The Visual Genome dataset provides the annotations of objects, attributes, and relationships between them. Specifically, this dataset contains over 100K images, where each image has an average of 21 objects, 18 attributes, and 18 pairwise relationships between the objects.

---

[8]http://test.baidu.com/

Apart from these datasets, several other datasets [221, 291, 99, 248, 101, 292] have been released to address the particular problem related to VQA and to challenge the VQA system by introducing the various level of difficulties. In the majority of these datasets, the images were taken from the large-scale image database MS COCO or artificially constructed [12, 10, 119]. The questions in the dataset were either formulated automatically [119, 71, 142] or by the crowd workers [12, 79, 221, 99]. There are a few datasets [71, 248] for multilingual VQA, but these are limited only to some chosen languages, and unlike our dataset, they do not offer any code-mixed challenges.

### 2.4.2 VQA Models

The popular frameworks for VQA in the literature are built to learn the joint representation of image and question using the attention mechanism.

[69] proposed a VQA system by introducing the bi-linear model. They proposed Multi-modal Compact Bilinear pooling (MCB), which uses the outer product of two feature vectors to produce a very high-dimensional feature for quadratic expansion. To reduce the computational cost, they used a sampling-based approximation approach that exploits the property that the projection of two vectors can be represented as their convolution. However, MCB needs high-dimensional features to get robust performance.

[132] overcome this problem by proposing Multi-modal Low-rank Bilinear Pooling (MLB). The technique is based on the Hadamard product of two feature vectors (image and question) in the common space with two low-rank projection matrices. However, MLB has a slow convergence rate and is sensitive to the learned hyper-parameters. [334] generalized the multi-modal fusion technique proposed by [132] and proposed Multi-modal Factorized Bilinear Pooling (MFB) which follows the two-stage process to fuse the information. The first stage is *expansion*, where the features from different modalities are expanded to a high-dimensional space and then integrated with element-wise multiplication. In the second stage (*squeeze*) sum pooling followed by normalization layers is performed to squeeze the high-dimensional feature into the compact output feature.

[130] propose bilinear attention networks (BAN) by extending the idea of co-attention into bilinear attention, which considers every pair of multimodal channels. BAN exploits bilinear interactions between the two groups of input channels, while low-rank bilinear pooling extracts the joint representations for each pair of channels. [168] proposed a

mechanism (co-attention) that jointly reasons about visual and question attention. They also exploited the question hierarchy to encode the question better. The hierarchical architecture co-attends to the image and question at three different levels (word level, phrase level, and question level). [112] proposed a technique to separately learn the answer embedding with best parameters such that the correct answer has a higher likelihood among all the possible answers.

There are also some works [33, 165, 217, 305] which exploited the adversarial learning strategy in VQA. These learned representations are passed to a multi-label classifier whose labels are the most frequent answers in the dataset. In addition to these, there are other VQA models [334, 128, 252, 253, 244, 17], which focused on attention mechanism to fuse the multiple modalities.

With the success of Transformer based pre-trained networks [52, 263, 55, 123] in NLP, there is a growing interest in solving a variety of computer vision and natural language processing problems, such as image captioning, visual question answering, image retrieval, etc. Towards this, [276] proposed a cross-modality framework, LXMERT, for learning the connections between vision and language. The LXMERT model is based on Transformer and cross-modality encoders. [129] adapt the LXMERRT model with knowledge-distillation for multilingual and code-mixed visual question answering. Similar to BERT, it is pre-trained with diverse pre-training tasks on the large-scale dataset of image and text pairs. [152] introduced a pre-trained model, VISUALBERT for joint language and vision representation. To learn the associations between the language and text, VISUALBERT is pre-trained with two visually-grounded language model tasks on COCO image caption data [37]. [270] proposed a VL-BERT pre-trained model, which is based on single cross-modal architecture for language-vision tasks. The network is pre-trained with the masked language model objective and masked the visual feature classification. They utilized the Conceptual Captions [245], BookCorpus [352], and English Wikipedia to perform the pre-training tasks. There are other notable works [346, 38, 167, 151, 156], where Transformer based models are pre-trained to learn the joint language-vision representation and later used for the downstream tasks, such as VQA, image captioning, etc.

It is to be noted that designing a VQA system for each language is quite expensive in terms of computation and time, where multiple languages are involved. Hence, a single VQA model that integrates multilinguality and code-mixing in its components is

**25**

extremely useful.

## 2.5 Semantic Question Matching

Question-question (QQ) matching is one of the very important problems in QA setup. The rapid growth of community question and answer (CQA) forums have intensified the necessity for QQ matching. Answer retrieval to similar questions has drawn the attention of researchers in recent times [176, 190]. It solves the problem of *question starvation* in CQA forums. There have been attempts in the literature to find the most similar match for a given question, which has already been answered [28, 182]. Along this direction, [289] has presented a syntactic tree-matching method for finding semantically similar questions. Similar question retrieval has been modeled using various techniques, such as topic modeling [154], knowledge graph representation [345], and machine translation [118]. Similarly, [149] has proposed a recurrent network over non-consecutive convolution to represent questions in mathematical vector space and use cosine similarity to retrieve similar questions.

In contrast to [149], our proposed approach uses taxonomy information along with the deeper semantic representation of questions to find similar questions. The system proposed in [57] uses the combination of *bag-of-words* (BoW) representation and Convolutional Neural Network (CNN) to retrieve similar questions. The technique proposed in [344] showed a way to learn word embeddings using category-based metadata information for questions. Popular approaches where taxonomy is used for question answering have been proposed in [155, 115]. These existing taxonomies are more suited for factoid question answering, primarily focusing on interrogative questions. Semantic kernel-based similarity methods for QA have also been proposed in [66, 46, 47]. [49] proposed an approach called SCQA design to find semantic similarity between the two questions. The approach is based on the architecture of the Siamese Convolutional Neural Network. The proposed network consists of two convolutional neural networks [314] with shared parameters and a loss function (contrastive) joining them. The aim of the proposed model to project the semantically similar questions close to each other and dissimilar questions far from each other in the semantic space.

Answer selection on QA forums is similar to question similarity task. In recent times, neural network-based model architecture was investigated, but mainly for answer selection in [287, 243, 64, 277]. Most of the existing works either focus on better representation of questions or linguistic information associated with the questions. On the other hand, the model proposed in this dissertation is a hybrid model, where we present an empirical study of how sophisticated deep learning models can be used along with a linguistically motivated taxonomy for QQ matching. Recently deep neural variational inference [177] has been present for answer sentence selection. [126] proposed a CNN-based model for answer sentence selection and released the SelQA dataset.

Generalization is a long-standing problem in a deep learning framework because sentences with different surface forms can convey the same meaning (paraphrases), and not all of them can be listed in the training set. Towards this, [342] introduced the NLP-Capsule Framework, based on the extension of the Capsule network [108, 228]. The proposed NLP-Capsule Framework is capable of scaling to large output spaces and higher reliability for routing processes at the instance level. The experiment on the TREC QA dataset [290] showed promising results and outperformed the competitive deep neural network-based systems on the answer sentence selection problem.

## 2.6 Multi-hop Question Generation

Previous works on question generation can be categorized into two types: rule-based methods and neural network-based methods. Traditional rule-based approaches involve the manual formulation of templates to generate questions and further refinement of the generated questions using features based on semantic information [162, 104], ontologies [145], argument structures [30] etc.

Our literature survey shows that the existing methods of question generation (general) include both rules [104, 7] and machine learning [241, 295, 93] techniques. The very first work was conducted by [104]. They developed a rule-based technique for question generation (QG). Another rule-based approach was proposed by [7], where they extracted elementary sentences from complex sentences using syntactic information and generated questions based on the subject, verb, object, and preposition using predefined interaction rules.

Recently, works on question generation have drifted towards neural-based approaches. These approaches typically involve end-to-end supervised learning to generate questions. [59] proposed the first neural question generation (NQG) model. However, in this work, the answer is not considered while generating the question. This leads to the generation of incoherent questions. [347] utilized the answer-position, and linguistic features, such as named entity recognition (NER) and parts-of-speech (POS) information, to further improve the QG performance as the model is aware that for which answer a question needs to be generated. In the work of [297], a multi-perspective context-matching algorithm was employed. Another study, conducted by [100], used a set of rich linguistic features along with an NQG model. [264] used the matching algorithm proposed by [297] to compute the similarity between the target answer and the passage for collecting relevant contextual information under the different perspectives, so that contextual information can be better considered by the encoder. More recently, [133] has claimed to improve the performance of the QG model by replacing the target answer in the original passage with a special token.

The model proposed by [240, 298] for single-document QA experienced a significant drop in accuracy when applied in multiple document settings. This shortcoming of the single-document QA dataset is addressed by newly released multi-hop datasets [299, 275, 330] that promote multi-step inference across several documents. So far, multi-hop datasets have been predominantly used for answer generation tasks [239, 279, 341]. Our work can be seen as an extension to single-hop question generation where a non-trivial number of supporting facts are spread across multiple documents. Our current work is different from the existing question generation works because our framework can generate the question by reasoning over multiple documents. QA systems' ability to perform multi-hop reasoning. There are some works [239, 279, 341] on multi-hop QA, but none of the works focus on question generation in multi-hop settings.

## 2.7 Code-Mixed Text Generation

The problem of data scarcity is one of the focuses of research in code-mixing. To solve this problem, some synthetic data generation techniques have been proposed in the literature. However, modeling code-mixed phenomenon in language pairs is not a trivial task, especially with limited supervision in terms of data. The challenges increase manifold while dealing with multiple language pairs. [2] extended the standard Recurrent Neural Network (RNN) [315, 250] by inducing POS tags into the input layer and language information to factorize the output layer. Followed by that, [3] paired up the RNN model with the n-grams backup model with linear interpolation and further augmented the syntactic and semantic features [1].

In another study, [42] exploited curriculum learning strategies to train the network on the code-mixed data efficiently. Firstly, they trained the network on the monolingual corpora of the two languages and then trained the resulting network on code-switched data. In the same line, [77] proposed a ranking-based evaluation method for the language models motivated by speech recognition and created an evaluation dataset for English-Spanish code-switching LM (SpanglishEval). They further presented discriminative training for this ranking task that is intended for ASR purposes. [210] explored the equivalence constraint (EC) theory to construct grammatically correct artificial code-mixed data. They analyzed that when training instances are correctly sampled from the synthetic data via training curriculum, together with code-mixed and monolingual data can help in reducing the perplexity of the RNN-based language model. [303] further validated the usefulness of EC and curriculum learning in the English-Spanish language modeling.

To address the critical issue of data scarcity in code-mixed settings, [303] proposed a multitask learning framework. Particularly, they leveraged the linguistic information of languages using common shared syntax representation jointly learned over POS and language modeling on code-switched utterances. [73] exploited SeqGAN in the generation of the synthetic code-mixed language sequence. Most recently, [302] utilized the language-agnostic meta-representation method to represent the code-mixed sentence.

The very first study was conducted by [158], where they presented the statistical approach. Their approach was linguistically driven by adapting the EC to monolingual sentence pairs during the decoding stage of an automatic speech recognition (ASR) method. To minimize search space for a code-mixed ASR method, [159] introduced a functional-head constraint lattice parser and weighted finite-state transducer. In a similar line work, [157] proposed a generative model for category text generation to enlarge the training dataset. They utilized the generative adversarial network (GAN) [78] with a recurrent neural network to build the text generation framework termed CS-GAN, which takes the category (label) and sentence distribution (latent variable) to generate the synthetic text.

In contrast to the previous works, in this dissertation, first, we provide a linguistically motivated technique to create code-mixed datasets from multiple languages with the help of corresponding language pair parallel sentences. Thereafter, we utilize the former datasets to propose a neural-based approach to generate the code-mixed sentence from the English sentence. The work presented in this dissertation has a wider scope as the underlying architecture can be used to harvest the code-mixed data for various NLP tasks not only limited to the language modeling and speech recognition as it has been focused on in the earlier works. In contrast to the previous studies, where only a few of the language pairs were considered for code-mixing, we propose an effective approach that shows its effectiveness in generating code-mixed sentences for eight different language pairs of different kinds and origins.

## 2.8 Evaluation

In this section, we provide the details of various evaluation metrics used in the thesis work. We use the following metrics [216] to evaluate the performance of a deep neural-based question-answering system.

- **Exact Match (EM):** It computes the percentage of system-predicted answers that match any ground truth answers.
- **F1 Score:** It measures the overlap between the system-predicted answer and ground truth answer. The predicted and ground truth answers are considered as a bag of tokens, thereafter the F1 score is computed between them. We compute the F1 score for each question in the dataset and take the average of them to compute the F1 score for the dataset.

To compute the performance of IR based question answering system and semantic question ranking, we compute the following metrics:

- **Precision at k (P@k):** Precision at $k$ is the proportion of system-predicted items in the top-k set that are correct.

$$P@k = \frac{\text{\# of correctly predicted items in the top-k}}{\text{\# of predicted items}} \tag{2.1}$$

- **Recall at k (R@k):** Recall at $k$ is the proportion of correctly predicted items by the system in the top-k predictions.

$$R@k = \frac{\text{\# of correctly predicted items in the top-k}}{\text{\# of correct items}} \tag{2.2}$$

- **Mean Average Precision (MAP):** MAP is the average precision averaged across a set of queries.

$$MAP = \frac{1}{|Q|} \sum_{j=1}^{|Q|} \frac{1}{m_j} \sum_{k=1}^{m_j} P@k_j \qquad (2.3)$$

where, $Q$ is the total set of queries, and $m_j$ is the number of correct items to be returned for the $j^{th}$ query.

- **Mean Reciprocal Rank (MRR):** The mean reciprocal rank is the average of the reciprocal ranks of results for a sample of queries.

$$MRR = \frac{1}{|Q|} \sum_{j=1}^{|Q|} \frac{1}{rank_j} \qquad (2.4)$$

where $rank_j$ refers to the rank position of the first correct item for the $j$-th query.

In order to compute the performance of the language generation system, we use the following metrics:

- **BLEU:** It stands for BiLingual Evaluation Understudy (BLEU) [200]. It is a popular evaluation metric in language generation which compares the generated sentence with the reference sentence based on the number of n-grams of the generated sentence that match with the reference sentence, along with the brevity penalty for shorter output. BLEU is computed using the multiple modified n-gram precisions[9]. Specifically,

$$\text{BLEU} = \text{BP} \cdot \exp\left( \sum_{n=1}^{N} w_n \log_e p_n \right) \qquad (2.5)$$

where $p_n$ is the modified n-gram precision, BP is the brevity penalty to penalize short answer, $w_n$ is the weight between 0 and 1 for $\log_e p_n$ and $\sum_{n=1}^{N} w_n = 1$, $N$ is the maximum length of n-gram. BP can be computed as follows:

$$\text{BP} = \begin{cases} 1 & \text{if } c > r \\ \exp\left(1 - \frac{r}{c}\right) & \text{if } c \leq r \end{cases} \qquad (2.6)$$

where $c$ is the number of unigrams in all the candidate sentences $r$ is the best match

---

[9]To compute modified n-gram precision, all candidate n-gram counts and their corresponding maximum reference counts are collected. The candidate counts are clipped by their corresponding reference maximum value, summed, and divided by the total number of candidate n-grams.

lengths for each candidate sentence in the dataset.

- **METEOR:** METEOR [14] stands for Metric for Evaluation for Translation with Explicit Ordering. It is another evaluation metric to assess the performance of language generation systems such as machine translation, summarization and sentence generation, etc. Given a pair of system-generated sentences and a reference sentence, METEOR creates an alignment between them. In the alignment, each uni-gram of system generated sentence is mapped to another uni-gram in the reference sentence. The matching between uni-gram is done on the basis of word-mapping modules:"Exact match", "Porter-Stem match" and "WordNet Synonyms match". The word-mapping modules generate all word matches between the pair of strings. Thereafter, the largest subset of these word mappings is considered the final alignment. Based on the number of mapped uni-grams between the two strings ($m$), the total number of uni-grams in system generated sentence ($t$), and the total number of uni-grams in the reference ($r$), uni-gram precision $P = m/t$ and uni-gram recall $R = m/r$ are computed. A parametrized harmonic mean [225] of $P$ and $R$ are computed as $F_{mean} = \frac{10PR}{9P+R}$. METEOR also computes the penalty to account for the word order in the reference sentence and candidate sentence.

- **ROUGE-L:** ROUGE [160] stands for Recall-Oriented Understudy for Gisting Evaluation. It measures the count of the number of overlapping n-grams between the system-generated sentence and the reference sentence. In this work, we use ROUGE-L a variant of ROUGE to compute the performance of the system. **L** in ROUGE-L refers to the Longest Common Subsequence (LCS) based statistics. The longest common subsequence problem takes into account sentence-level structure similarity naturally and identifies the longest co-occurring in sequence n-grams automatically.

## 2.9 Conclusion

The first part of this chapter describes the existing literature on several tasks related to QA, i.e., code-mixing, multilingual question answering, semantic question matching, question generation, and code-mixed text generation. In the later part of this chapter, we discuss the evaluation scheme for the various problems, as discussed in Chapter 1.

The next chapter is the first contributory chapter, where we discuss in detail our proposed framework for solving the multilingual question-answering task in English and Hindi languages. We will begin by describing the annotation scheme to create a high-quality benchmark multilingual QA dataset. Later, we will elaborate on our proposed methods based on IR and neural network techniques for efficiently modeling the QA in a multilingual setting.

# Chapter 3

# Multilingual Question Answering

## 3.1 Introduction

This is the first contributing chapter, where we focus on multilingual question answering, particularly in English-Hindi languages. In the first part of this chapter, we discuss the dataset creation process for the English-Hindi question-answering task followed by the analysis of the developed dataset, and finally, the IR-based framework, utilizing the machine translation for non-English questions and answers. The second part of this chapter is devoted to the advance the IR method. Towards this, we proposed a novel neural network approach for solving the multilingual question-answering problem.

With the abundance of digital information on the web, the need to access precise information has increased tremendously during the past few years. However, the information on the web is not limited to any particular language. There are vast sources of information available in multiple languages, making the web a pool of multilingual information sources. A multilingual question-answering system can extract the precise answer(s) to a given question from various sources of information, regardless of the language of the question or the information sources. Such a system allows users to interact and receive query-specific information from various multilingual information sources, which may not be available in their native languages.

A system, which can process the questions asked in multiple languages and comprehend to retrieve the answer from multilingual documents, is termed an MQA system. It is an interesting yet challenging research area in QA. Such a system allows users to interact in their native languages, facilitating multilingual information access, which is immensely useful in a country like India. MQA system can also contribute to conserving the endangered languages, which are losing their existence and prestige as mentioned in [138]. Hindi is a widely spoken language in India, and in terms of native speakers, it ranks fourth all over the world. According to 2001 Census[1], 57.10% of the total population speaks Hindi compared to English (10.60%). English, on the other hand, is used for all kinds of official communications. There is often a need to exchange information from Hindi with other popular languages such as English. The primary goal of our MQA framework is to set up a common system to evaluate both bilingual and cross-lingual question answering that process queries in either Hindi or English language and retrieve answers in either language from documents in Hindi or English.

Inspired by this, we propose a unified deep neural network framework to retrieve the multilingual answer by exploring the attention-based recurrent neural network to generate an adequate representation of multilingual questions and snippets. We utilize the soft-alignment of words from English and Hindi questions to generate a single shared representation of questions. The proposed system's effectiveness is demonstrated by extracting the answer to an English and/or Hindi question from English and/or Hindi snippets. Our experiments on a recently released multilingual QA dataset show that our proposed model achieves state-of-the-art performance. For multilingual settings, our model has shown significant performance improvement over the baselines.

### 3.1.1 Problem Statement

Given a natural language question, $\mathcal{Q}$ in English or Hindi and multilingual documents $D$, the task is to extract the answer $A$ from the multilingual documents $D$. Let us consider the following example from Table 3.1:
**Question:** शिमला का क्षेत्रफल कितना है?
(**Translation:** *What is the area of Shimla?*).
Even though the answer to this question is not available in the Hindi passage, it can be retrieved (**25 sq km**) from the English passage. An efficient MQA system provides the facility to retrieve the answers across multilingual information sources.

---

[1] https://en.wikipedia.org/wiki/2011_Census_of_India

**36**

| |
|---|
| **English Passage:** Shimla is the capital of Himachal Pradesh and was also the summer capital in pre-independence India. Covering an area of 25 sq km at a height of 7,238 ft Shimla is surrounded by pine, deodar and oak forests. <br> **Hindi Passage:** शिमला, एक ख़ूबसूरत हिल स्टेशन है जो हिमाचल प्रदेश की राजधानी है। <br> (**Trans:** Shimla is a beautiful hill station, which is the capital of Himachal Pradesh.) |
| **Question:** हिमाचल प्रदेश की राजधानी क्या है? <br> (**Trans:** What is the capital of Himachal Pradesh?) <br> **Answer(s)**: [Shimla, शिमला (**Trans:** Shimla)] <br> **Question:** What is the capital of Himachal Pradesh? <br> **Answer(s):** [Shimla, शिमला (**Trans:** Shimla)] |
| **Question:** शिमला का क्षेत्रफल कितना है? <br> (**Trans:** How much area is covered by Shimla?) <br> **Answer(s):** [25 sq km] <br> **Question:** What is the height of Shimla from sea level? <br> **Answer(s):** [7,238 ft] |

Table 3.1: Sample multilingual questions, answers and passages.

### 3.1.2 Motivation

- The current QA system is centered around a monolingual setup, which does not account for the documents having multiple languages. Besides, transliterated text from non-roman alphabets could be included in both questions and documents, which leads to high spelling variations. Although language identification for monolingual text is a well-established research area, for multilingual documents, it is still an unexplored research area.

- The ability to handle Indic language queries/questions is limited in the current search engine. We have shown the comparison of search engine response to the English and Hindi query in Fig 3.1. Here, a search engine can provide the exact answer to the question:

  **Question:** *"What is the population of Sikhs in Canada?"*.

  However, when a query is made in Hindi language, search engines fail to provide the exact answer to the question:

  **Question:** कनाडा में सिखों की जनसंख्या कितनी है ?"

  (***Translation:****"What is the population of Sikhs in Canada?"*).

  It is to be noted that although both questions are the same, the search engine is unable to return the exact answer for the Hindi question. Instead, it returns a web snippet.

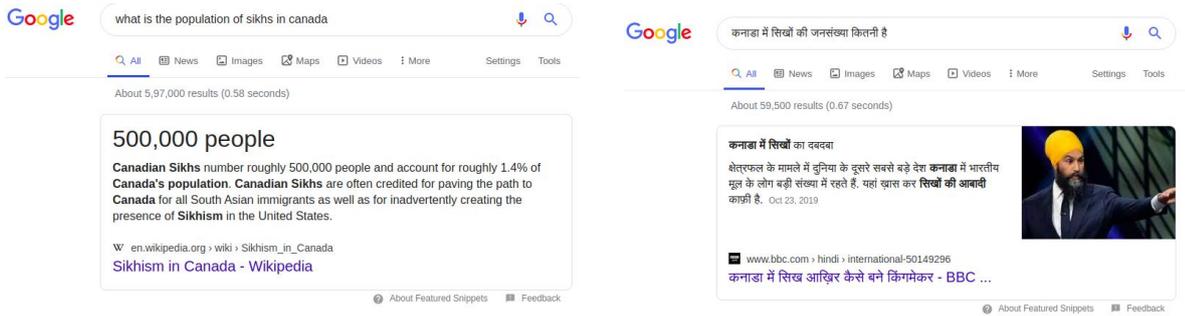

Figure 3.1: Comparison of search engine response to the English query (left) and Hindi query (right).

- Traditional approaches deal with multilingual QA by translating either questions or documents and converting the problem to a more familiar monolingual QA setup. However, one of the critical issues with the traditional approach is its dependency on the quality of the underlying machine translation system. Although the recent advancement of neural machine translation has shown promising performance on resource-rich language pairs such as English-French, it is still a long way to go for Indian languages.
- Most of the existing research works are in resource-rich languages such as English and French. For building the multilingual QA system focused on resource-scarce Indian languages (such as Hindi), the transfer learning method can be adapted to utilize the resources and tools developed on the English language

### 3.1.3 Challenges

- **Linguistic Diversities:** The linguistic diversities (e.g., morphological, lexical, syntactical) across the languages of a question, document, and answer add to the challenge in developing a robust MQA system.
- **Lack of Multilingual NLP Tools:** There are currently very limited language-dependent tools (such as POS taggers and NER recognizers) in a multilingual domain. These tools are often required in MQA to perform the upstream task of information retrieval. However, developing such tools is expensive and requires plenty of human effort.
- **Data Scarcity:** Recently, there has been much progress [216, 281, 330] towards creating the QA/RC dataset for the English language. However, there is a scarcity of large-scale multilingual QA datasets, which are often required for training gigantic Transformers [284] and deep learning networks to achieve similar performance as observed in the English language. Training such systems requires a huge set of examples from diverse languages which is very tedious to create manually.

- **Quality of the Data:** We can utilize the recent advancement of NMT and NQG to create questions or documents in multiple languages. However, this can leads to a quality problem. It may arise the following issues in the generated data: (a) **Answerability**: the semantics of the translated passage or generated question may shift, which may lead to a situation where the question gets unanswerable after generation, or (b) **Misalignment**: the answer span in a target language may become incorrect. This issue is often observed in the case of multilingual reading comprehension.

## 3.2 Resource Creation

There is no existing benchmark setup for a multilingual QA system in Indian languages. Building a multilingual QA system will be highly beneficial for multilingual Indian communities to facilitate information needs in their native languages. Towards this, we created a QA dataset, named 'MMQA' (Multi-domain Multilingual Question Answering) in Hindi and English languages covering multiple domains. We only focus on creating *factoid* and *short descriptive* questions. The MMQA dataset is created in three different stages: *Comparable Article Curation*, *Question Answer Formulation*, and *Validation*.

| Domains | English/Hindi | | | |
|---|---|---|---|---|
| | # Articles | # Paragraphs | # Sentences | # Words |
| **Tourism** | 112/112 | 1,569/1,186 | 5,077/3,799 | 90,222/71,863 |
| **History** | 68/68 | 563/597 | 2,518/2,224 | 40,368/46,418 |
| **Diseases** | 31/31 | 441/298 | 1,932/1,171 | 33,787/28,128 |
| **Geography** | 16/16 | 81/171 | 304/520 | 8,915/9,443 |
| **Economics** | 13/13 | 146/144 | 667/477 | 10,633/10,875 |
| **Environment** | 10/10 | 64/54 | 290/272 | 5,319/6,109 |
| Total (EN/HI) | 250/250 | 2,864/2,450 | 10,788/8,463 | 189,244/172,836 |
| Total (EN+HI) | 500 | 5,314 | 19,251 | 362,080 |

Table 3.2: Statistics of comparable English and Hindi articles from various domains.

### 3.2.1 Comparable Article Curation

Since our objective is to build a multilingual QA dataset, therefore we curated comparable articles from various web sources covering different domains. We provide the statistics of the comparable articles in Table 3.2. Rather than translating an article from one language to the other, we curated comparable articles, mainly for two reasons: **(i)** to bridge the information gap between two different language articles, and **(ii)** to assess the challenges of dealing with two syntactically divergent texts to retrieve answers for the given questions. From each of these articles, we extracted individual paragraphs and removed images, links, and tables. We curated a total of 500 articles from 6 different domains. We curated these texts from the different web sources using a web crawler[2].

| Domains | QA pair in only English (Fact/Desc) | QA pair in only Hindi (Fact/Desc) | QA pair in Both Languages (Fact/Desc) | Total QA pair (Fact/Desc) | Total QA pair |
|---|---|---|---|---|---|
| Tourism | 456/14 | 403/5 | 422/10 | 1,281/29 | 1,310 |
| History | 110/75 | 126/78 | 1,118/588 | 1,354/741 | 2,095 |
| Diseases | 81/54 | 33/26 | 48/40 | 162/120 | 282 |
| Geography | 55/7 | 29/10 | 174/202 | 258/219 | 477 |
| Economics | 25/4 | 14/5 | 682/218 | 721/227 | 948 |
| Environment | 9/3 | 2/1 | 226/142 | 237/146 | 383 |
| Total | 736/157 | 607/125 | 2,670/1,200 | 4,013/1,482 | 5,495 |

Table 3.3: Statistics of QA pairs for factoid and short descriptive questions in English and Hindi.

### 3.2.2 Question and Answer Formulation

We engage annotators[3] to formulate the question-answer pairs in their own words. The annotators are provided with an interface displaying the domain name, article name, and comparable article in parallel. They were asked to formulate questions in English and Hindi by looking into both comparable articles. When they formulate a question by looking at a paragraph in one (English or Hindi) article, they also have to verify whether the question of interest is available in the comparable article or not. Specifically, they have to provide the *question, answer, answer source (sentence or paragraph where the answer exists), and type of question (factoid, descriptive)* in both languages, if it exists. Additionally, annotators were encouraged to paraphrase questions in their own words.

---

[2]Tourism (EN):www.india.com/travel
Tourism (HI): https://hindi.nativeplanet.com Rest of the domains are curated from http://www.jagranjosh.com/

[3]The annotators are equally proficient in both languages

Statistics[4] of question and answers in both the languages are shown in Table 3.3. An example of QA pairs formulated from a comparable article is given in Table 3.4.

| Article (English) | Article (Hindi) |
|---|---|
| Shimla is the capital of Himachal Pradesh and was also the summer capital in pre-Independence India. Shimla derives its name from Shyamala Devi, an incarnation of the goddess Kali, whose temple existed in the dense forest covering the Pradesh Jakhu Hill in the early 19th century. Shimla is the capital of Himachal in pre-Independence India. Covering an area of 25 sq km at a height of 7,238 ft, Shimla is surrounded by pine, deodar and oak forests. | शिमला, एक खूबसूरत हिल स्टेशन है जो हिमाचल प्रदेश की राजधानी है। समुद्र की सतह से 2202 मीटर की ऊँचाई पर स्थित इस जगह को 'समर रिफ्यूज' और 'हिल स्टेशनों की रानी' के रूप में भी जाना जाता है। वर्तमान का शिमला जिला 1972 में निर्मित किया गया था। इस जगह का यह नाम 'माँ काली' के दूसरे नाम 'श्यामला' से व्युत्पन्न है। जाखू, प्रॉस्पेक्ट, ऑब्सर्वेटरी, एलीसियम और समर इस जगह की महत्वपूर्ण पहाड़ियाँ हैं। |

| Question (English) | Answer (English) | Question (Hindi) | Answer (Hindi) | Answer Availability |
|---|---|---|---|---|
| What is the capital of Himachal Pradesh? | Shimla | हिमाचल प्रदेश की राजधानी क्या है? | शिमला (*Shimla*) | EN and HI |
| How much area is covered by Shimla? | 25 sq km | शिमला का क्षेत्रफल कितना है? | - | EN |
| From which goddess name the Shimla word derived? | Shyamala Devi | शिमला शब्द किस देवी के नाम से लिया गया है? | श्यामला (*Shyamala*) | EN and HI |
| What is the height of Shimla from sea level? | 7,238 ft | समुद्र के स्तर से शिमला की ऊंचाई क्या है? | 2202 मीटर (*2202 meter*) | EN and HI |
| In which year Shimla established? | - | शिमला की स्थापना किस साल हुई? | 1972 | HI |

Table 3.4: An example of comparable articles from English and Hindi and a set of question-answer pairs created from the given articles.

### 3.2.3 Validation

The validation stage is performed to ensure that we obtain a high-quality dataset at the end. We asked two other annotators to verify the questions and answers generated in both languages. Annotators were given a free hand to correct the answers to some extent or eliminate the question-answer pairs if found not fitting. The validation stage is applicable for the question-answer pair of both languages.

### 3.2.4 Analysis

We analyze the questions and answers of the MMQA dataset. It is essential to understand the characteristics and usefulness of the created multilingual dataset. Our analysis focuses on studying the difficulty level of questions and the diversity of answers. We provide some examples in Table 3.5 to give some ideas about the difficulty levels associated with the questions. For better understanding and thorough analysis of various answer types, we categorize the answers of factoid questions into 8 entities and phrases similar to [216] and [281]. Statistics of the answer types for English and Hindi QA pairs are provided in Table 3.6.

Some examples of short descriptive QA pairs from our dataset are given in Table 3.7.

The Cross-Language Evaluation Forum (CLEF[5]) has introduced different multilinguality challenges for question answering in their respective editions from 2003 to 2009. The direct comparison of our dataset with the CLEF datasets [198] is not possible because we have created question answers pairs in both languages (MQA). In contrast, the CLEF dataset has the question and answer pair in different languages. However, we have shown the comparison in various terms as shown in Table 3.8.

---

[4]Total of 7120 questions (English+Hindi) for which the answer exists in either of two language documents.

[5]http://www.clef-initiative.eu/track/qaclef

| Reasoning Types | Question & Answer Sentence | Descriptions |
|---|---|---|
| Interlingual semantic word | **Q:** What was introduced in the Sixth Five-Year Plan? **Answer Sentence:** छठी पंचवर्षीय योजना (1980-1985) ने भी आर्थिक उदारीकरण की शुरुआत को चिन्हित किया । (*Chhati panchvarshiya yojna (1980-1985) ne bhi arthik udarikaran ki shuruaat ko chinhit kiya.*) | An interlingual semantic word knowledge is required to provide the answer. Only translation may not help to extract the answer. |
| Multiple sentence reasoning | **Q:** What is the part of the Adam's Bridge? **Answer Sentence: Pamban Island** is situated in the Gulf of Mannar between India and Srilanka. It is a part of the Adam's Bridge. | There is an anaphora, or fusion of multiple sentences is required to answer. |
| Word matching | **Q:** What is the collection of Vedic hymns or mantras called? **Answer Sentence:** The collection of Vedic hymns or mantras is called the **Samhita**. | Word matching between question and answer sentence can provide the answer. |
| Interlingual syntax variation | **Q:** प्लेग किसके लिए जिम्मेदार है ? (*Plaque kiske liye jimmedar hai?*) **Answer Sentence:** Plaque deposited on the teeth and under the gumline irritates the gum tissue, and causes **gingivitis**. | Syntactic structure of question and answer sentence vary across the language. |
| Single Sentence | **Q:** What is Sustainable development? **Answer Sentence:** Scale defines the relationship between distance on a map and on the earth's surface. Sustainable development: **Development that does not exploit resources more rapidly than the renewal of those resources**, ... | Answer of short descriptive question can be a single sentence from a paragraph. |
| Multiple Sentence | **Q:** Why India is considered to be an eastern country? **Answer Sentence:** ... **India lies east of the Prime Meridian. Therefore India is considered to be an eastern country because of its situation in the Eastern Hemisphere**.... | Answer of short descriptive question can be multiple sentences from a single or multiple paragraphs. |

Table 3.5: The set of possible reasoning types with the corresponding question-answer pair example and descriptions. Reasoning types show the difficulty of the question in terms of finding their answer. The answer in the answer sentence has been shown in bold font.

| Answer type | Proportion (English/Hindi) | Examples (English/Hindi) |
|---|---|---|
| **Person** | 12.22 / 14.28 | Krishna / तुलसीदास (*Tulasidas*) |
| **Location** | 17.26 / 14.89 | Madurai / भुवनेश्वर (*bhubaneswar*) |
| **Organization** | 7.69 / 8.96 | International Monetary Fund/ टाटा मूलभूत अनुसंधान संस्थान (*Tata Mulbhut Anusandhan Sansthan*) |
| **Noun Phrase** | 23.79 / 24.57 | Hotel Apsara / स्टील प्लांट (*Steel plant*) |
| **Verb Phrase** | 2.58 / 1.57 | planned economic development/ तनाव से छुटकारा (*Tanav se chhutkara*) |
| **Adjective Pharse** | 1.98 / 1.02 | Smiling Buddha/ 14 प्रमुख भारतीय बैंक (*14 pramukh Bhartiya bank*) |
| **Date / Numbers** | 32.43 / 33.17 | 580 / 7 किलो (*7 kilo*) |
| **Other** | 2.05 / 1.54 | at least two / सापूतरा का मतलब है 'नागों का वास'। (*Saputra ka matlab hai 'Nagon ka vas* ) |

Table 3.6: Set of various answer type categories (only for factoid questions) from the dataset with their proportion (in %) for English and Hindi answer.

The created MMQA dataset can be scaled to multiple languages by introducing the appropriate comparable documents. One way of scaling is to extract the corresponding Wikipedia articles in multiple languages and use the existing sets of questions from MMQA to create cross-lingual QA datasets supporting multiple languages. Translation of the existing article from MMQA is another way to scale the dataset in a multilingual QA dataset aiding multiple languages. The MMQA dataset can be used to evaluate the monolingual, cross-lingual, and multilingual QA systems. It can also create a code-mixed QA dataset, as discussed in Chapter 4 of this dissertation.

| | | | | | | | | |
|---|---|---|---|---|---|---|---|---|
| **Question (English):** Why did Alexander marched back in 325 BC? | | | | | | | | |
| **Question (Hindi):** अलेक्जेंडर 325 ईसा पूर्व में क्यों चला गया? | | | | | | | | |
| **Answer (English):** After Alexander's last major victory in India as his forces refused to go any further. They were too tired to carry on with the Alexander's expedition and wanted to return home... | | | | | | | | |
| **Answer (Hindi):** हालांकि, यह जीत भारत में उसकी आखिरी बड़ी जीत साबित हुई क्योंकि उसकी सेना ने इसके बाद आगे जाने से इनकार कर दिया था... | | | | | | | | |
| **Question (English):** What does Buddhist texts such as Jatakas reveal? | | | | | | | | |
| **Question (Hindi):** बौद्ध ग्रंथों जैसे जतकस क्या बताते हैं? | | | | | | | | |
| **Answer (English):** Buddhist texts such as Jatakas reveal socio-economic conditions of Mauryan period while Buddhist chronicles Mahavamsa and Dipavamsa throws light on the role of Ashoka in spreading Buddhism to Sri Lanka. | | | | | | | | |
| **Answer (Hindi):** Not Available | | | | | | | | |

Table 3.7: Examples of short descriptive QA pairs from the dataset.

| | 2003 | 2004 | 2005 | 2006 | 2007 | 2008 | 2009 | **Our data** |
|---|---|---|---|---|---|---|---|---|
| Target lang. | 3 | 7 | 8 | 9 | 10 | 11 | 9 | 2 |
| Collection | News 1994 | + News 1995 | | | + Wikipedia Nov. 2006 | | JRC-Acquis | Web |
| No. of questions | 200 | | | | | | 500 | **7120** |
| Type of questions | 200 Factoid | + Temp. restrict +Defn | | -Type of question +List | + Linked question +Closed lists | | -Linked +Reason +Purpose +Procedure | Factoid Descriptive |
| Supporting info. | Document | | | | Snippet | | Paragraph | Document |

Table 3.8: Comparison of our dataset with the various released Cross-Language Evaluation Forum (CLEF) dataset

## 3.3 IR based Framework for MQA

We develop a translation-based approach for multilingual QA. As English is a resource-rich language, we translate Hindi questions and articles into English. Our proposed model comprises of *Knowledge Source Preparation*, *Question Processing*, *Passage Retrieval*, *Candidate Answer Extraction*, and *Answer Scoring and Ranking* components. We describe the details of each component in the following subsections.

### 3.3.1 Knowledge Source Preparation

In this step, the passage/article from which answers need to be extracted is indexed using the efficient indexing mechanism. We translate Hindi questions and articles into English by Google Translate[6]. Total English articles are indexed at the passage level using an inverted indexing mechanism. The inverted indexing facilitates the fast retrieval of relevant articles against the questions. We use the Lucene[7] implementation for inverted indexing.

### 3.3.2 Question Processing:

The question processing (QP) step is responsible for analyzing and understanding the questions posed to the QA system. This step comprises two sub-steps **(a)** Question Classification and **(b)** Query Formulation.

**Question Classification**

This is a vital component of any scoring-based answer extraction technique. In general, question classification categorizes a question at a coarser and finer level based on the answer type. For example, when considering the question, *"When did Mandi become a part of India?"*, we wish to classify this question as coarse class: *Numeric* and finer class: *date*, implying that only candidate answers that are '*dates*' need to be considered. With the recent developments for multiple text classification [325, 323, 312] using deep learning, neural network models have shown promise for QA. A deep neural network shows exceptional performance in other NLP problems. Inspired by the deep neural network's success, we adapt neural network architecture to develop our question classification model. Our question classification model is based on convolution and recurrent units. The model comprises *Question embedding layer, Convolution layer, Recurrent layer, Softmax classification layer*. Our question classification model is inspired by [134] and [307]. The input to the model is an English question. Now we describe each component of the model:

- **Question embedding layer:** It is responsible for obtaining the sequence of dense, real-valued vectors, $E = [v_1, v_2 \ldots v_T]$ of a given question having $T$ tokens. We keep the maximum size of token $T = 15$ in this layer. The distributed representation $v_i \in R^k$ is the $k$-dimensional word vector. The distributed representation $v$ is looked up into the word embedding matrix $W$. In our experiment, we have used the pre-trained word embedding [8] matrix by [180].

---

[6]https://translate.google.com
[7]https://lucene.apache.org/
[8]https://code.google.com/archive/p/word2vec/

- **Convolution layer:** This layer performs convolution operation, which extracts the useful set of n-gram features from the questions. The extracted n-gram features help the network to identify the question categories correctly. Similar to [307] and [134] we obtain convolution feature $c_t$ at given time $t$. Then we generate the feature vectors $C = [c_1, c_2 \ldots c_T]$. The convolution operations are performed with the filter size of 3, 4, and 5.
- **Recurrent layer:** The shallow convolutional network has the shortcoming of capturing the long-term dependencies. To overcome this, we introduce the recurrent layer on top of the convolution layer. This layer performs recurrent operations over the convolution output $c$ at a given time $t$. Similar to [307], we obtained the forward and backward hidden states at every step time $t$ using the gated recurrent unit (GRU). [307] have used the LSTM unit; however, we have employed GRU due to its less complex architecture compared to LSTM.
- **Softmax classification layer:** Finally, the fixed-dimensional vector $h$ is fed into the softmax classification layer to compute the predictive probabilities for all the question classes (coarse or fine).

**Query Formulation**

To form the query, we remove all the stop words, and punctuation symbols from the question. We tag the question with Stanford POS tagger [280]. Then we concatenate all the noun, verb, and adjective in the same order in which it appears in the question.

### 3.3.3 Passage Retrieval

The candidate passage that contains the answer(s) to the given question(s) is extracted in this stage. We exploit Lucene's text retrieval functionality to retrieve passages. It retrieves and ranks the passages using a combination of a Boolean model and the BM25 vector space model [337]. The query obtained from the *question processing* stage serve as input to the scorer module. The most relevant 30 passages were retrieved for subsequent processing.

### 3.3.4 Candidate Answer Extraction

This depends on the output of the question classification. For factoid questions, the coarse and finer class guide this stage to extract the appropriate entities (person, location, organization, number, etc.) from the candidate passage(s). We tag the candidate passage with Stanford named entity tagger [67]. We utilize the coarse class and finer class of a question to extract suitable candidate answers. For descriptive questions, candidate answers are extracted by segmenting the relevant passage.

### 3.3.5 Answer Scoring and Ranking

Each candidate answer is assigned a score using the candidate answer extraction phase. We segment the candidate passage into several candidate answer sentences. Thereafter, we calculate the score for each of the candidate answer sentences.

1. **Term Coverage Score (TCS):** It calculates the number of query terms appearing in the candidate answer sentence. This is normalized w.r.t the number of terms present in the given query.
2. **Proximity Score (PS):** It calculates the length of the shortest span that covers the query contained in the candidate answer sentence. This is again normalized in the same way.
3. **N-gram Coverage score (NCS):** We compute the n-gram coverage till $n = 4$. Finally, the $n$-gram score between a query (q) and a candidate answer sentence (S) is calculated based on the following formula.

$$NGCoverage(q, S, n) = \frac{\sum_{ng_n \in S} Count_{common}(ng_n)}{\sum_{ng_n \in q} Count_{query}(ng_n)} \quad (3.1)$$

$$NGScore(q, S) = \sum_{i=1}^{n} \frac{NGCoverage(q, S, i)}{\sum_{i=1}^{n} i} \quad (3.2)$$

4. **Semantic Similarity Score (SSS):** Query and candidate answers are represented using the semantic vectors. Cosine similarity is then computed between the query and candidate answers.

$$\text{VEC}(X) = \frac{\sum_{t_i \in X} \text{VEC}(t_i) \times \text{tf-idf}_{t_i}}{\textit{number of look-ups}} \quad (3.3)$$

where $X$ is query $q$ or candidate answer sentence $S$, $\text{VEC}(t_i)$ is the word vector of word $t_i$. *number of look-ups* represents the number of words in the question for which pre-trained word embeddings[9] are available.

5. **Pattern Matching Score (PMS):** This score is used in the descriptive question only. We design a set of patterns similar to the [120] to match a query against the candidate answers. We set up a score for each pattern according to its importance. For factoid and descriptive questions the weighted aggregate score for each candidate answer (A) is calculated as:

$$\begin{aligned} S_f(Q, A) &= W_1^f * TCS + W_2^f * PS + W_3^f * NCS + W_4^f * SSS \\ S_d(Q, A) &= W_1^d * TCS + W_2^d * PS + W_3^d * NCS + W_4^d * SSS + W_5^d * PMS \end{aligned} \quad (3.4)$$

Here, $W_k^f$ and $W_k^d$ are the learning weights for factoid and descriptive questions, respectively. Optimal values [10],[11] are determined through the validation data. For the factoid question, the candidate having the maximum score is returned as an answer to the given question. Answers to the descriptive questions may sometimes cover multiple sentences. At first, we consider the sentence having the maximum score and then include the other sentences which have scores closer to the highest one.

### 3.3.6 Experimental Result and Analysis

The experiments performed on the benchmark English-Hindi dataset can be categorized in two-fold: English question classification and answer extraction. For question classification, network training and hyper-parameters are discussed as follows:

**Network Training and Hyper-parameters:** We have applied the rectified linear units (ReLu) as the activation function in our experiment. We use the development data to fine-tune the hyper-parameters. In order to train the network, the stochastic gradient descent (SGD) over mini-batch is used and the backpropagation algorithm [103] is used to compute the gradients in each learning iteration. In order to prevent the model from over-fitting, we employed a dropout regularization (set to 50%) proposed by [268] on the penultimate layer of the network. We have used cross-entropy loss as the loss function.

---

[9] https://code.google.com/archive/p/word2vec/
[10] optimal weights for factoid (0.31, 0.18, 0.39, 0.12)
[11] optimal weight for descriptive (0.21, 0.09, 0.23, 0.19, 0.28)

The experimental results for question classification and answer extraction are described as follows:

- **Question Classification:** We perform the experiment on a coarse and fine class set of the questions using the model discussed in Section 3.3.2. For training, we use the following datasets:

  1. A dataset[12] of 5,452 questions collected from [111], TREC 8 and TREC 9 questions dataset,
  2. A dataset of 500 questions from TREC 10[13].
  3. We also manually label 1,022 questions at coarse and finer labels with the taxonomy guidelines provided by [155]. These questions were randomly taken from the set of curated questions.

  We perform 5-fold cross-validation to evaluate the question classification model. Our proposed model obtains the accuracy of 90.12% and 80.30% for question classification under coarse (i.e. 63 classes), respectively. This model is used to classify the incoming questions while we perform answer extraction.

- **Answer Extraction:** We perform experiments for the factoid and descriptive questions using the model proposed in Section 3.3. We use 10% of the total dataset of factoid and descriptive QA pairs, shown in Table 3.3, as the validation dataset to fine-tune the weight parameters. MRR and EM are used to evaluate the model performance on factoid questions. For descriptive questions, we use well-known machine translation evaluation metrics like BLEU [200] and ROUGE [160]. While evaluating, we translate the Hindi answer to English and create a gold answer set by combining the actual English answer and the translated English answer for each question. The performance of the system is reported in Table 3.9.

  For factoid questions, we obtain the maximum MRR value of 65.72 for the domain *Environment*. We obtain the lowest EM and MRR values for the domain *Diseases*. One possible reason could be that most of the factoid answers are the phrases, and the PoS tagger could not extract these correctly. The system achieves the maximum BLEU of 48.51 and ROUGE-L of 45.72 scores for the domains *Diseases* and *History*, respectively. Our model could not perform well for the descriptive questions of the domain *Tourism*. However, it is to be noted that *Tourism* contains only a few (29) short descriptive questions. Our close analysis reveals that the system suffers due to the errors encountered in the linguistic components, such as the PoS tagger and NEs tagger(RNN. The NE tagger could not detect some of the entities present in the translated Hindi passage, may be due to the errors encountered during translation.

---

[12] http://cogcomp.org/Data/QA/QC/
[13] http://cogcomp.org/Data/QA/QC/TREC_10.label

| Domains | Factoid | | Descriptive | |
|---|---|---|---|---|
| | EM | MRR | BLEU | ROUGE-L |
| Environment | 39.13 | 65.72 | 45.81 | 42.56 |
| History | 29.53 | 57.19 | 42.84 | 45.72 |
| Geography | 35.55 | 52.27 | 43.02 | 44.61 |
| Diseases | 23.29 | 34.78 | 48.51 | 39.19 |
| Economics | 26.28 | 46.89 | 45.12 | 44.77 |
| Tourism | 27.68 | 37.79 | 22.96 | 24.29 |
| Total | 30.24 | 49.10 | 41.37 | 40.19 |

Table 3.9: Performance (in %) of the proposed model for factoid and descriptive questions

## 3.4 Deep Neural Model for MQA

We propose a unified deep neural network-based approach for multilingual QA. The proposed network takes as an input the triplets of $<question, snippet, answer>$ for both English and Hindi languages. Specifically, the model takes the multilingual question and snippet[14] as an input and provide the answer, irrespective of the language of the question or snippet.

We have conducted experiments with two datasets, **(1)** Translated SQuAD and **(2)** Multilingual QA. The multilingual QA dataset consists of documents containing the passages for each question. We generate the snippet from the whole document in a question-focused summarization fashion. In the case of Translated SQuAD dataset, the paragraph (snippet) containing the answer is available for each question. The proposed algorithm for snippet generation is described as follows:

---
[14]In this work, we use the term snippet to represent the paragraph containing the answer.

### 3.4.1 Snippet Generation

In the recent work on English/Hindi QA [236, 231, 269], the focus is on passage extraction, considering only the lexical similarity. However, lexical similarity does not take into account the semantic information to curate the probable sentences where the answer could lie. This set of curated sentences is also known as a snippet. The snippets are automatically anchored around the question terms. To consolidate the relevant sentences from multiple documents, we propose a snippet generation algorithm. The inputs to the algorithm are questions and a set of documents. The output(s) is(are) the most probable sentence(s) supporting the evidence containing the answer(s). The algorithm takes into account the semantic information with lexical similarity to rank the probable sentences by considering their relevance to the question. Along with this, we represent the sentences of documents as a graph. Each pair of sentences is linked to the question based on their lexico-semantic similarity (obtained through word embeddings).

In our work, we have questions and documents in multilingual forms. The existing deep learning-based approaches [328, 327, 283] may not be feasible in our work because of the following reasons: **(a)** they require a sufficient amount of labeled data to train the model, and **(b)** the model should have the capability to process the multilingual inputs. Therefore, in this work, we propose an unsupervised approach with the flexibility to deal with the language-independent question/passage.

Our snippet generation algorithm is motivated by the passage retrieval task [197], where a graph-based query-focused summarization technique is used to retrieve the relevant passage. For a given question $q$ and a set of sentences $S = \{s_1, s_2, \ldots, s_n\}$, the proposed algorithm calculates the relevance score to each sentence $s \in S$ with respect to the question, as shown below:

$$p(s|q) = d \frac{rel(s,q)}{\sum_{p \in C} rel(p,q)} + (1-d) \sum_{v \in C} \frac{rel(s,v)}{\sum_{z \in v} rel(z,v)} p(v|q) \qquad (3.5)$$

where $d$ is termed as 'question bias' factor and $C = S - \{s\}$.

The first component of E.q. 3.5 determines the relevance of sentence $s$ to the question $q$, and the second component finds out its relevance to the other sentence. The term $d$ is a trade-off between the two components in the equation and is determined empirically[15]. We force the system to give more importance to the relevance of the question by providing a higher value of $d$ in the Eq. 3.5. We compute $p(s|q)$ with the help of the power method as discussed in [197]. The term $rel(X, Y)$ is the standard relevance score, which can be computed as follows:

$$V_{X(Y)} = \sum_{w \in X(Y)} log(1 + tf_{w,X(Y)}) * idf_w * Ma_w$$
$$rel(X, Y) = cosine(V_X, V_Y)$$
(3.6)

---
[15]The value of $d$ is set to 0.8 in our experiment.

Here, $tf_{w,X(Y)}$ is the frequency of word $w$ in $X(Y)$, $idf_w$ is the inverse document frequency of word $w$. $M \in \mathbb{R}^{d \times |V|}$ is the $d$ dimensional word embedding matrix of vocabulary $V$ word $w$ represented by their one hot vector representation $a_w$. The terms $V_X$ and $V_Y$ are the lexico-semantic representation of the entities $X$ and $Y$, respectively. The vector $V_{X(Y)}$ is normalized to avoid biases towards the long sentences. The sentences are ranked based on their relevance to the user's question. The top-most ranked three sentences are considered the candidate to belong to a snippet in our proposed multilingual network. We use the English-Hindi multilingual embedding trained via the technique discussed in [255], which helps the snippet generation technique to consider the multilingual words.

In this work, we attempt to solve the multilingual question-answering problem, especially in English-Hindi languages. Our proposed method employs a unified deep neural network-based model with the capability to process the English and Hindi question/document/snippet and provide the answer.

Our model is trained with the questions and snippets from English-Hindi languages simultaneously. The reason to train questions and snippets from both languages simultaneously is to adopt cross-lingual and multilingual settings in a proposed unified model. The first *Multilingual Sentence Encoding* layer encodes the question and snippet, which are in English and/or Hindi. This layer exploits the multilingual embedding to represent the multilingual words from question and snippet. Thereafter, the obtained question/snippet word representations are fed into a Bi-GRU network to obtain the sequence-level representation. The second *Shared Question Encoding* layer takes the English and Hindi question representation obtained from the previous layer and generates the shared representation of the question. We generate the shared representation of questions because the questions posed in English and Hindi languages are semantically similar. The shared representation is generated by the soft-alignment of words between English and Hindi questions. After that, *Snippet Encoding Layer*, which is a self-matching layer, provides the flexibility to dy-

**51**

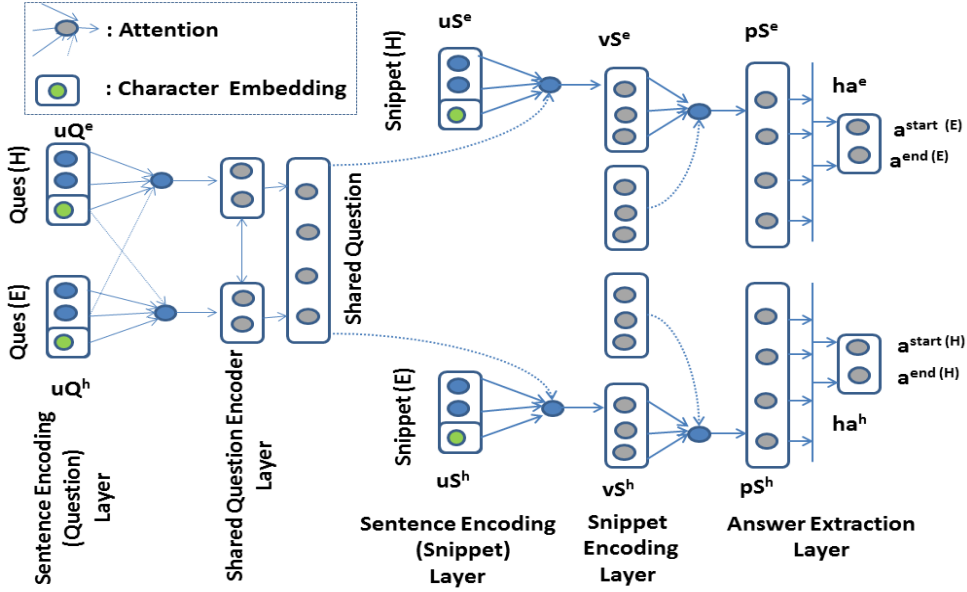

Figure 3.2: Structure of the proposed unified deep neural network for MQA.

namically collect information for each word by exploiting the whole snippet's information. Finally, we have *Answer Extraction Layer* that is based on the pointer network, which points to the start and end answer indices from the snippet. The structure of the model is depicted in Figure 3.2. We now describe the individual components of the proposed neural network model as follows:

### 3.4.2 Multilingual Sentence Encoding Layer

This layer is responsible to encode the multilingual question and snippet. Given an English question $Q_e = \{w_1^{Q_e}, \ldots, w_{m_e}^{Q_e}\}$, English snippet $S_e = \{w_1^{S_e}, \ldots, w_{n_e}^{S_e}\}$, Hindi question $Q_h = \{w_1^{Q_h}, \ldots, w_{m_h}^{Q_h}\}$ and English snippet $S_h = \{w_1^{S_h}, \ldots, w_{n_h}^{S_h}\}$, word-level embeddings $\{x_t^{Q_e}\}_{t=1}^{m_e}$, $\{x_t^{S_e}\}_{t=1}^{n_e}$, $\{x_t^{Q_h}\}_{t=1}^{m_h}$ and $\{x_t^{S_h}\}_{t=1}^{n_h}$ are generated from pre-trained multilingual word embedding table. To tackle the out-of-vocabulary (OOV) words, we employ character-level embedding $\{c_t^{Q_e}\}_{t=1}^{m_e}$, $\{c_t^{S_e}\}_{t=1}^{n_e}$, $\{c_t^{Q_h}\}_{t=1}^{m_h}$ and $\{c_t^{S_h}\}_{t=1}^{n_h}$. The character-level embeddings are generated using a Bi-GRU network which takes the characters as input. The final representation of each word $u_t^{Q_e}$ ($u_t^{Q_h}$) of English (Hindi) question and snippet $u_t^{S_e}$ ($u_t^{S_h}$) are obtained as follows:

$$\begin{aligned} u_t^{Q_k} &= \text{Bi-GRU}(u_{t-1}^{Q_k}, [x_t^{Q_k} \oplus c_t^{Q_k}]) \\ u_t^{S_k} &= \text{Bi-GRU}(u_{t-1}^{S_k}, [x_t^{S_k} \oplus c_t^{S_k}]) \end{aligned} \quad (3.7)$$

where $k \in \{e, h\}$ denotes the English(e) and Hindi(h) languages, $\oplus$ is the concatenation operator.

### 3.4.3 Shared Question Encoding Layer

In this layer, we obtain a shared representation of the encoded English $\{u_t^{Q_e}\}_{t=1}^{m_e}$ and Hindi question $\{u_t^{Q_h}\}_{t=1}^{m_h}$. Basically, we obtain the shared representation *via* soft-alignment of words [226, 296] between English and Hindi questions. Since both the questions are the same irrespective of their languages, therefore it contains the same information across the languages. With the help of soft-alignment of words between the questions of both languages, we obtain a better representation of a given question (in a language), which considers the same information in other languages. Given English and Hindi question representation $\{u_t^{Q_e}\}_{t=1}^{m_e}$ and $\{u_t^{Q_h}\}_{t=1}^{m_h}$, at first we obtain the co-language (English) aware Hindi question representation:

$$v_t^{Q_h} = \text{Bi-GRU}(v_{t-1}^{Q_h}, p_t^Q) \tag{3.8}$$

where $p_t^Q$ is an attention-based pooling vector. It is calculated as follows:

$$\begin{aligned} k_j^t &= \text{V}^T \tanh\Big( \big[W_u^{Q_e} W_u^{Q_h} W_v^{Q_h}\big] \big[u_j^{Q_e} u_t^{Q_h} v_{t-1}^{Q_h}\big]^T \Big) \\ p_t^Q &= \sum_{i=1}^{m_e} \Big( \exp(k_i^t) / \sum_{j=1}^{m_e} \exp(k_j^t) \Big) u_i^{Q_e} \end{aligned} \tag{3.9}$$

where $\text{V}^T$ is a weight vector, $W_u^{Q_e}$, $W_u^{Q_h}$ and $W_v^{Q_h}$ are the weight matrices.

To compute the representation $(v_t^{Q_h})$ at time $t$ of Hindi question (equation 3.8) using Bi-GRU, we concatenate the pooling vector $p_t^Q$ with the representation $(v_{t-1}^{Q_h})$ at time $(t-1)$. The pooling vector is computed by the weighted representation of Hindi question representation $u_t^{Q_e}$ at time $t$ in Eq. 3.9. The Hindi question representation is computed by considering the English question representation therefore, we called it co-language aware question representation. Similarly, we compute the English question representation $v_t^{Q_e}$. The shared question representation is obtained by concatenating both the language-aware question representations. The final question representation will be $\{v_t^Q\}_{t=1}^{(m_e+m_h)} = \{v_t^{Q_e}\}_{t=1}^{m_e} \oplus \{v_t^{Q_h}\}_3 t = 1^{m_h}$.

### 3.4.4 Snippet Encoding Layer

The snippet encoding generated from the sentence encoding layer (c.f. Section 3.4.2) does not account for question information. To incorporate the question information into the snippet representation, we follow the attention-based RNN. We generate the snippet representation of both English and Hindi by taking the shared question information into account. The English snippet representation is computed by:

$$v_t^{S_e} = \text{Bi-GRU}(v_{t-1}^{S_e}, c_t^{S_e}) \tag{3.10}$$

where $c_t^{S_e}$ is an attention based pooling vector, which is derived *via* the following equations:

$$\begin{aligned}
k_j^t &= V^T \tanh\Big( \big[W_v^Q W_u^{S_e} W_v^{S_e}\big] \big[v_j^Q u_t^{S_e} v_{t-1}^{S_e}\big]^T \Big) \\
c_t^{S_e} &= \sum_{i=1}^{m_e+m_h} \Big( \exp(k_i^t) / \sum_{j=1}^{m_e+m_h} \exp(k_j^t) \Big) v_i^Q
\end{aligned} \tag{3.11}$$

where, $W_v^Q$, $W_u^{S_e}$ and $W_v^{S_e}$ are the learnable weight matrices. The snippet representation $v_t^{S_e}$ dynamically incorporates aggregated matching information from the whole question. Similarly, we compute the Hindi snippet representation $v_t^{S_h}$. In order to capture the context information while generating the snippet representation, we introduce an additional layer similar to [296]. The context plays an important role in discovering the answer from a snippet. This additional layer matches the obtained snippet representation from the *snippet encoding layer* against itself. This layer provides the facility to dynamically collect evidence from the whole snippet. It encodes the evidence relevant to the current snippet word and its matching question information into the snippet representation. The final snippet representation for the English snippet is computed as follows:

$$p_t^{S_e} = \text{Bi-GRU}(p_{t-1}^{S_e}, [v_t^{S_e}, c_t^{S_e}]) \tag{3.12}$$

where $c_t^{S_e}$ is an attention-based pooling vector for the entire English snippet, it is computed in the following manner:

$$\begin{aligned}
k_j^t &= V^T \tanh\Big( \big[W_{p'}^{S_e} W_{p''}^{S_e}\big] \big[v_j^{S_e} v_t^{S_e}\big]^T \Big) \\
c_t^{S_e} &= \sum_{i=1}^{n_e} \Big( \exp(k_i^t) / \sum_{j=1}^{n_e} \exp(k_j^t) \Big) v_i^{S_e}
\end{aligned} \tag{3.13}$$

where, $W_{p'}^{S_e}$ and $W_{p''}^{S_e}$ are the learnable weight matrices. We compute the snippet representation for the Hindi snippet in the same way as the English snippet. The final snippet representations that we obtain are $\{p_t^{S_e}\}_{t=1}^{n_e}$ and $\{p_t^{S_h}\}_{t=1}^{n_h}$ for English and Hindi, respectively.

### 3.4.5 Answer Extraction Layer

We utilize the pointer network proposed by [286] to extract the answer from the snippet. We use two pointer networks, one to select the start ($a_e^{start}$) and end ($a_e^{end}$) index of the answer from the English snippet and another from the Hindi snippet. Given the English snippet representation $\{p_t^{S_e}\}_{t=1}^{n_e}$, with the help of the attention mechanism, networks select the start and end indices of the answer. The hidden state of pointer network is calculated by $h_t^{a_e} = \text{Bi-GRU}(h_{t-1}^{a_e}, c_t^{S_e})$, where $c_t^{S_e}$ is the attention pooling vector. It can be computed as follows:

$$\begin{aligned}
k_j^t &= V^T \tanh\left(\left[W_p^{S_e} W_h^{a_e}\right]\left[p_j^{S_e} h_{t-1}^{a_e}\right]^T\right) \\
a_i^t &= \exp(k_i^t)/\sum_{j=1}^{n_e} \exp(k_j^t) \\
c_t^{S_e} &= \sum_{i=1}^{n_e} a_i^t p_i^{S_e} \\
a_e^t &= argmax(a_1^t, .., a_{n_e}^t)
\end{aligned} \quad (3.14)$$

At the first step ($t = 1$) network will predict $a_e^{start}$, and in the next step, it will predict $a_e^{end}$. In a similar way, we compute $a_e^{end}$. Following E.q. 3.14, the answer index $a_h^{start}$ and $a_h^{end}$ from the Hindi snippet are extracted.

## 3.5 Datasets and Experimental Details

### 3.5.1 Experimental Setup

We perform experiments in six different multilingual settings.

1. **$Q_E - S_{E+H}$**: The question is in *English* and the answer exists in both *English* and *Hindi* snippets. The model has to retrieve the answer from both snippets. This setting is equivalent to the cross-lingual and multilingual evaluation setup of QA.
2. **$Q_H - S_{E+H}$**: The question is in *Hindi* and the answer exists in both *English* and

*Hindi* snippets. The model has to retrieve the answer from both snippets. This setting is equivalent to the cross-lingual and multilingual evaluation setup of QA.

3. **$Q_E - S_E$**: Both question and answer are in *English*. The model has to retrieve the answer from the *English* snippet. This setting is equivalent to the monolingual evaluation setup of QA.

4. **$Q_H - S_H$**: Both question and answer are in *Hindi*. The model has to retrieve the answer from the *Hindi* snippet. This setting is equivalent to the monolingual evaluation setup of QA.

5. **$Q_E - S_H$:** The question is in *English* and the answer exist in *Hindi* snippet. The model has to retrieve the answer from the Hindi snippet. This setting is equivalent to a cross-lingual evaluation setup of QA.

6. **$Q_H - S_E$:** The question is in *Hindi* and the answer exist in *English* snippet. The model has to retrieve the answer from the English snippet. This setting is also equivalent to the cross-lingual evaluation setup of QA.

It is to be noted that we train our model with the bi-triplet $<question_e, snippet_e, answer_e>$ and $<question_h, snippet_h, answer_h>$ input from the English and Hindi languages, respectively. Both the triplets have the same information in two different languages. The proposed network is trained to minimize the sum of the negative log probability of the ground truth start and end indices of the answers in both languages by the predicted probability distributions of the model. By training the network with the bi-triplet of both languages, the network learns to handle the different settings of multilingual questions and snippets. At the time of evaluation, when the network receives questions or snippets from one language, we replicate the same for the other language to keep the inputs compatible with the model. For experiments, we use the publicly available *fastText* [22] pre-trained English and Hindi word embeddings of dimension 300. For multilingual word embedding, we align monolingual vectors of English and Hindi in a unified vector space using a learned linear transformation matrix [255]. We use the Stanford CoreNLP [175] to pre-process all the English sentences. The model with character-level embeddings of dimension 45 shows the highest performance on the validation set. The optimal dimension of hidden units for all the layers is set to 45 in the experiment. We exploit two layers of Bi-GRU to compute character embedding and three layers to obtain the question and snippet representation, respectively. Mini-batch gradient descent (batch size of 50) with the AdaDelta optimizer [338] is used to train the network with a learning rate of 1.

The network is trained for 70 epochs. The hyper-parameters are tuned using a validation dataset.

### 3.5.2 Datasets

We use two different multilingual question-answering datasets in our experiment to evaluate the performance of the proposed model.

**Translated SQuAD dataset:** We translate 18,454 random English <*question, passage, answer*> triplet from Squad dataset [216] into Hindi. These translated triplets ensure that the answer is a substring of passage. We divide this dataset into the train, validation, and test sets. We use a set of 10,454 QA pairs in English and Hindi for training the network. Another set of 2000 QA pairs are used to validate the system performance over every epoch. We use a set of 6,000 QA pairs for evaluating the system performance.

**Multilingual QA dataset:** We use the created MMQA dataset (c.f. Section 3.2) to evaluate the model. The detailed statistics of this dataset are given in Table 3.10. This dataset also provides us with the source documents where the answer exists for the questions. In the practical scenario, we only have a question and need to retrieve its answer from different documents, not necessarily in the same language as that of the question. With this fact in mind, we perform the experiments in different multilingual settings (c.f. Section 3.5.1). For each question, we generate the snippet following the approach discussed in Section 3.4.1. This dataset is only used for evaluating model performance. To compare the performance between the different multilingual settings, we could only use the data samples listed in the category of $Q_E - S_{E+H}$ and $Q_H - S_{E+H}$.

| Domains | $Q_E - S_E$ | $Q_H - S_H$ | $Q_E - S_H$ | $Q_H - S_E$ | $Q_E - S_{E+H}$ | $Q_H - S_{E+H}$ | Overall |
|---|---|---|---|---|---|---|---|
| **Tourism** | 456 | 403 | 456 | 403 | 422 | 422 | 1,703 |
| **History** | 110 | 126 | 110 | 126 | 1,118 | 1,118 | 2,472 |
| **Diseases** | 81 | 33 | 81 | 33 | 48 | 48 | 210 |
| **Geography** | 55 | 29 | 55 | 29 | 174 | 174 | 432 |
| **Economics** | 25 | 14 | 25 | 14 | 682 | 682 | 1,403 |
| **Environment** | 9 | 2 | 9 | 2 | 226 | 226 | 463 |
| **Overall** | **736** | **607** | **736** | **607** | **2,670** | **2,670** | **6,683** |

Table 3.10: Statistics of the multilingual QA dataset.

### 3.5.3 Evaluation Scheme

We evaluate the system performance using EM and F1 metrics following [216]. For multilingual setting $Q_E - S_{E+H}$ and $Q_H - S_{E+H}$, we count the correct prediction only when the model produces the correct answer from both the snippets. For the rest of the experimental settings, we count the correct prediction when the model produces the correct answer from the particular snippet.

## 3.6 Results and Analysis

### 3.6.1 Baselines

1. **IR-based QA model:** We develop a translation-based baseline model for the comparison. This baseline is adopted from the state-of-the-art models in English-Hindi QA, as discussed in Section 3.3.

2. **RNN-based QA model:** Similar to the IR-based baseline, we translate[16] the Hindi question and Snippet into English. The question and snippet encodings are performed, as discussed in Section 3.4.2. Thereafter, we incorporate the question information into the snippet by applying the attention mechanism similar to E.q. 3.10 and 3.11 to regenerate the snippet representation. This snippet representation of a word (from snippet) at time $t$ is fed to a feed-forward neural network. This network computes the vectors of the probability score of $p_t$. The length of the probability vector is set to 3, representing the BIO encoding (B-beginning, I-intermediate, and O-outside) of the answer. This model is similar to the attention-based QA-LSTM model proposed by the [277], but instead of computing the similarity between question and snippet as in [277], we classify the token at time $t$ from the snippet into 'B-answer', 'I-answer' and 'O'.

3. **Monolingual (English) QA model:** This baseline is similar to the monolingual version of the proposed network (c.f. Section 3.4). In the first layer of this baseline model, the English question and snippet are encoded as discussed in Section 3.4.2. As we are dealing with only one language, *shared question encoding layer* does not exist in this particular baseline model. The output of *sentence encoding layer* is passed to the *snippet encoding layer* (c.f. Section 3.4.4). Finally, *answer extraction*

---
[16]In all baseline models, translation is performed using Google translation.

*layer* (c.f. Section 3.4.5) predicts the start and end indices of the answer from the snippets.

4. **Monolingual (Hindi) QA model:** We propose a fourth baseline similar to the monolingual (English) baseline. The input question and snippet are in the Hindi language. Hyperparameters of both monolingual models are kept the same as those of the multilingual model.

5. **Deep Canonical Correlation Analysis (Deep CCA):** Deep CCA [11] computes representations of the two views by passing them through multiple stacked layers of nonlinear transformation. We experiment with Deep CCA by treating English and Hindi question representations as two different views of the same question. In our experiment, we use four layers of the GRU network to compute the representation of both views. Basically, from our proposed model, we replace the *Shared Question Encoding* layer with Deep CCA, which computes the shared representation by taking the two question views (representation) as inputs. The goal is to learn parameters for both views jointly such that the correlation between the final obtained representations is as high as possible. The hyperparameters of the Deep CCA model are kept the same as those of the proposed multilingual model.

### 3.6.2 Results

We evaluate the performance of the proposed snippet generation algorithm in terms of MRR. We achieve the MRR values of 95.48% over the standard Biased LexRank [197] of 91.71%, on the ground truth passage provided in the multilingual QA dataset. We show the evaluation results on MQA for the multilingual question answering and translated SQuAD dataset in Table 3.11 and Table 3.12 for multilingual QA and Translated SQuAD dataset, respectively. The proposed model achieves 7.23, and 11.7 absolute F1 point increments over the attention-based RNN baseline for the multilingual QA and Translated SQuAD datasets, respectively. Similarly, the proposed model achieves 5.86 and 5.14 absolute F1 point increments over the Deep CCA baseline for the multilingual QA and Translated SQuAD datasets, respectively. Statistical t-test confirms this improvement to be statistically significant (t-test, $p < 0.05$). We observe that $Q_H - S_{E+H}$ performs slightly lower than $Q_E - S_{E+H}$. It may be because of the smaller size corpus used for generating the Hindi embeddings.

| | Models | $Q_E - S_E$ EM (F1) | $Q_H - S_H$ EM (F1) | $Q_E - S_H$ EM (F1) | $Q_H - S_E$ EM (F1) | $Q_E - S_{E+H}$ EM (F1) | $Q_H - S_{E+H}$ EM (F1) | Overall EM (F1) |
|---|---|---|---|---|---|---|---|---|
| Baselines | IR based QA | 33.46 (39.81) | 32.63 (38.12) | 30.24 (32.94) | 27.67 (30.04) | 32.17 (39.67) | 30.78 (37.97) | 31.15 (36.42) |
| | RNN based QA | 37.18 (41.74) | 34.75 (40.32) | 32.14 (33.85) | 28.22 (29.61) | 35.49 (41.85) | 33.79 (39.12) | 33.59 (37.74) |
| | Monolingual (Hindi) | 36.12 (42.67) | 41.38 (47.79) | 30.97 (33.54) | 28.41 (30.08) | 38.31 (44.61) | 38.71 (44.94) | 35.65 (40.60) |
| | Monolingual (English) | 44.17 (49.35) | 35.52 (41.11) | 31.23 (33.97) | 29.11 (31.71) | 39.18 (46.64) | 35.17 (41.29) | 35.73 (40.67) |
| | Deep CCA | 41.21 (43.48) | 37.79 (40.23) | 31.62 (33.89) | 30.34 (32.65) | 39.76 (42.23) | 38.23 (42.19) | 36.49 (39.11) |
| | Proposed Multilingual | 44.78 (50.27) | 41.46 (48.14) | 34.68 (37.89) | 33.41 (37.02) | 42.28 (49.01) | 40.06 (47.49) | 39.44 (44.97) |

Table 3.11: Performance comparison of proposed MQA model (on Multilingual QA dataset) with the various baseline models.

| | Models | $Q_E - S_E$ EM (F1) | $Q_H - S_H$ EM (F1) | $Q_E - S_H$ EM (F1) | $Q_H - S_E$ EM (F1) | $Q_E - S_{E+H}$ EM (F1) | $Q_H - S_{E+H}$ EM (F1) | Overall EM (F1) |
|---|---|---|---|---|---|---|---|---|
| Baselines | IR based QA | 35.17 (37.78) | 32.87 (36.55) | 31.45 (33.13) | 28.12 (30.69) | 34.67 (36.54) | 31.22 (35.16) | 32.25 (34.97) |
| | RNN based QA | 44.68 (45.51) | 41.24 (44.71) | 33.27 (36.89) | 31.59 (33.86) | 42.56 (46.94) | 39.33 (44.54) | 38.77 (42.07) |
| | Monolingual (Hindi) | 43.78 (47.41) | 49.81 (53.27) | 35.01 (38.78) | 37.14 (41.85) | 47.77 (51.29) | 48.18 (52.21) | 43.61 (47.46) |
| | Monolingual (English) | 52.49 (56.11) | 43.17 (48.37) | 41.54 (35.53) | 33.11 (37.54) | 52.38 (56.61) | 45.11 (49.35) | 44.63 (47.25) |
| | Deep CCA | 44.78 (50.27) | 41.46 (48.14) | 42.04 (46.68) | 40.84 (44.86) | 51.19 (53.38) | 45.06 (48.49) | 44.28 (48.63) |
| | Proposed Multilingual | 53.15 (57.29) | 51.34 (53.87) | 45.34 (50.24) | 44.19 (48.21) | 54.38 (58.39) | 52.27 (54.67) | 50.11 (53.77) |

Table 3.12: Performance comparison of proposed MQA model (on test set of Translated SQuAD dataset) with the various baseline models.

To ensure the quality of translation from Google Translate, we perform a human evaluation of the Google translation. We randomly choose 100 question-snippet pairs from the English (SQuAD) dataset and translate them to Hindi. For translation, we employ two annotators having expertise in both English and Hindi. We computed the BLEU score [200] and found the score as 72.13.

### 3.6.3 Analysis and Discussion

In this section, we present the analysis of the results obtained in terms of the effect of shared question encoding and the ablation study. In addition to this, we also compare the quality of answers extracted using the proposed multilingual model and the Deep CCA model.

**Effect of Shared Question Encoding:** This layer learns the word or phrase of the question, which needs to provide more focus to the question of the other language while generating the question representation. We analyze through attention weight that the model learns to align the same/similar words from the questions across the languages (English and Hindi). The effect of shared question representation is evident when we look at the Monolingual (English) and Monolingual (Hindi) baselines performance in Table 3.11 and Table 3.12, respectively. Both of these baselines do not have a shared question encoding layer. The Monolingual (Hindi) model favors the question and snippet, which are in Hindi, and it shows a comparable performance close to the RNN-based baseline for the English question and/or snippet ($Q_E - S_E$, $Q_E - S_{E+H}$). We also observed quite a similar trend for the Monolingual (English) baseline model. The evaluation shows that the proposed multilingual system performs better in all the multilingual settings compared to the monolingual baselines.

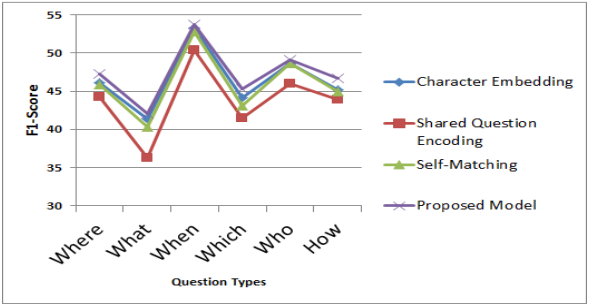

| Models | Multilingual QA | | Translated SQuAD | |
|---|---|---|---|---|
| | EM | F1 | EM | F1 |
| Proposed Model | 39.44 | 44.97 | 50.11 | 53.77 |
| -Shared Question Embedding | 35.62 | 41.18 | 46.37 | 49.91 |
| -Character Embeddings | 38.12 | 43.26 | 48.84 | 52.53 |
| -Self Matching | 37.23 | 42.39 | 48.02 | 51.84 |

Figure 3.3: Effect of model components on the various type of questions from the MQA dataset.

Table 3.13: Results of ablation study (by removing one model component at a time) on both the dataset.

**Ablation Study:** We carefully observe the effect of various components of the model. We show the ablation study in terms of EM and F1 scores on the multilingual QA dataset in Table 3.13. The analysis reveals that shared question encoding represents the questions of two languages very effectively, by aggregating the information from the questions. The character embedding helps the model overcome the out-of-vocabulary words and short words, which are often in Hindi questions and snippets. The self-matching of snippets assigns more weights to the words (in a snippet) related to the question and the context in which the answer appears. We extend our experiment by analyzing the model performance on the various question types such as *what, where, when, how, which, who*. Figure 3.3 shows the impact (in terms of F1 score) of model components (by removing a component at a time) on different types of questions of the multilingual QA dataset. Our model achieves the best performance on *'when'* type question because *'when'* type question generally looks for 'date' and 'time' as the answer. However, for *'what'* type of question, the model achieves a comparatively low F1 score. This is because the *'what'* type of questions look for a long phrase as the answer. The study reveals that the shared question encoding has a higher impact on the performance of the model for all types of questions.

We have translated the question/snippet in baseline 1 and baseline 2 only. We did not translate the question/snippet in our proposed model. The Monolingual (English) and Monolingual (Hindi) models are trained on the question and snippet from the English and Hindi languages. In the $Q_E - S_{E+H}$ and $Q_H - S_{E+H}$ settings, the model receives the cross-lingual inputs. Therefore the monolingual model could not achieve as good performance as our proposed multilingual model. The proposed model has a shared question encoder and has the capability of processing cross-lingual and multilingual inputs. This is the reason why the proposed model achieves the improvements on $Q_E - S_{E+H}$ and $Q_H - S_{E+H}$ settings compared to the monolingual (English) and monolingual (Hindi) model.

We observe that the model performance on the multilingual QA dataset is relatively lower than the Translated SQuAD multilingual dataset. This is because the model is trained on the Translated SQuAD multilingual dataset and learns the diverse answers from the dataset, which may not exist in the multilingual QA dataset. Due to the unavailability of any other MQA (EN-HI) dataset, we can not make any direct comparison. However, our IR-based baseline is the re-implementation of the state-of-the-art work of [236] on EN-HI cross-lingual QA and obtains significantly better performance compared to the state-of-the-art model. Most of the available French/German-English dataset (CLEF) is small and developed in the cross-lingual setting. However, the dataset used in this work provides monolingual, cross-lingual, and multilingual settings. To the best of our knowledge, this is the very first work in multilingual question answering where for a given multilingual question, the corresponding answer is extracted from various multilingual snippets.

### 3.6.4 Qualitative Analysis

We qualitatively analyze the answers predicted by the proposed system. The examples are shown in Table 3.14. The analysis shows that the proposed system performs very well for the question which is looking for the named entity type answer. Our further analysis reveals that the proposed system performs exceptionally well to identify the '*number*', '*date*', '*quantity*', "*person name*" types of answers.

We closely analyze the major sources of errors in Section 3.6.5. The model learns to identify the semantically similar words in snippets, and sometimes it predicts the semantically similar words as the answer. We compare the performance of the CCA-based model to the proposed model- both quantitatively and qualitatively. We show the question, snippet along with their answers predicted from the proposed model and Deep CCA in Table 3.14. The Deep CCA model suffers from out-of-context answers. In cross-lingual setups ($Q_H-Q_E$) and ($Q_E-S_H$), the Deep CCA model does not perform well compared to the proposed model. We also observe that the Deep CCA model extracts the long sentence answer. The Deep CCA model tries to maximize the correlation between English and Hindi representation and learns the shared question representation. While maximizing the correlation Deep CCA focuses on the question representation as a single vector. In contrast, our shared question encoding layer tries to find the alignment between the English and Hindi question representation by considering each word from the English and Hindi questions. In addition, our model generates the shared question representation by considering the English-aware Hindi and Hindi-aware English representation (c.f. Section 3.4.3).

### 3.6.5 Error Analysis

We closely analyze the outputs on the multilingual QA dataset and come up with the following observations:

1. The system suffers to predict the correct answer, where the answer entity is the anaphor or cataphor in the snippet. E.g.
   **Q:** *What is the part of the Adam's Bridge?*,
   **Gold Answer:** Pamban Island
   **Snippet:** ***Pamban Island*** *is situated in the Gulf of Mannar between India and Srilanka...* ***It*** *is a part of the Adam's Bridge.*
   As shown in the example the word '**it**' (pronoun) is referring to the phrase **'Pamban Island'**, and these two words are far apart (in terms of the number of words between these two words) in the passage. Therefore, the model could not identify the correct referred phrase **'Pamban Island'**. Resolving such pronouns in the snippet before passing it into the network should lead to performance improvements.
2. Sometimes the system predicts the wrong answer from the snippet. This generally happens in the case the named entity (NE) appears in the vicinity. E.g.
   **Q:** *How far is the Taj Mahal from New Delhi?*
   **Gold Answer:** 230 KM;

| |
|---|
| **Question (1):** Which company adopted the ASA scale in 1946? |
| **Snippet:** General Electric switched to use the ASA scale in 1946. Meters manufactured since February 1946 were equipped with the ASA scale -LRB- labeled " Exposure Index " -RRB- already . For some of the older meters with scales in " Film Speed " or " Film Value " -LRB- e.g. models DW-48 , DW-49 as well as early DW-58 and GW-68 variants -RRB- , replaceable hoods with ASA scales were available from the manufacturer ... |
| **Gold Answer:** General Electric |
| **Answer using Deep CCA:** DW-48 |
| **Answer using Proposed Model:** General Electric |
| **Question (2):** एलजीबीटी के अधिकारों के लिए कौन सा मील का पत्थर माना जाता है? |
| **Trans**: Which landmark is considered the spark for LGBT rights? |
| **Snippet:** The Statue of Liberty National Monument and Ellis Island Immigration Museum are managed by the National Park Service and are in both the states of New York and New Jersey . ... Hundreds of private properties are listed on the National Register of Historic Places or as a National Historic Landmark such as, for example , the Stonewall Inn in Greenwich Village as the catalyst of the modern gay rights movement . |
| **Gold Answer:** Stonewall Inn |
| **Answer using Deep CCA:** Governors Island National Monument |
| **Answer using Proposed Model:** Stonewall Inn in Greenwich Village |
| **Question (3):** How did naturalism effect the greater world? |
| **Snippet:** ...But as the 19th-century went on , European fiction evolved towards realism and naturalism , the meticulous documentation of real life and social trends. Much of the output of naturalism was implicitly polemical, and influenced social and political change , but 20th century fiction and drama moved back towards the subjective, emphasising unconscious motivations and social and environmental pressures on the individual ... |
| **Gold Answer:** influenced social and political change |
| **Answer using Deep CCA:** primacy of individual experience |
| **Answer using Proposed Model:** social and political developments |
| **Question (4):** ज़ार अलेक्ज़ेंडर ने चोपिन को क्या दिया? |
| ( **Trans**: What did Tsar Alexander I give to Chopin?) |
| **Snippet:** सितंबर 1823 से 1826 तक चोपिन वारसा लिसेयुम में भाग लिया जहां उन्होंने अपने पहले वर्ष के दौरान चेक संगीतकार विल्हेम वार्फ़ेल से अंग सबक प्राप्त किए ज़ार ने उसे एक हीरे की अंगूठी प्रस्तुत किया 10 जून 1825 को बाद के ईओलोमेलोडिकॉन कॉन्सर्ट में चोपिन ने अपने रोंडो ओप का प्रदर्शन किया |
| ( **Trans:** From September 1823 to 1826 Chopin attended the Warsaw Lyceum , where he received organ lessons from the Czech musician Wilhelm Wurfel during his first year. Tsar presented him with a diamond ring . At a subsequent eolomelodicon concert on 10 June 1825 , Chopin performed his Rondo Op)... |
| **Gold Answer:** हीरे की अंगूठी |
| **Answer using Deep CCA:** रोंडो ओप ( **Trans:** Rondo Op ) |
| **Answer using Proposed Model:** हीरे की अंगूठी ( **Trans:** diamond ring ) |
| **Question (5):** Who is responsible for appointing the Lieutenant Governor of the Union Territory of Delhi? |
| **Snippet:** The head of state of Delhi is the Lieutenant Governor of the Union Territory of Delhi, appointed by the President of India... |
| **Gold Answer:** President of India |
| **Answer using Deep CCA:** Lieutenant Governor |
| **Answer using Proposed Model:** President of India |
| **Question (6):** What particle is associated with the yellowing of newspapers? |
| **Snippet:** Paper made from mechanical pulp contains significant amounts of lignin , a major component in wood . In the presence of light and oxygen , lignin reacts to give yellow materials , which is why newsprint and other mechanical paper yellows with age ... |
| **Gold Answer:** lignin |
| **Answer using Deep CCA:** lignin |
| **Answer using Proposed Model:** lignin |

Table 3.14: Examples of the question, snippet, gold answer, and the predicted answer using Deep CCA and our proposed model. The answers are shown in red.

**Predicted Answer:** 310 KM

**Snippet:** *Taj is located within the distance of 310 km and **230 Km** from Lucknow and national capital New Delhi respectively...*

In this example, there are two numbers (*310 km* and *210 km*) appear very near in the snippet. The network fails to correctly map the associated number (**230 km**).

3. While analyzing the outputs of snippet generation, we observe that during the translation of Hindi sentences in snippet generation, some synonym words and named entities were incorrectly translated. E.g. **Q:** *When Mahatma Gandhi visited Darjeeling?*

The prompt translation of documents: *"..Mahatma Gandhi traveled to Darjeeling in 1925...".* The word *visited* has been replaced with *traveled*, so the snippet generation algorithm ranks it to the lower in order.

4. Our proposed network sometimes was unable to identify the correct start or end index of the answer in the snippet. It contributes to the major sources of errors. An example of this type of error is shown as question (2) in Table 3.14. This phenomenon is observed more often in cross-lingual settings. The prediction of the end index can be improved by providing the predicted start index information to the network before making the prediction of the end index.

5. The network was also unable to provide an answer where reasoning across multiple sentences is required. We also observe a similar behavior, signifying that the network fails to provide the correct answer, where the answer and the headwords (query) in the question are far apart (2 to 3 sentences away). Example:

   **Q**: *The climate of Greece in the Northwest is known as what?*,

   **Snippet**: *The mountainous areas of Northwestern Greece -LRB- parts of Epirus , Central Greece , Thessaly , Western Macedonia -RRB- as well as in the mountainous central parts of Peloponnese – including parts of the regional units of Achaea , Arcadia and Laconia – feature an Alpine climate with heavy snowfalls . ... . Snowfalls occur every year in the mountains and northern areas , and brief snowfalls are not unknown even in low-lying southern areas , such as Athens.*

   **Gold Answer**: Alpine climate

   **Predicted Answer**: Western Macedonia

   In this example model has to perform the reasoning across multiple sentences to conclude the correct answer. This type of errors can be addressed by the multi-step of reasoning similar to the work of [50].

6. One of the limitations of the network is that it does not correctly identify the answer to short descriptive questions started with '*why*' or '*how*'. In these types of errors, the network could not predict the correct answer indices. It is because of the network that has to predict the correct phrase not limited to nouns, verbs, or adjectives. The prediction of the complex phrase is difficult as compared to the prediction of the named entities. Example:

**Q**: *Why did they miss that competition?*,

**Snippet**: *Defending holders Manchester United did not enter the 1999 – 2000 FA Cup, as they were already in the inaugural Club World Championship, with the club stating that entering both tournaments would overload their fixture schedule and make it more difficult to defend their Champions League and Premiership titles. The club claimed that they did not want to devalue the FA Cup by fielding a weaker side ...*

**Gold Answer**: The club claimed that they did not want to devalue the FA Cup by fielding a weaker side.

**Predicted Answer**: their handling of the situation

7. The network also suffers to find the correct answer in the cross-lingual setup ($Q_E - S_H$) where the answer words are not named entities and consist of descriptive answers. Example:

   **Q:** कई चीनी सैनिकों की एक बड़ी चिंता क्या थी?

   **Trans**: *What was a biggest concern of many Chinese troops?*,

   **Snippet**: *.. What Chinese soldiers feared, Hong said, was not the enemy, but that they had nothing to eat, no bullets to shoot, and no trucks to transport them to the rear when they were wounded ...*

   **Gold Answer**: they had nothing to eat

   **Predicted Answer**: supply

## 3.7 Conclusion

In the first part of this chapter, we propose a new multilingual QA dataset: MMQA. The dataset has a broad coverage of various entities as the answer. It can be used to build a monolingual (EN: English, HI: Hindi), cross-lingual (EN → HI, HI → EN), and multilingual (EN ↔ HI) QA system. We also proposed and discussed IR based framework for multilingual question answering. Our analysis reveals that a significant proportion of questions require some reasoning ability to provide the correct answers.

In the second part of this chapter, we tackle the limitations of IR based system by proposing a unified deep neural network technique for multilingual question answering. The proposed model is a generic framework with the flexibility of being adaptable to any number of languages. To provide the input snippet (if not available) to the proposed network, we introduce an effective language-independent snippet generation algorithm. Our snippet generation algorithm exploits the lexico-semantic similarity between the sentences. The proposed network utilizes the soft alignment of the question words from the English and Hindi languages to learn the question's shared representation. The learned shared representation of question and attention-based snippet representation is passed as an input to the answer extraction layer of the network, which extracts the snippet's answer span. We achieve state-of-the-art performance on the multilingual benchmark QA dataset. The evaluation shows that our proposed model attains 39.44 EM and 44.97 F1 values.

In the next chapter, we will discuss our proposed approach for code-mixed question generation that aims to automatically generate code-mixed questions by attentively modeling various linguistic phenomena of code-mixed languages. Later, we elaborate on our novel algorithm based on the concept of answer-type aware extraction technique for solving code-mixed question answering task.

# Chapter 4

# Code-Mixed Question Answering

## 4.1 Introduction

In the previous chapter, we developed the neural network-based framework focusing on soft alignment of the words from the English and Hindi languages to learn the effective, shared representation of questions/documents to address the problem of multilingual question answering from textual documents. In this chapter, we aim to solve some of the challenges of code-mixed question answering from textual documents. Towards this, first, we create a benchmark dataset for code-mixed question answering focused on Hindi-English language pair. Second, we propose an effective neural network method to model the phenomena of code-mixing in question-answering tasks.

Generally, multilingual people often switch back and forth between their native languages and foreign (popular) languages to express themselves on the web. This is common nowadays, mainly when people express their opinions (or make any communication) through various social media platforms. This phenomenon of embedding the morphemes, words, phrases, etc. of one language into another is popularly termed code-mixing [188, 189]. The recent study [230] has uncovered that users frequently use question patterns, namely 'how' (38%), 'why' (24%), 'where' (15%), 'what' (11%), and 'which' (12%) in their queries as opposed to a 'statement query'.

Presently, search engines have become more intelligent and are capable enough to provide a precise answer to a natural language query/question[1] .Several virtual assistants, such as Siri, Cortana, Alexa, and Google Assistant are also equipped with these facilities.

---

[1]The capability of handling factoid questions is higher than the complex questions or descriptive questions.

However, these search engines and virtual assistants are efficient in only handling queries made in the English language. Let us consider the following two representations (English and code-mixed) of the same question.

(i) **Question:** *"Who is the foreign secretary of USA?"*

(ii) **Question:** *"USA **ke** foreign secretary **kaun hai**?"*

(**Translation:***"Who is the foreign secretary of USA?"*)

Search engines can provide the exact answer to the first question. It is to be noted that although both questions are the same, the search engine is unable to return the exact answer for the second question, which is code-mixed. It instead returns the topmost relevant web pages.

In this chapter, we propose a framework for CMQG and CMQA involving English and Hindi. First, we propose a linguistically-motivated technique for generating code-mixed questions. We follow this approach as we do not have access to any labeled data for code-mixed question generation. After that, we propose an effective framework based on a deep neural network for CMQA. In our proposed CMQA technique, we use multiple attention-based recurrent units to represent the code-mixed questions and the English passages. Finally, our answer-type focused network (attentive towards the answer type of the question) extracts the answer for a given code-mixed question.

### 4.1.1 Problem Statement

Given a code-mixed question $Q$ with tokens $\{q_1, q_2 \ldots, q_m\}$ and an English passage $P$ having tokens $\{p_1, p_2 \ldots, p_n\}$, where $m$ and $n$ are the numbers of tokens in question and passage, respectively. The task is to identify the answer $A$ with tokens $\{p_i, p_{i+1} \ldots, p_j\}$ of length $j - i + 1$, where $1 \leq i \leq n$ and $i \leq j \leq n$.

The example of code-mixed questions is shown in Table 4.1. As observed in Table 4.1, the question is in a mixed script (English and Hindi), while the passage is in the English language. The system needs to identify and retrieve the start and end position of the answer from the passage for a given code-mixed question. In this work, since we aim to retrieve the answer from a single passage, we formulated the task of code-mixed QA as a reading comprehension problem.

> **Passage:** English literature began to reappear after 1200 when a changing political climate and the decline in Anglo-Norman made it more respectable. The Provisions of Oxford, released in **1258**, was the first English government document to be published in the English language after the Norman Conquest. In 1362, **Edward III** became the first king to address Parliament in English. Official documents began to be produced regularly in English during the 15th century. **Geoffrey Chaucer**, who lived in the late 14th century, is the most famous writer from the Middle English period, and The Canterbury Tales is his best-known work.

Table 4.1: Sample code-mixed questions, answers and passage. The answer(s) are shown in **bold-faced** in the passage. The mixed word from the Hindi script in the questions is highlighted using underline.

### 4.1.2 Motivation

- The recent study [230] has uncovered that users often apply question formats in their queries. They often utilize Wh-words[2] as opposed to a 'statement query'. Any multilingual speaker has a natural tendency to formulate the question format query by mixing the query words from their native language into the English query.
- There are plenty of information sources available nowadays on the web– the majority being in English. In order to leverage this information (in English), we require an effective QA model, such that for a given code-mixed question/query, it would be able to retrieve the answers from the English source documents.
- The incapability to handle code-mixed queries/questions by the search engine is one of the many factors that inspired us to develop a code-mixed question-answering system. We have shown the comparison of search engine response to the English query and code-mixed query in Fig 4.1. Here, a search engine can provide the exact answer to the question posed in the English language:

  **Question:** *"Where is the headquarter of United Nations?"*.

  However, in a code-mixed query, search engines fail to provide the exact answer to the same question:

  **Question:** *"United nations ka headquarter kaha hai?"*

  (**Translation:***"Where is the headquarter of United Nations?"*).

  It is to be noted that although both the questions are the same, the search engine is unable to return the exact answer for the code-mixed question. Instead, it returns the topmost relevant web pages.

---

[2]*who, when, where, which, why, etc.*

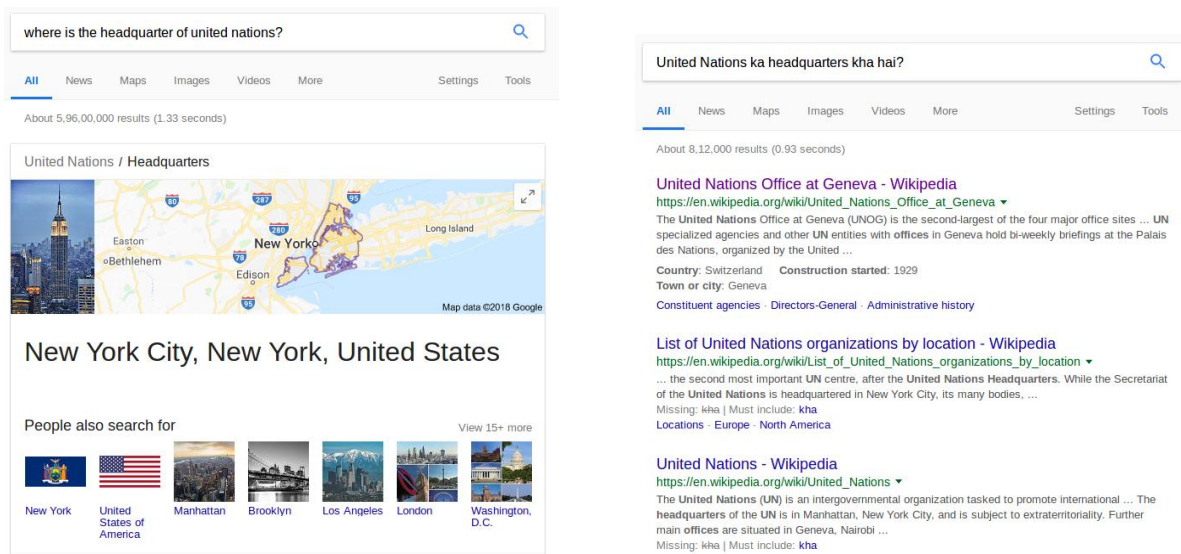

Figure 4.1: Comparison of search engine response to the English query (left) and code-mixed query (right).

- Nowadays, there is an increasing trend for virtual assistants, such as Siri[3], Cortona[4], Alexa[5], Google Assistant[6] etc. However, these search-based virtual assistants are efficient only in handling queries made in the English language. For the widespread use and accessibility of these devices, it is highly important to expand their horizon in handling code-mixed and multilingual queries.

### 4.1.3 Challenges

- **Mixed Script:** Code-mixed questions are formulated by mixing words from multiple languages; therefore, often, they do not follow any particular grammatical structure. Due to the nature of the mixed script, a strategy for processing a specific language can not be directly applied to code-mixed questions, making the problem more complex than the general QA system. A code-mixed QA system has to deal with the appropriate mechanism to capture the questions' semantics to retrieve the correct answers.
- **Lack of Code-Mixed NLP Tools:** There is a lack of efficient code-mixed NLP tools such as parts-of-speech tagger, named entity recognizer, dependency parser, etc. As the code-mixed text contains the script from multiple languages, state-of-the-art NLP tools of any particular language can not be used directly on the code-mixed text. To build code-mixed specific NLP tools focused on particular mixed languages, we need the labeled data in those scripts, which are often very limited and expensive to create.

---

[3]https://www.apple.com/in/siri/
[4]https://en.wikipedia.org/wiki/Cortana
[5]https://alexa.amazon.com/
[6]https://assistant.google.com/

- **Data Scarcity:** There is a scarcity of the code-mixed QA dataset to train the advanced deep neural networks and transformer-based models. Most of the existing code-mixed data are user-generated (either by voice-controlled devices or social-media text). It is difficult to obtain a large volume of code-mixed text from speech data as it requires transcribing with language experts. On the other hand, social media texts are profuse with misspellings, ungrammatical structures, and canonical forms of mentioning entities. Additionally, social media data are mixed with many scripts in an uneven switching manner.
- **Romanization Variability:** It is to be noted that, there is variation in a romanized script in code-mixing. For example, the Hindi word '*padhai*' (Meaning: '*study*') can be written as '*padhaai*' or '*pdhai*'. The model has to handle all those variations effectively.
- **Ambiguity**: It is also possible that the romanized and English scripts are the same but have different meanings. For example, the Hindi word '*rat*' ('*night*' in English) and English word '*rat*' (as in Animal: '*Rattus*') share the same script with different meanings.

## 4.2  Code-Mixed Question Generation

We focus on a code-mixed scenario involving two languages, *viz.* English and Hindi. Due to the scarcity of labeled data, we could not employ any sophisticated machine learning technique for question generation. Therefore, we propose an unsupervised algorithm (Algo 1) that automatically formulates the code-mixed questions. The algorithm uses several NLP components, such as PoS tagger, transliteration, and lexical translation to generate the code-mixed questions. We construct a Hindi-English code-mixed question from a given Hindi question utilizing our proposed unsupervised method. Let us consider the following three questions:

- **Q₁**: *What is the name of the baseball team in Seattle?*

- **Q₂**: सिएटल में बेसबॉल दल का नाम क्या है?

  (**Translation:** *What is the name of the baseball team in Seattle?*)

  (**Transliteration**: Seattle mai baseball dal ka naam kya hai?)

- **Q₃**: *Seattle mein baseball team ka naam kya hai?*

  (**Translation:** *What is the name of the baseball team in Seattle?*)

All three questions are the same but are asked in English, Hindi, and the code-mixed English-Hindi languages. It can be observe that **Q₂** and **Q₃** are similar and share many *false cognates [187]* [(Seattle, सिएटल), (naam, नाम), (kya, क्या), (mai, में), (baseball, बेसबॉल)]. The question **Q₃** has the direct transliteration of the Hindi words (सिएटल → Seattle), (नाम → naam), (क्या → kya), (में → mai) and (बेसबॉल → baseball). There are some words (e.g. '*team*') in **Q₃** which are the English lexical translations from Hindi. We perform a thorough study of Hindi sentences and their corresponding code-mixed Hindi-English sentences, and observe the following:

1. NEs of type person (PER) remain the same in both Hindi as well as the code-mixed English-Hindi (En-Hi) sentence. These NEs are only transliterated. For example.
   **Hindi:** महात्मा गांधी का जन्म कब हुआ था?
   (**Translation:** When was Mahatma Gandhi born?)
   **Code-Mixed (En-Hi):** *Mahatma Gandhi ka birth kab hua tha?*
   The NE *Mahatma Gandhi* of type 'PER' is transliterated in the code-mixed sentence.

2. NEs of type location (LOC) and organization (ORG) present in a Hindi sentence are replaced with their best lexical translations in English. For example:
   **Hindi:** लखनऊ दिल्ली से कितनी दूर है?
   (**Translation:** How far Lucknow is from Delhi?)
   **Code-Mixed (EN-HI):** *Lucknow New Delhi se kitni dur hai?*
   The 'PER' type NEs are only transliterated as the names do not have any variation in Hindi or English. For example, सचिन तेंदुलकर (**Translation:** Sachin Tendulkar) → 'Sachin Tendulkar'. It just needed to be written in Roman script. However, the same does not hold for 'LOC' and 'ORG'. For example, transliterating भारतीय अंतरिक्ष अनुसंधान संगठन (**Translation:** Indian Space Research Organisation) → 'Bharatiya Antriksh Anushandhan Sangathan' is incorrect.

3. PoS tags such as singular or plural noun (NN), proper noun (NNP), spatio-temporal noun (NST), and adjective (JJ) present in a Hindi sentence are often replaced with

**74**

their context-aware lexical translation in English. For example:

**Hindi:** व्यक्तियों को उनकी रचनात्मकता के लिए कौन से अधिकार दिए जाते हैं?

(**Translation:** Which rights are given to individuals for their creativity?)

**Code-Mixed (En-Hi):** *Individuals ko unki creativity ke liye koun se rights diya jate hain?*

The underlined words in the Hindi sentence have noun (NN) PoS tags. Therefore, the corresponding words are replaced with their best lexical translations in their respective code-mixed sentence.

4. The remaining words in the Hindi sentence are transliterated (in English), and a code-mixed En-Hi sentence is formed. For example, the remaining words of the previous Hindi sentence are transliterated (in underlines), and the code-mixed En-Hi sentence is formed.

**Code-Mixed (En-Hi):** *Individuals ko unki creativity ke liye koun se rights diye jate hain?*

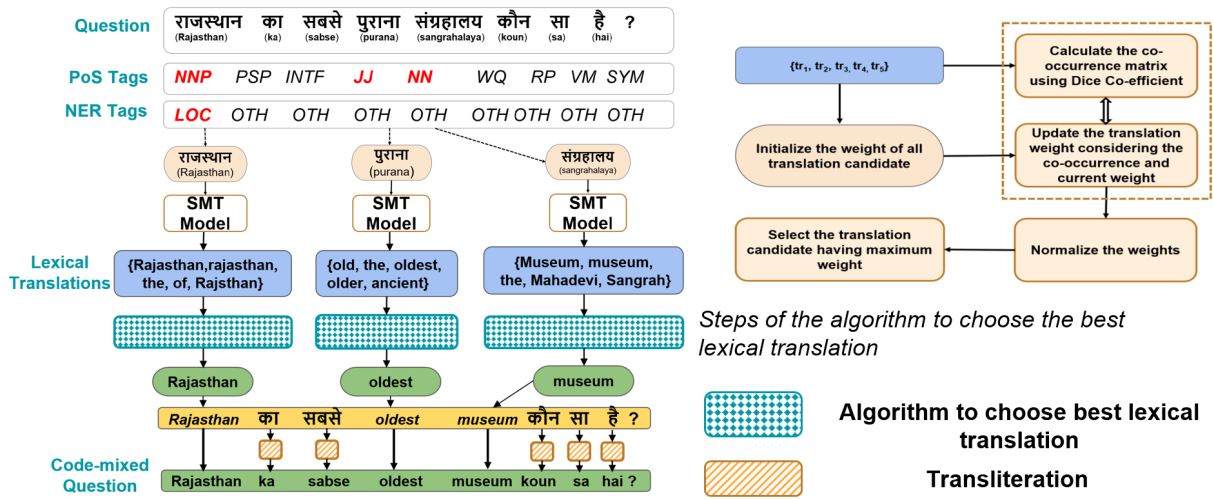

Figure 4.2: Illustrations of the proposed CMQG. The English transliterations are given in the bracket. The right part of the image shows the basic steps to select the best lexical translation. The red color tags in PoS and NE lists denote the tags of the words that qualify for the next step.

The main challenge of automatic CMQG is to identify the best lexical translation that is most appropriate in the given context of the particular question. Let us consider the various lexical translation choices $tr_i = \{tr_{(i,\ 1)}, tr_{(i,\ 2)}, \ldots, tr_{(i,\ l_i)}\}$ for the token $(t_i)$, where $l_i$ is the number of lexical translations available for the token $t_i$. The lexical trans-lation disambiguation algorithm `getBestLexicalTranslation(:,:)` (Algo 2) selects the most probable lexical translation of token $t_i$ from a set of $l_i$ possible translations. We generate a new query by adding the previous token $t_{i-1}$ and the next token $t_{i+1}$ with the token of interest $t_i$.

**Algorithm 1** Code-Mixed Question Generation
---
1: **Input**: a Hindi question ($Question$)
2: **Output**: an equivalent code-mix question ($CM-Question$)
3: **procedure** GETCODEMIXEDQUESTION($Question$)
4:     $question \leftarrow tokenize(Question)$
5:                        ▷ Tokenize the Hindi question and return the list of tokens
6:     $CMquestion \leftarrow question$
7:                            ▷ Initialize the Hindi question tokens list as CM question
8:     $pos \leftarrow getPartsOfSpeechTags(question)$
9:                                ▷ Parts-of-speech tagging of Hindi question
10:     $ner \leftarrow getNERTags(question)$
11:                           ▷ Named entity recognition tagging of Hindi question
12:     **for** $(i = 0; i < length(question); i++)$ **do**
13:                                 ▷ Looping for each token of Hindi question
14:         **if** $(ner[i] \in \{`PER'\})$ **then**
15:             $CMquestion[i] = getTransliteration(question[i])$
16:                                            ▷ Transliterate the $i^{th}$ token
17:         **else**
18:             **if** $(pos[i] \in \{`NN',`NNP',`NST',`JJ'\}$ OR $ner[i] \in \{`LOC',`ORG'\})$ **then**
19:                 $CMquestion[i] = $ `getBestLexicalTranslation`$(question[i], question)$
20:             **else**
21:                 $CMquestion[i] = getTransliteration(question[i])$
22:                                            ▷ Transliterate the $i^{th}$ token
23:     $CM-Question \leftarrow$ ' ' $.join(CMquestion)$
24:                               ▷ Join each token to form the code-mixed question
25:     **return** $CM-Question$

The context within a query also provides important clues for choosing the right transliteration for a given query word. For example, for a query $S$ ={शहर, पूर्व, स्कॉटलैंड} (**Translation:** {city, east, Scotland}), where the word 'पूर्व' is the word in interest for which the most probable lexical translation needs to be identified from the list {*BC, East*}. Here, based on the context, we can see that the choice of translation for the word 'पूर्व' is *'east'* . Since the combinations of {*city, east*} and {*east, Scotland*} are more likely to co-occur in the corpus than {*city, BC*} and {*BC, Scotland*}. We follow the iterative disambiguation algorithm (IDA) [186] based on the graph-based TextRank [178]

**Algorithm 2** Best Lexical Translation
1: **Input**: token $t$ and token list *question* from Hindi question
2: **Output**: best lexical translation $T$ for token $t$
3: **procedure** GETBESTLEXICALTRANSLATION( token $t$, token_list *question*)
4:     $i, t_i \leftarrow getTokenIndex(t, question)$     ▷ Index of the token $t$ in token list *question*
5:     $t_{i-1}, t_{i+1} \leftarrow getNeighbouringTokens(i, question)$
6:                                                         ▷ Previous and next token to token $t_i$
7:     $tr_{i-1}, tr_i, tr_{i+1} \leftarrow getLexicalTranslationSet(t_{i-1}, t_i, t_{i+1}, threshold)$
8:
9:     ▷ Lexical translations of the given token with a certain threshold on lexical translation probability
10:    $S \leftarrow \{t_{i-1}, t_i, t_{i+1}\}$                                    ▷ query terms
11:    $TR \leftarrow \{tr_{(i-1,\ 1)}, \ldots, tr_{(i-1,\ l_{i-1})}, tr_{(i,\ 1)} \ldots, tr_{(i,\ l_i)}, tr_{(i+1,\ 1)} \ldots, tr_{(i+1,\ l_{i+1})}\}$
12:    ▷ List of lexical translations for each tokens $t_{i-1}, t_i$ and $t_{i+1}$
13:    $r[0: length(TR), 0: length(TR)] \leftarrow -\infty$     ▷ Initialization of relatedness weight
14:    **for** $(x = 1;\ x \mathrel{<{=}} length(TR);\ x{+}{+})$ **do**
15:        **for** $(y = 1;\ y = length(TR);\ y{+}{+})$ **do**
16:            $a_{m,\ n} = TR[x]$                          ▷ $n^{th}$ lexical translation of $m^{th}$ token
17:            $b_{p,\ q} = TR[y]$                          ▷ $q^{th}$ lexical translation of $p^{th}$ token
18:            **if** $(m == p)$ **then** *continue*;
19:                                ▷ No action for the lexical translation of the same token
20:            **else**
21:                $r[x,\ y] = \frac{freq(TR[x],\ TR[y])}{freq(TR[x]) +\ freq(TR[y])}$
22:                                        ▷ Set co-occurrence weight using Dice Coefficient
23:    **for** *each* $tr_k \in \{tr_{i-1}, tr_i, tr_{i+1}\}$ **do**     ▷ Iterate over the lexical translation set
24:        **for** *each* $m \in\ tr_k$ **do**      ▷ Iterate over the lexical translation candidate
25:            $w_0(m|t_k) = \frac{1}{|tr_k|}$      ▷ Initialize the lexical translation with equal weight
26:            **for** $(i = 1;\ w_i(m|t_k) - w_{i-1}(m|t_k) \mathrel{<{=}} threshold;\ i{+}{+})$ **do**
27:                $w_i(m|t_k) = w_{i-1}(m|t_k) + \sum_{n\in\ connectedNode(m)} r[m, n] \times w_{i-1}(n|t)$
28:                ▷ Update the weight until the difference between two successive weights is less than the threshold
29:            **for** *each* $m \in\ tr_k$ **do**
30:                $w_i(m|t_k) = \frac{w_i(m|t_k)}{\sum_{j=1}^{|tr_k|} w_i(m_j|t_k)}$ ▷ Normalize the weight to ensure all sum up to one
31:        $m_k^* = argmax_{m \in \{tr_{(k,\ 1)}, \ldots, tr_{(k,\ l_k)}\}}(\{w_i(m|t_k)\})$
32:        ▷ Select the lexical translation candidate from the pool with maximum weight value
33:    $T = m_i^*$
34: ▷ Assign the maximum weighted lexical translation $m_i^*$ of token $t$ having index $i$ to $T$
35:    **return** $T$

---

algorithm. IDA judges a pair of items to gather partial evidence for the likelihood of a translation in a given context. Towards this, we construct an occurrence graph using the query term $S$ and the translation set $TR$, such that the translation candidates of different query terms are connected with the associated Dice Co-efficient weight between them. At the same time, it is also ensured that there should not be any edge between the different translation candidates of the same query term. We initialize each translation candidate with an equal likelihood of a translation. After initialization, each translation candidate's weight is iteratively updated using the weights of its neighboring candidates. At the end of the iteration, each translation candidate's weight is normalized to ensure that these all sum up to 1.

## 4.3 Proposed Approach for CMQA

We extend the task of QA by introducing code-mixed questions. We aim to extract the answer to code-mixed questions from the English passage. Our proposed approach for the CMQA technique consists of four major modules: **(i) Token and Sequence Encoding** to encode the Code-mixed questions and English passages using multilingual word and character embeddings followed by Bi-GRU network, **(ii) Question-aware Passage Representation** that takes input as the encoded code-mixed question and English passages and output the passage representation. This module enables the model to learn the code-mixed question-informed representation of the passage, **(iii) Context-aware Passage Representation** to capture the relevant context from the passage. It takes the question-aware passage representation as input, **(iv) Answer-type Focused Answer Extraction** is a final module of our proposed method, that brings the attention towards the answer-type of the question being asked and extracts the answer for a given code-mixed question. Each component of the model along with its description and intuition are described below:

### 4.3.1 Token and Sequence Encoding

From the given code-mixed question $Q$ and passage $P$, we first obtain the respective token-level embeddings $\{x_t^Q\}_{t=1}^m$ and $\{x_t^P\}_{t=1}^n$ from the pre-trained word embedding matrix. Due to the code-mixed nature, our model faces the out-of-vocabulary (OOV) word issue. To tackle this, we adopt character-level embedding to represent each token of the question and passage. These are denoted by $\{c_t^Q\}_{t=1}^m$ and $\{c_t^P\}_{t=1}^n$ for question and passage, respectively. The character-level embeddings are generated by taking the final hidden states of a bi-directional gated recurrent unit (Bi-GRU) [43] applied to the character embedding of the tokens. The final representations of each token $u_t^Q$ and $u_t^P$ of question and passage are obtained through the Bi-GRU as follows:

$$\begin{aligned} u_t^Q &= \text{Bi-GRU}(u_{t-1}^Q, [x_t^Q \oplus c_t^Q]) \\ u_t^P &= \text{Bi-GRU}(u_{t-1}^P, [x_t^P \oplus c_t^P]) \end{aligned} \quad (4.1)$$

where $\oplus$ is the concatenation operator. In order to encode the token sequence, we apply convolution followed by Bi-GRU operation as follows: First, the convolution operation is performed on the zero-padded sequence $\bar{u}^P$ over the passage sequence $u^P$, where $\bar{u}_t^P \in \mathbb{R}^d$. A set of $k$ filters $F \in \mathbb{R}^{k \times l \times d}$, is applied to the sequence. We obtain the convoluted features $c_t^P$ at given time $t$ for $t = 1, 2, \ldots, n$ by the following equation:

$$c_t^P = tanh(F[\bar{u}_{t-\frac{l-1}{2}}^P \ldots \bar{u}_t^P \ldots \bar{u}_{t+\frac{l-1}{2}}^P]) \quad (4.2)$$

The feature vector $\bar{C}^P = [\bar{c}_1^P, \bar{c}_2^P \ldots \bar{c}_n^P]$ is generated by applying the max pooling on each element $c_t^P$ of $C^P$. This sequence of convolution feature vector $\bar{C}^P$ is then passed through a Bi-GRU network, which captures the sequence information across the time steps. The same convolution operations are also performed over the question sequence $u^Q$ and the convolution feature vector $\bar{C}^Q$ is obtained. Similar to e.q. 4.1, we compute Bi-GRU outputs $v_t^P$ ($v_t^Q$) by giving the inputs $v_{t-1}^P$ ($v_{t-1}^Q$) and $\bar{c}_t^P$ ($\bar{c}_t^Q$). We represent the question and passage representation matrix by $V^Q \in \mathbb{R}^{m \times h}$ and $V^P \in \mathbb{R}^{n \times h}$, respectively, where $h$ is the number of hidden units of the Bi-GRUs.

### 4.3.2 Question-aware Passage Representation

When a single passage contains the answer of multiple different questions, then the passage encoding, obtained from the previous layer (c.f. section 4.3.1) will not be sufficient enough to provide the answer of each question. It is because the obtained passage encoding does not take into account the question of information. Towards this, first, we compute an attention matrix $M \in \mathbb{R}^{n \times m}$ as follows:

$$M_{i,\,j} = 1/(1 + dist(V^P[i,:],\ V^Q[j,:])) \quad (4.3)$$

Here, $M_{i,\,j}$ is the similarity score between the $i^{th}$ element of the passage encoding $V^P$ and $j^{th}$ element of the question encoding $V^Q$. The $dist(x, y)$ function is a euclidean distance[7] between $x$ and $y$. Thereafter, the normalization of an element $M_{i,j}$ of matrix $M$ is performed with respect to the $i^th$ row as following:

$$\overline{M}_{i,\ j} = \frac{M_{i,\ j}}{\sum_{k=0}^{m} M_{i,\ k}} \tag{4.4}$$

Intuitively, it computes the relevance of a word in the given passage with each word in the question. We compute the question vector $Q \in \mathbb{R}^{n \times h}$ corresponding to all the words in the passage as $Q = \overline{M} \times V^Q$. Each row $t$ of the question vector $Q$ denotes the encoding of the passage word $t$ with respect to all the words in the question. The question-aware passage encoding will be computed by the word-level concatenation of the passage encoding $v_t^P$ and question vector of the $t^{th}$ row $Q_t$. More formally, the question aware passage encoding $a_t$ of the word at time $t$ will be $a_t = v_t^P \oplus Q_t$. Finally, we apply a Bi-GRU to encode the question of aware information over time. It is computed as follows:

$$s_t = \text{Bi-GRU}(s_{t-1}, a_t) \tag{4.5}$$

We can represent the question aware passage encoding matrix as $S \in \mathbb{R}^{n \times h}$.

### 4.3.3 Context-aware Passage Representation

Question-aware passage encoding accounts for the relevance of the words in a question with the given passage. If the answer spans more than one token (i.e., multi-word tokens), it is crucial to compute the relevance between the multi-word token's constituents. We encounter this situation by introducing bi-linear attention. We calculate the bilinear attention matrix $B \in \mathbb{R}^{n \times n}$ on question aware passage encoding $S \in \mathbb{R}^{n \times h}$ as follows:

$$B = S \times \mathbf{W_b} \times S^T \tag{4.6}$$

where, $\mathbf{W_b} \in \mathbb{R}^{h \times h}$ is a bilinear weight matrix. Similar to e.q. 4.4, normalization is performed on $B$, and the normalized attention matrix is denoted as $\overline{B}$. The element $\overline{B}_{i,j}$ is the measure of relevance between the $i^{th}$ and $j^{th}$ words of the passage. Similar to the question vector $Q$, we calculate the passage vector $R \in \mathbb{R}^{n \times h}$ as computed on $R = \overline{B} \times S$. The concatenation (word wise) of question-dependent passage encoding vector $s_t$ and passage vector $r_t$ is performed to obtain $\overline{R_t}$ and form the matrix $\overline{R} \in \mathbb{R}^{n \times 2h}$.

---

[7]We observe that e.q. 4.3 performs well, when $dist$ is an euclidean distance.

To determine the relevant spans of passage and attend to the ones relevant to the given question, we introduce the gating mechanism [296] with Sigmoid unit and recompute the passage representation as $G \in \mathbb{R}^{n \in 2h}$. In order to identify the start and end indices of the answer from the passage, we employ two Bi-GRUs with input as $G$, and the output of the Bi-GRUs is computed as $P_s \in \mathbb{R}^{n \times h}$ and $P_e \in \mathbb{R}^{n \times h}$.

### 4.3.4 Answer-type Focused Answer Extraction

The answer-type of a question provides clues to detect the correct answer from the passage. Consider a code-mixed question

**Question:** *Kaun sa Portuguese player, Spanish club Real Madrid ke liye as a forward player khelta hai?*

(**Translation:** *Which Portuguese player plays as a forward for Spanish club Real Madrid?.*)
The answer-type of the question $Q$ is 'person'. Even though the network has the capacity to capture this information up to a certain degree, it would be better if the model takes into account this information in advance while selecting the answer span. [155] proposed a hierarchical question classification based on the answer-type of a question. Based on the coarse and fine classes of [155], we train two separate answer-type detection networks on the Text REtrieval Conference (TREC) question classification dataset[8]. First, we translate[9] 5952 TREC English questions into Hindi and after that transform the Hindi questions into the code-mixed questions by using our proposed CMQG algorithm. We train the answer-type detection network with code-mixed questions and their associated labels using the question classification technique, as discussed in Section 3.3.2. The network learns the encoding of coarse ($C_{at} \in \mathbb{R}^h$) and fine class ($F_{at} \in \mathbb{R}^h$) of answer-types obtained from the answer-type detection network. The attention matrix $M$ calculated in e.q. 4.3 undergoes the max-pooling over the columns to capture the most relevant parts of the question.

$$Q_p^j = \text{max-pool}(M[:, j]) \tag{4.7}$$

The max-pooled representation of question and answer-type representation are concatenated in the following way:

$$Q_f = Q_p.V^P \oplus C_{at} \oplus F_{at} \tag{4.8}$$

---

[8] http://cogcomp.org/Data/QA/QC/
[9] We use Google Translate because of its better performance on EN → HI translation.

A feed-forward neural network with *tanh* activation function is used to obtain the final output $\overline{Q}_f \in \mathbb{R}^h$. The probability distribution of the beginning of answer $A_s$ and the end of answer $A_e$ is computed as:

$$\begin{aligned} prob(A_s) &= \text{softmax}(\overline{Q}_f \times P_s) \\ prob(A_e) &= \text{softmax}(\overline{Q}_f \times P_e) \end{aligned} \quad (4.9)$$

To train the network, we minimize the sum of the negative log probabilities of the ground truth start and end position by the predicted probability distributions.

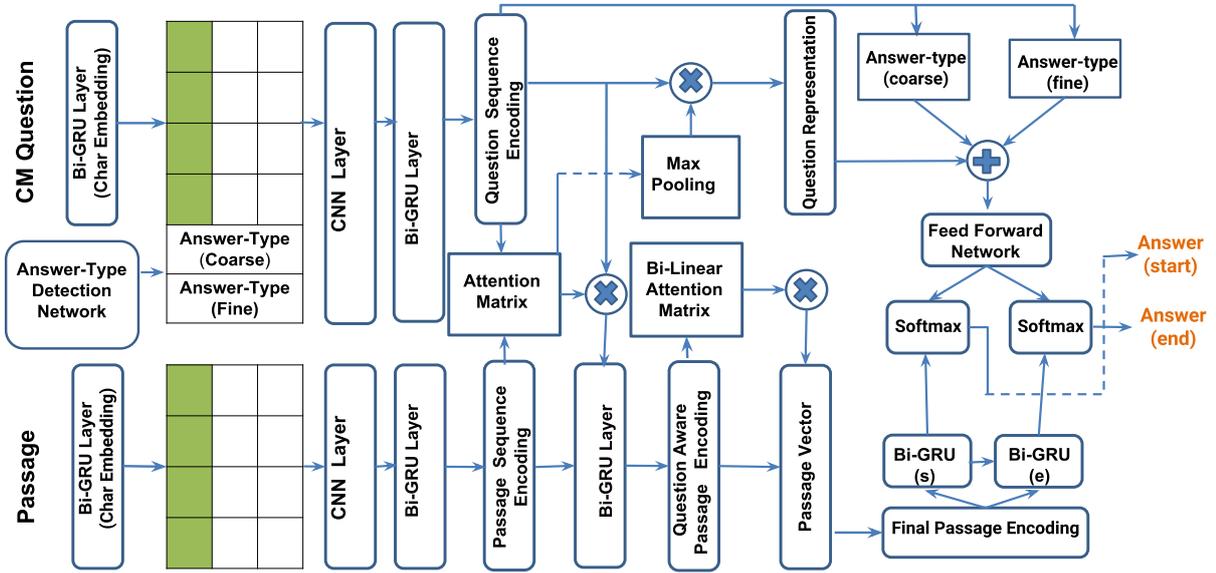

Figure 4.3: Proposed CMQA model architecture. The green color column denotes the character embeddings.

## 4.4 Datasets and Experimental Details

In this section, we report the datasets and the experimental setups. The details are as follows:

### 4.4.1 Datasets for CMQG

For the CMQG task, we require the input question to be in Hindi. We use our created Hindi-English question-answering dataset MMQA (Section 3.2) to generate the code-mixed questions by our proposed approach (c.f. Section 4.2). In order to evaluate the performance of our proposed CMQG algorithm, we also manually formulate[10] the Hindi-English code-mixed questions. Details of this dataset are shown in Table 4.2. We compute the complexity of code-mixing using the Code-mixing Index (CMI) score [70] metric. We name this code-mixed question dataset as '*HinglishQue*'. We observe that our *HinglishQue* dataset has a higher CMI score over the FIRE[11] 2015 (CMI=11.65) and ICON[12] 2015 (5.73) CM corpus [267][13]. This implies that our *HinglishQue* dataset is more complex and challenging compared to the other Hindi-English code-mixing (CM) dataset. The CMI score of the system-generated code-mixed questions is 37.22.

| # of CM Questions | 5,535 | # of Hindi Words | 37,300 |
|---|---|---|---|
| # of words | 59,733 | Average # of Hindi Words/Question | 6.7389 |
| Average Length of CM Questions | 10.79 | # of English Words | 22,433 |
| Code-Mixing Index (CMI) Score | 37.14 | Average # of English Words/Question | 4.05 |

Table 4.2: Statistics of manually formulated CM questions

| Datasets | Train | Dev | Test | Total |
|---|---|---|---|---|
| CM-SQuAD | 16,632 | 2,080 | 2,080 | 20,792 |
| CM-MMQA | 2,746 | 341 | 341 | 3,428 |

Table 4.3: Detailed statistics (# of question-passage pairs) of the derived CMQA datasets

### 4.4.2 Datasets for CMQA

1. **CM-SQuAD:** We generate the CMQA dataset from the portion of the SQuAD [216] dataset. We translate the English questions into Hindi and use our approach of CMQG (c.f. Section 4.2) to transform the Hindi questions into code-mixed questions. We manually verify the questions to ensure quality. We use the corresponding English passage to find the answer pair of the code-mixed questions. Detailed statistics of the dataset are shown in Table 4.3. We randomly split the dataset into training, development, and test set.

2. **CM-MMQA:** We experiment with our multilingual QA dataset (c.f. Section 3.2). Similar to the CM-SQuAD dataset, we create code-mixed questions and their respective answer. Details of the dataset are shown in Table 4.3.

---

[10]The question formulators are the undergraduate and postgraduate students having good proficiencies in English and Hindi.
[11]http://fire.irsi.res.in/fire/2015/home
[12]http://ltrc.iiit.ac.in/icon2015/
[13]Please note that these two datasets are not related to QA

### 4.4.3 Experimental Setup for CMQG

The tokenization and PoS tagging are performed using the publicly available Hindi Shallow Parser[14]. The Polyglot[15] Named Entity Recognizer (NER) [5] is used for named entity recognition. The lexical translation set is obtained by the lexical translation table generated as an intermediate output of Statistical Machine Translation (SMT) training by Moses [141] on publicly available[16] English-Hindi (En-Hi) parallel corpus [23]. We aggregate the output probability $p(e|h)$ and inverse probability $p(h|e)$ along with their associated words in both English (e) and Hindi (h) languages. We choose a threshold (5) to filter out the least probable translations. The co-occurrence weight (Dice Co-efficient) is calculated on the available[17] n-gram dataset consisting of unique $2,86,358$ bigrams and $3,33,333$ unigrams. For Devanagari (Hindi) to Roman (English) transliteration, we use the transliteration system[18] based on [62]. We evaluate the performance of CMQG in terms of accuracy, BLEU [200] and ROUGE [160] score.

### 4.4.4 Experimental Setup for CMQA

CMQA datasets contain the words both in Roman script and English. For English, we use the *fastText* [22] word embedding of dimension 300. We use the Hindi sentences from [23], and then transliterate it into the Roman script. These sentences are used to train the word embeddings of dimension 300 by the word embedding algorithm [22]. Finally, we align monolingual vectors of English and Roman words in a unified vector space using a linear transformation matrix learned by the approach, as discussed in [255]. Other optimal hyperparameters are set to: character embedding dimension=50, GRU hidden unit size=150, CNN filter size=150, filter size=3, 4, batch size=60, # of epochs=100, initial learning rate=0.001. The optimal values of the hyperparameters are decided based on the model performance on the development set of CM-SQuAD dataset. Adam optimizer [136] is used to optimize the weights during training. For the evaluation of CMQA, we adopt the EM and F1-score [216].

---

[14]http://ltrc.iiit.ac.in/showfile.php?filename=downloads/shallow_parser.php
[15]http://polyglot.readthedocs.io/en/latest/NamedEntityRecognition.html
[16]http://ufal.mff.cuni.cz/hindencorp
[17]http://norvig.com/ngrams/
[18]https://github.com/libindic/indic-trans

| Datasets → | CM-SQuAD (1) | | | | | | CM-MMQA (2) | | | |
|---|---|---|---|---|---|---|---|---|---|---|
| | Dev | | Test | | Test (2) | | Dev | | Test | |
| Models | EM | F1 | EM | F1 | EM | F1 | EM | F1 | EM | F1 |
| IR [31] | 5.82 | 9.51 | 5.02 | 8.92 | - | - | 5.52 | 9.66 | 6.10 | 10.64 |
| BiDAF [240] | 21.44 | 29.18 | 21.63 | 28.45 | 22.26 | 37.54 | 22.38 | 33.10 | 22.09 | 32.82 |
| R-Net [296] | 24.17 | 31.12 | 23.76 | 30.74 | 24.47 | 39.15 | 24.27 | 37.33 | 23.72 | 36.86 |
| Proposed Approach | **31.12** | **37.78** | **31.05** | **36.97** | **30.91** | **46.18** | **28.14** | **46.25** | **30.56** | **46.10** |

Table 4.4: Performance comparison of the proposed CMQA algorithm with the IR-based and neural-based baselines. **Test (2)** refers the test set of CM-MMQA.

| Models | Accuracy | Bleu | ROUGE-1 | ROUGE-2 | ROUGE-L |
|---|---|---|---|---|---|
| Seq2Seq | 39.24 | 52.18 | 53.28 | 56.05 | 52.11 |
| Proposed Algorithm | 67.11 | 86.17 | 95.15 | 90.53 | 95.13 |

Table 4.5: Performance comparison of the proposed CMQG algorithm with *seq2seq* baseline.

## 4.5 Results and Analysis

### 4.5.1 Baselines

**Baselines (CMQG):** We portray the problem of code-mixed question generation as the sequence to sequence (seq2seq) learning task where the input sequence comprises of Hindi question, and the output sequence is the code-mixed En-Hi question. Our first baseline is the seq2seq model equipped with attention mechanism [273, 13]. The model is trained using the default parameters of Nematus [237]. The training dataset consists of a pair of Hindi-translated questions, and code-mixed questions from the CM-SQuAD dataset (c.f. Section 4.4.2). For the final evaluation of the model, we use the manually created CMQG dataset (c.f. Section 4.4.1).

**Baselines (CMQA):** To compare the performance of our proposed CMQA model, we define the following baseline models.

1. **IR based model:** This baseline is our implementation of the *WebShodh* [31] with improvements in some existing components. We replaced *WebShodh*'s support vector machine (SVM) based question classification with our recurrent CNN-based answer-type detection network (c.f. Section 4.3.4). Despite searching for the answer on the web (as *WebShodh* does), we search for it within the passage. We choose the highest-ranked answer as our final answer.

2. **R-Net [296]:** This is a deep neural network-based reading comprehension (RC) model. We train the R-Net model with the hyperparameters as described in [296].

3. **BiDAF [240]:** This is another state-of-the-art neural model for RC. We trained this model with the same hyperparameters as given in [240].

### 4.5.2 Experimental Results

We demonstrate the evaluation results of our proposed CMQG algorithm on the *HinglishQue* dataset in Table 4.5. For evaluation, we employed three annotators who were instructed to assign the label (*same* or *different*) depending upon whether the system-generated and manually created questions are similar or dissimilar. The agreement among the annotators was calculated by Cohen's Kappa [45] coefficient, and it was found to be 92.45%. Evaluation of question generation shows that our proposed CMQG algorithm performs better than the seq2seq-based baseline. One reason could be the insufficient amount (16,632) of training instances and the out-of-vocabulary (only 62.35% words available in the vocab) issue. Performance improvement in our proposed model over the baseline is statistically significant as $p < 0.05$. In the literature, we find only one study on English-Hindi code-mixed question classification i.e., [215]. They used only 1,000 code-mixed questions and used a Support Vector Machine (SVM) to classify the questions into coarse and fine-grained answer types. They reported 63% and 45% accuracies for coarse and fine-grained answer-type detection, respectively, under a 5-fold cross-validation setup. In contrast, we manually create 5,535 code-mixed questions and train a CNN model that shows 87.21% and 83.56% accuracies for coarse and fine answer types, respectively, for the 5-fold cross-validation.

| Model | CM-SQuAD | | CM-MMQA | |
|---|---|---|---|---|
| Components | EM | F1 | EM | F1 |
| Proposed | 31.12 | 37.78 | 28.14 | 46.25 |
| (-) Convolution | 29.46 | 36.14 | 26.19 | 43.76 |
| (-) Bilinear Attention | 26.42 | 33.31 | 25.36 | 41.29 |
| (-) Answer-type Focused | 28.41 | 35.14 | 26.69 | 42.37 |

Table 4.6: Effect of the various components of the CMQA model on the development set of CM-SQuAD and CM-MMQA dataset. **(-) X** denotes the model architecture after removal of '**X**'.

Results of CMQA for both datasets are shown in Table 4.4. The performance of IR-based baseline [31] on both datasets is poor. This may be because [31]'s system was mainly developed to answer simple factoid questions based only on the named entities denoting person, location, and organization. However, the datasets used in this experiment have different types of answers beyond the basic factoid questions. We also perform a cross-domain experiment, where the test data of CM-MMQA is used to evaluate the system trained on CM-SQuAD. Performance improvements in our proposed model over the baselines are statistically significant as $p < 0.05$. Experiments show that the performance of CM-MMQA is better than CM-SQuAD. This might be due to the relatively smaller length passages in CM-MMQA, extracting answers from which are easier.

| Sr. No. | Reference Questions | System Generated Questions |
|---|---|---|
| 1 | Maharaja Ranjit Singh ne Mandi par kab **occupy** kar liya tha? | Maharaja Ranjit Singh ne Mandi par kab *Czechoslovakia* kar liya tha? |
| 2 | Babur kaa **death** kab ho gaya tha? | Babur kaa *died* kab ho gaya tha? |
| 3 | **IMF** kaa primary purpose kya hai? | *Imef* kaa primary purpose kya hai? |
| 4 | **Demographics** kya hai? | *Population* kya hai? |

Table 4.7: Some examples from the *HinglishQue* dataset depicting the errors occurred. The **correct** and *incorrect* words in the questions are denoted with **bold** and *italic* fonts, respectively.

We perform an ablation study to observe the effects of various components of the CMQA model. Results are shown in Table 4.6. The component convolution refers to the convolution operation performed before the Bi-GRUs in sequence encoding.

### 4.5.3 Error Analysis

We analyze the errors encountered by our CMQG and CMQA systems. The CMQG algorithm uses several NLP components such as PoS tagger, NE tagger, translation, and transliteration. The errors that occurred in these components propagate toward the final question generation. We list some of the major causes of errors with examples in Table 4.7. As in **(1)**, the algorithm could not find the correct lexical translation from the lexical table itself and therefore selected an irrelevant word 'Czechoslovakia' instead of 'occupy'. In **(2)** and **(4)**, the algorithm picked the words 'died' and 'population' instead of 'death' and 'demographics', respectively. It is because the word 'died' and 'population' have higher n-gram frequencies than the words 'death' and 'demographics' in the n-gram corpus. In **(3)**, the system generated an incorrect word ('imef') instead of 'IMF'. Here, the Hindi word 'आईएमएफ' is incorrectly tagged as 'Other' instead of 'Organization'. Therefore, the transliteration system provides an incorrect transliteration ('imef') of the abbreviated

Hindi word 'आईएमएफ' (**Translation**: IMF).

We observe that our CMQA system sometimes incorrectly predicts the answer words that are very close to some other word in the shared embedding space (c.f. section 4.4.4), and hence gets a high attention score in the bilinear attention module. For example, in this passage '...*India* was ruled by the **Bharata** clan and ...', the system predicted the answer 'India' instead of 'Bharata' (reference answer) because the word 'Bharata' is the transliteration form of भारत and भारत is the correct translation form of the word 'India'.

Our close analysis of the prediction of CM-SQuAD and CM-MMQA development data reveals that the system mostly suffers from errors, where the answer strings are relatively longer. We found that the CM-MMQA dataset has some definitional questions (requires at least one sentence long answer). Therefore, we evaluated the performance on the CM-MMQA dataset after removing those questions (92) and obtaining the EM and F1 scores of 40.50% and 53.73%, respectively. These are much higher (28.14%, 46.25%) than the model where all the questions are considered. Due to ambiguity in selecting answers (between two candidate answers, location type answer), the system sometimes predicts incorrectly. We also observed some other types of errors, which were mainly due to the context mismatch and long-distance dependence between the answer and the context words.

## 4.6 Conclusion

In this chapter, we have discussed the proposed linguistically motivated unsupervised algorithm for CMQG and the neural framework for CMQA. Specifically, we have developed bilinear attention and answer-type focused neural architecture to deal with CMQA. We have evaluated the performance of CMQG on manually created code-mixed questions involving English and Hindi. For CMQA, we have created two CMQA datasets: CM-MMQA and CM-SQuAD to enable end-to-end supervised training for code-mixed question answering task. Experiments show that our proposed models attain state-of-the-art performance on both datasets. The proposed approach achieves the EM and F1 scores of 31.05 and 36.97, respectively, for the CM-SQuAD dataset and 30.56 and 46.10, respectively, for the CM-MMQA dataset.

In the next chapter, we will extend our study of MQA on visual question answering. We will discuss the limitation and challenges of the existing VQA model on the multilin-gual setup. Furthermore, we will also devise a way to mitigate the existing challenges and propose a solution to build the VQA model in the code-mixed and multilingual setup.

# Chapter 5

# Multilingual and Code-Mixed VQA

## 5.1 Introduction

In the previous chapter, we propose and discuss the task of code-mixed question answering. Toward that, we create the dataset based on the linguistic theory of code-mixing. In addition to that, we introduce the neural network-based solution to extract the answer for code-mixed questions. This chapter focus on solving the issue of multilingual and code-mixed QA from textual and visual documents. We assume that there are several techniques available for VQA in a monolingual (mostly English) setting. However, we are interested in building a system that can answer questions from different languages (multilingual) and the language formed by mixing multiple languages (code-mixed). Toward this, we create a dataset for VQA in a multilingual and code-mixed environment. After that, we utilize the dataset to propose a unified model that can handle natural language questions from multiple languages to find the answer from a given image.

Multilingualism and Code-mixing are common in chats, conversations, and messages posted over social media, especially in bilingual/multilingual countries like India, China, Singapore, and most other European countries. Sectors like tourism, food, education, and marketing, etc. have recently started using code-mixed languages in their advertisements to attract their consumer base. In order to build an Artificial Intelligent (AI) agent which can serve multilingual end users, a VQA system should be put in place that would be language-agnostic and tailored to deal with the code-mixed and multilingual environment. It is worth studying the VQA system in these settings, which would be immensely useful to a very large number of the population who speak/write in more than one language.

A recent study [201] also shows the popularity of the Hinglish (code-mixed Hindi and English) language and the dynamics of language shift in India.

We show that in a cross-lingual scenario due to language mismatch, applying a learned system directly from one language to another language results in poor performance. Thus, we propose a technique for multilingual and code-mixed VQA. Our proposed method mainly consists of three components. The first is the *multilingual question encoding*, which transforms a given question into its feature representation. This component handles multilingualism and code-mixing in questions. We use multilingual embedding coupled with a hierarchy of shared layers to encode the questions. To do so, we employ an attention mechanism on the shared layers to learn language-specific question representation. Furthermore, we utilize self-attention to obtain an improved question representation by considering the other words in the question. The second component (*image features*) is to obtain an effective image representation from object-level and pixel-level features. The last component is *multimodal fusion*, which is accountable to encode the question-image pair representation by ensuring that the learned representation is tightly coupled with both the question (language) and image (vision) feature.

It is to be noted that designing a VQA system for each language separately is computationally very expensive (both time and cost), especially when multiple languages are involved. Hence, an end-to-end model that integrates multilinguality and code-mixing in its components is advantageous.

### 5.1.1 Problem Statement

Given a natural language question $\mathcal{Q}$ in English, Hindi, or code-mixed and a correlative image $\mathcal{I}$, the task is to perform complex reasoning over the visual elements of the image to provide an accurate natural language answer $\hat{\mathcal{A}}$ from all the possible answers $\mathcal{A}$. Formally:

$$\hat{\mathcal{A}} = \underset{\hat{\mathcal{A}} \in \mathcal{A}}{\arg\max}\, p(\hat{\mathcal{A}}|\mathcal{Q}, \mathcal{I}; \phi) \tag{5.1}$$

An example of a multilingual and code-mixed VQA system is shown in Fig 5.1.

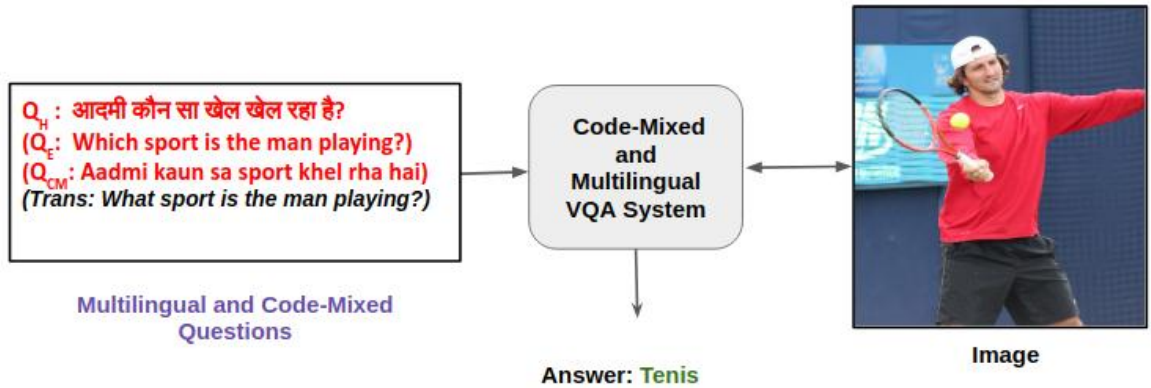

Figure 5.1: Example of multilingual and code-mixed VQA system where the system takes multilingual questions and interact with the image to retrieve the answer. $Q_H$: question in Hindi, $Q_E$: question in English and $Q_{CM}$: question in English-Hindi code-mixed.

### 5.1.2 Motivation

- As the existing research on VQA is mainly focused on natural language questions written in English [12, 114, 69, 9, 153, 309, 247], their applications are often limited to the English speaker.
- Code-mixing phenomena are common in chats, conversations, and messages posted over social media, especially in bilingual/multilingual countries like India, China, Singapore, and most other European countries. Sectors like tourism, food, education, and marketing, etc. have recently started using code-mixed languages in their advertisements to attract their consumer base. In order to build an AI agent which can serve multilingual end users, a VQA system should be put in place that would be language-agnostic and tailored to deal with the code-mixed and multilingual environment. It is worth studying the VQA system in these settings, which would be immensely useful to a vast population of people who speak/write in more than one language.
- The existing VQA systems lack generalizability and therefore have shown poor performance in code-mixed and multilingual settings. Let us consider the examples shown in Fig 5.2. The majority of the VQA models [9, 153, 335] would be capable enough to provide the correct answers for English questions **Q<sub>E</sub>**, but our evaluation shows that the same model could not predict the correct answers for Hindi **Q<sub>H</sub>** and Code-mixed question **Q<sub>CM</sub>**. The questions **Q<sub>H</sub>** and **Q<sub>CM</sub>** correspond to the same

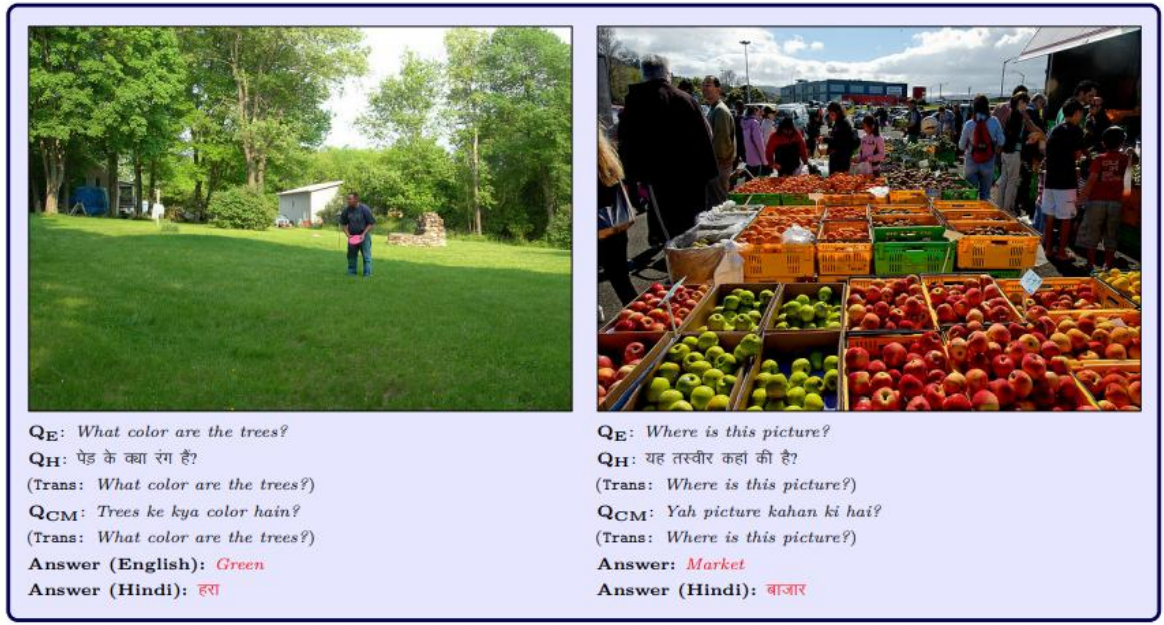

Figure 5.2: Examples of questions (English, Hindi and Code-mixed) with their corresponding images and answers

question $Q_E$, but are formulated in two different languages.

- **Easy to interact:** The code-mixed enables the system (QA or VQA) is easy to interact with the end-user with a limited understanding of the English language. There is a variety of much-needed virtual assistants (Health care, Transportation query, etc.) that can be offered through code-mixed QA.

### 5.1.3 Challenges

- **Code-Mixed Dataset:** The existing datasets [174, 71, 351, 12, 79] in VQA is only limited to English questions. It is necessary to have the dataset covering multilingual and code-mixed questions to facilitate learning in VQA, irrespective of language.
- **Complexity of Multilingual Questions:** Multilingual questions differ from each other on various fronts, such as morphologically, syntactically, etc. These make the task of question encoder difficult. The question encoder should be capable enough to encode multilingual and code-mixed question variations without losing its semantics.
- **Language and Vision Fusion:** This is one of the crucial challenges in VQA where the language feature from a question and vision feature from the image are amalgamated to generate a joint representation of the language-vision feature. It is

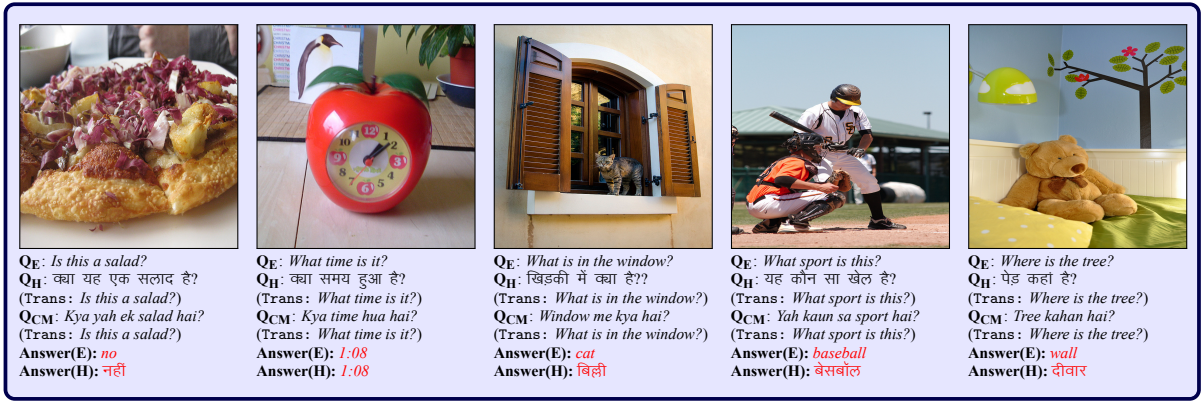

Figure 5.3: Sample questions (in English, Hindi, and Code-Mixed) with their corresponding images and answers (in English, Hindi) from our `MCVQA` dataset

essential to have an efficient mechanism to fuse the language and vision features.

## 5.2 `MCVQA` Dataset

### 5.2.1 Dataset Creation

The popular VQA dataset released by [12] contains images, with their corresponding questions (in English) and answers (in English). This is a challenging large-scale dataset for the VQA task. To create a comparable version of this English VQA dataset in Hindi and code-mixed Hinglish languages, we introduce a new VQA dataset named "**M**ultilingual and **C**ode-mixed **V**isual **Q**uestion **A**nswering" (`MCVQA`) which comprises of questions in Hindi and Hinglish. Our dataset, in addition to the original English questions, also presents the questions in Hindi and Hinglish languages. This makes our `MCVQA` dataset suitable for multilingual and code-mixed VQA tasks. A sample of question-answer pairs and images from our dataset are shown in Figure 5.3.

We do not construct the answer in code-mixed language because our study (Chapter 4) shown that code-mixed sentences and their corresponding English sentences share the same nouns (common nouns, proper nouns, Spatio-temporal nouns), adjectives, etc. For example, given an English and its corresponding code-mixed question:

$Q_E$ : *Where is the **tree** in this **picture**?*

$Q_{CM}$ : *Is **picture** me **tree** kahan hai?*

It can be observed that both $Q_E$ and $Q_{CM}$ share the same noun { ***picture***, ***tree***}. The majority of answers in the VQA v1.0 dataset are of type '*yes/no*', '*numbers*', '*nouns*',

'*verbs*' and '*adjectives*'. Therefore, we keep the same answer in both the English and Code-Mixed VQA dataset.

We follow the techniques proposed in Section 4.2 for code-mixed question generation, which takes a Hindi sentence as input and generates the corresponding Hinglish sentence in the output. We translate original English questions and answers using the Google Translate[1] that has shown remarkable performance in translating short sentences [306]. We use the Google Translate service as our original questions and answers in English are very short. For the code-mixed question generation, we first obtain the POS[2] and Named Entity[3] tags of each question. Thereafter, we replace the Hindi words with the PoS tags (common noun, proper noun, Spatio-temporal noun, adjective) with their best lexical translations. The same strategy is also followed for the words having the NE tags as *LOCATION* and *ORGANIZATION*. The remaining Hindi words are replaced with their Roman transliteration. In order to obtain the best lexical translation, we follow the iterative disambiguation algorithm [186]. We generate the lexical translation by training the Statistical Machine Translation (SMT) model on the publicly available English-Hindi (En-Hi) parallel corpus [23]. We also present the MCVQA comparison with other VQA datasets in Table 5.1.

| Dataset | Images used | Created by | Multilingual | Code-Mixed |
|---|---|---|---|---|
| DAQUAR [174] | NYU Depth V2 | In-house participants Automatically generated | ✗ | ✗ |
| FM-IQA [71] | MSCOCO | Crowd workers (Baidu) | ✔ | ✗ |
| VQA v1.0 [12] | MSCOCO | Crowd workers (AMT) | ✗ | ✗ |
| Visual7W [351] | MSCOCO | Crowd workers (AMT) | ✗ | ✗ |
| CLEVR [119] | Synthetic Shapes | Automatically generated | ✗ | ✗ |
| KB-VQA [291] | MSCOCO | In-house participants | ✗ | ✗ |
| FVQA [292] | MSCOCO | In-house participants | ✗ | ✗ |
| Japanese VQA [248] | MSCOCO | Crowd workers (Yahoo) | ✔ | ✗ |
| **MCVQA (Ours)** | **MSCOCO** | **Automatically generated** | ✔ | ✔ |

Table 5.1: Comparison of VQA datasets with our `MCVQA` dataset. The images used are: MSCOCO [161] and NYU Depth v2 [251]
.

---

[1] https://cloud.google.com/translate
[2] https://bit.ly/2rpNBJR
[3] https://bit.ly/2Qljan5

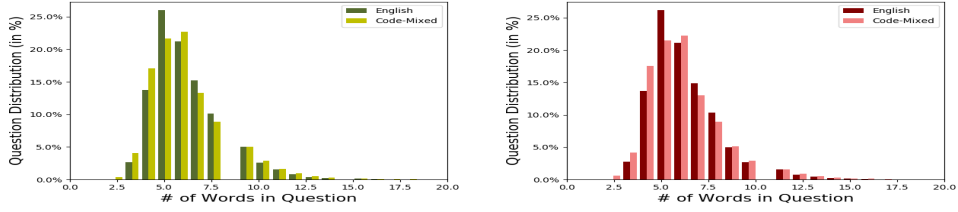

Figure 5.4: Analysis to show the comparison of question distribution w.r.t the question length between VQA v1.0 English and code-mixed, train and test dataset.

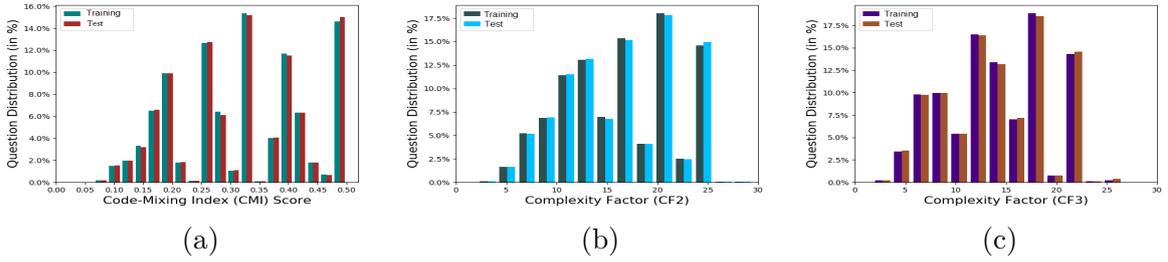

(a)          (b)          (c)

Figure 5.5: Analysis of code-mixed VQA dataset on various code-mixing metrics: **(a)**, **(b)** and **(c)** show the distribution of code-mixed questions from training and test set w.r.t to the CMI score, Complexity Factor, CF-2 and CF-3, respectively.

### 5.2.2 Dataset Analysis:

The `MCVQA` dataset consists of 248,349 training questions and 121,512 test questions for real images in Hindi and Code-Mixed languages. For each Hindi question, we also provide its 10 corresponding answers in Hindi. In order to analyze the complexity of the generated code-mixed questions, we compute the CMI and Complexity Factor (CF) [74]. These metrics indicate the level of language mixing in the questions. A details distribution of the generated code-mixed questions with respect to various metrics is shown in Figure 5.4, 5.5.

We perform qualitative analysis by randomly selecting 5,200 questions from our `MCVQA` dataset. A bilingual (EN, HI) expert was asked to manually create the code-mixed questions and translate the English questions into Hindi. We compute the BLEU [200], ROUGE [160] and Translation Error Rate (TER) [256] on the human-translated questions and the translations obtained from Google Translate. We achieve high BLEU and Rouge scores (BLEU 3: 80.22; ROUGE - L: 92.20) and lower TER (9.63). Similarly, we compute the complexity factor of automatically generated code-mixed questions and found these to be highly comparable to the ground truths.

## 5.3 Methodology for MVQA

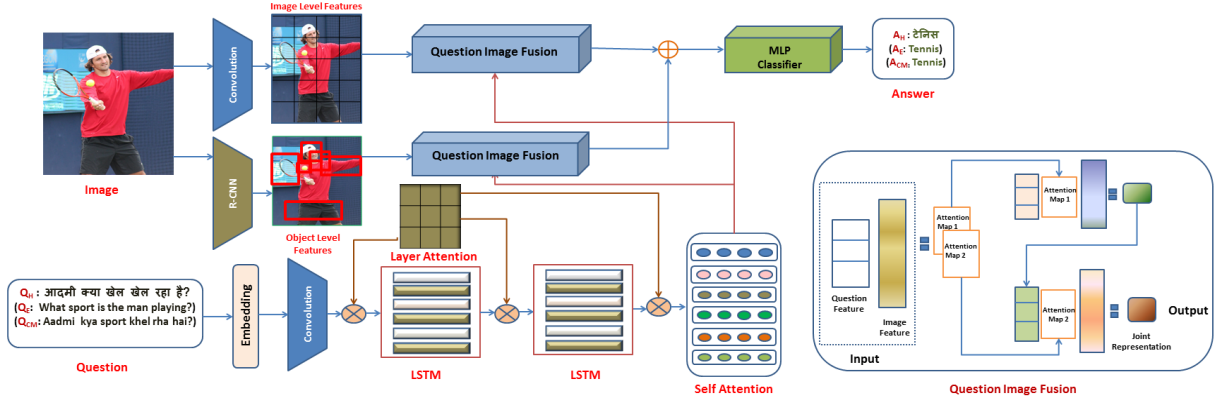

Figure 5.6: Architecture of the proposed multilingual VQA model. The input to the model is the multilingual question (one at a time). The bottom-right part of the image describes the *Question Image Fusion* component.

The architecture of our proposed methodology is depicted in Fig 5.6. Our proposed model has the following components:

### 5.3.1 Multilingual Question Encoding

Given a question[4] $Q = \{q_1, q_2, \ldots, q_T\}$ having $T$ words, we obtain the multilingual embedding $q_t^e \in \mathbb{R}^d$ (c.f. Section 5.4.1) for each word $q_t \in Q$. The resulting representation is denoted by $\{q_t^e\}_{t=1}^T$. We use multilingual word embedding to obtain lower-level representations of the words from English, Hindi, and Code-Mixed questions. However, only word embedding is not capable enough to offer multilingual and code-mixing capabilities. For an improved multilingual and code-mixing capability at a higher level, we introduce the shared encoding layers. In order to capture the notion of a phrase, first, the embedded input $\{q_t^e\}_{t=1}^T$ is passed to a CNN layer. Mathematically, we compute the inner product between the filter $F_l \in \mathbb{R}^{l \times d}$ and windows of $l$ word embedding. In order to maintain the length of the question after convolution, we perform appropriate zero-padding to the start and end of the embedded input $\{q_t^e\}_{t=1}^T$. The convoluted feature $q_t^{l,c}$ for $l$ length filter is computed as follows:

$$q_t^{l,c} = tanh(F_l q_{t:t+l-1}^e) \tag{5.2}$$

---
[4] It denotes the question in English, Hindi or Code-Mixed

A set of filters $L$ of different window sizes are applied on the embedded input. The final output $q_t^c$ at a time step $t$ is computed by the max-pooling operation over different window size filters. Mathematically, $q_t^c = max(q_t^{l_1,c}, q_t^{l_2,c}, \ldots, q_t^{l_L,c})$. The final representation computed by the CNN layer can be denoted as $\{q_t^c\}_{t=1}^T$. Inspired by the success in other NLP tasks [169, 336], we employ stacking of multiple Bi-LSTM layers to capture the semantic representation of an entire question. The input to the first layer of LSTM is the convoluted representation of the question $\{q_t^c\}_{t=1}^T$.

$$q_t^r = \text{Bi-LSTM}(q_{t-1}^r, q_t^c) \quad (5.3)$$

where, $q_t^r$ and $q_{t-1}^r$ are the hidden representations computed by the Bi-LSTM network at time $t$ and $t-1$, respectively. Specially, we compute the forward $\overrightarrow{q_t^r}$ and backward hidden representation $\overleftarrow{q_t^r}$ at each time step $t$ and concatenate them to obtain the final representation $q_t^r = \overrightarrow{q_t^r} \oplus \overleftarrow{q_t^r}$. The output from the previous layer of LSTM is passed as input to the next layer of LSTM.

**Layer Attention:** The encoding layers discussed in Section 5.3.1 are exploited by the questions from English, Hindi, and Code-Mixed languages. It might not be the case that the representation of a question (in a given language) obtained from a particular encoding layer would make a meaningful representation for the same question (in another language). We introduce a (similar to) attention-based mechanism over the encoding layer to learn the language-specific parameters in the network. Basically, our model learns an attention vector over each encoder layer for each language. Our model learns a language importance weight matrix $W \in \mathbb{R}^{m \times n}$, where $m$ and $n$ correspond to the number of encoding layers and the languages, respectively. The language importance weight matrix $W$ is applied to a given language's ($i$) question representation in the $j^{th}$ encoding layer. Let us assume that the $j^{th}$ multilingual encoding layer generates the question representation: $Q^{i,j} = \{q_1^{i,j}, q_2^{i,j}, \ldots, q_T^{i,j}\}$. The language attentive representation for a language $i$ and layer $j$ is computed as follows:

$$\begin{aligned}\overline{q_t^{i,j}} &= \overline{W_{i,j}} q_t^{i,j}, \quad t = \{1, 2, \ldots, T\} \\ \overline{W_{i,j}} &= \frac{e^{-W_{i,j}}}{\sum_{k=1}^n e^{-W_{k,j}}}\end{aligned} \quad (5.4)$$

The weighted question representation of $i^{th}$ language obtained from the $j^{th}$ layer can be denoted as $\overline{Q^{i,j}} = \{\overline{q_1^{i,j}}, \overline{q_2^{i,j}}, \ldots, \overline{q_T^{i,j}}\}$.

In our work, we use one layer of CNN and two layers of Bi-LSTM to encode multilingual questions. At each layer of encoding, we apply language-specific weight to obtain the language-specific encoding layer representation. We denote the question representation obtained from the final encoding layer after applying the language-specific attention as $h = \{h_t\}_{t=1}^{T}$.

**Self-Attention on Question:** Inspired by the success of self-attention on various NLP tasks [284, 137]; we adopt self-attention to our model for an improved representation of a word. As the model processes the current word from the question, self-attention allows it to look at the other words in the input question for clues that can help lead to a better encoding for the current word. The representations obtained from the multilingual encoding layer (c.f. Section 5.3.1) are passed to the self-attention layer. The multi-head self-attention mechanism [284] used in our model can be precisely described as follows:

$$Attention(Q, K, V) = \text{softmax}(\frac{QK^T}{\sqrt{d_h}})V \quad (5.5)$$

where, $Q$, $K$, $V$, and $d_h$ are the query, key, value matrices, and dimension of the hidden representation obtained from the multilingual encoding layer, respectively.

These matrices are obtained by multiplying different weight matrices to $h$. The value $d_h$ is the dimension of the hidden representation obtained from the multilingual encoding layer. Firstly, multi-head attention linearly projects the queries, keys, and values to the given head ($p$) using different linear projections. These projections then perform the scaled dot-product attention in parallel. Finally, these results of attention are concatenated and once again projected to obtain new representation. Formally, attention head ($z_p$) at given head $p$ can be expressed as follows:

$$\begin{aligned}z_p &= Attention(hW_p^Q, hW_p^K, hW_p^V) \\ &= softmax(\frac{(hW_p^Q)(hW_p^K)^T}{\sqrt{d_h}})(hW_p^V)\end{aligned} \quad (5.6)$$

where $W_p^Q$, $W_p^K$ and $W_p^V$ are the weight matrices.

We exploit multiple heads to obtain the attentive representation. Finally, we con-catenate all the attention heads to compute the final representation. The final question encoding obtained from the multilingual encoding layer can be represented by $U = \{q_1^h, q_2^h, \ldots, q_T^h\}$.

### 5.3.2 Image Features

Unlike the previous works [69, 334, 16] on VQA, in this work we extract two different levels of features, *viz.* image level and object level. We employ ResNet101 [102] model pre-trained on ImageNet [51] to obtain the image level features $V_i \in \mathbb{R}^{d_i \times n_i}$, where $n_i$ denotes the number of the spatial location of dimension $d_i$. We take the output of the pooling layer before the final softmax layer. To generate object-level features, we use the technique as discussed in [9] by using Faster R-CNN framework [222]. The resulting object-level features $V_o \in \mathbb{R}^{d_o \times n_o}$ can be interpreted as ResNet features focused on the top-$n_o$ objects in the image.

### 5.3.3 Multimodal Fusion

We fuse the multilingual question encoding (c.f. Section 5.3.1) and image features $V_i \in \mathbb{R}^{d_i \times M}$ by adopting the attention mechanism described in [130]. Let us denote the question encoding feature by $U \in \mathbb{R}^{n_1 \times T}$ and the image feature by $V \in \mathbb{R}^{n_2 \times R}$. The $k^{th}$ element representation using the bi-linear attention network can be computed as follows:

$$f_k = (U^T X)_k^T \mathcal{M}(V^T Y)_k \quad (5.7)$$

where $X \in \mathbb{R}^{n_1 \times K}$, $Y \in \mathbb{R}^{n_2 \times K}$, $(U^T X)_k \in \mathbb{R}^T$, $(V^T Y)_k \in \mathbb{R}^R$ are the weight matrices and $\mathcal{M} \in \mathbb{R}^{T \times R}$ is the bi-linear weight matrix. The E.q. 5.7 computes the 1-rank bi-linear representation of two feature vectors. We compute the $K$-rank bi-linear pooling for $f \in \mathbb{R}^K$. With $K$- rank bi-linear pooling, the bi-linear feature representation will be computed by multiplying a pooling vector $P \in \mathbb{R}^{K \times C}$ with $f$.

$$\overline{f} = P^T f \quad (5.8)$$

where $C$ is the dimension of the bi-linear feature vector. The $\overline{f}$ is a function of $U$, $V$ with the parameter (attention map) $\mathcal{M}$. Therefore, we can represent $\overline{f} = fun(U, V; \mathcal{M})$. Similar to [130], we compute multiple bi-linear attention maps (called visual heads) by introducing different pooling vectors. To integrate the representations learned from multiple bi-linear attention maps, we use the multi-modal residual network (MRN) [131]. Using MRN, we can compute the joint feature representation in a recursive manner:

$$\overline{f_{j+1}} = fun_j(\overline{f_j}, V; \mathcal{M}_j).\mathbf{1}^T + \overline{f_j} \tag{5.9}$$

The base case $\overline{f_0} = $ U and $\mathbf{1} \in \mathbb{R}^T$ is the vector of ones. We extract the joint feature representation for image level $\overline{f}_i$ as well as object-level feature $\overline{f}_o$.

### 5.3.4 Answer Prediction

Given the final joint representation of the question with image level and object level features (c.f. Section 5.3.3), we augment both of these features to the counter feature ($c_f$) proposed in [340]. The counter feature helps the model to count the objects. Finally, we employ a two-layer perceptron to predict the answer from a fixed set of candidate answers ( answers obtained from the training set). Towards this, the logits can be computed by the following equation:

$$A_{logits} = Relu\big(MLP(\overline{f}_i \oplus \overline{f}_o \oplus c_f)\big) \tag{5.10}$$

The $A_{logits}$ is passed to a *softmax* function to predict the answer.

## 5.4 Datasets and Experimental Details

### 5.4.1 Datasets and Network Training

In our experiments, we use the VQA v1.0 dataset for English questions. There isn't a single setup for a multilingual VQA system that can handle both multilingual and code-mixed questions at the same time. Therefore, our primary motivation has been to set up a basic VQA system using the VQA v1.0 dataset. For Hindi and Code-mixed questions, we use our own multilingual VQA dataset (c.f. Section 5.2). Both the datasets have 248, 349 and 121, 512 questions in their training and test set, respectively. Each question has 10 answers. The test dataset of English VQA does not have publicly available ground truth answers. In order to make a fair comparison of the results in all three setups, *viz.* English,

Hindi, and Code-mixed, we evaluate our proposed multilingual model on the validation set of English and test set of Hindi and Code-Mixed dataset (`MCVQA` dataset).

The training is performed jointly with English, Hindi, and Code-Mixed QA pairs by interleaving batches. We update the gradient after computing the loss of each mini-batch from a given language of the sample (question, image, answer). The other baselines are trained and evaluated for each language separately. For evaluation, we adopt the accuracy metric as defined in [12].

### 5.4.2 Hyperparameters

For English, we use the *fastText* [22] word embedding of dimension 300. We use Hindi sentences from [23], and then train the word embedding of dimension 300 using the word embedding algorithm [22]. In order to obtain the embedding of Roman script, we transliterate[5] the Hindi sentence into the Roman script. These sentences are used to train the code-mixed embedding using the same embedding algorithm [22], and we generate the embedding of dimension 300. These three word embeddings have the same dimensions but they are different in vector spaces. Finally, we align monolingual vectors of Hindi and Roman words into the vector space of English word embedding using the approach as discussed in [36]. While training, the model loss is computed using the categorical cross-entropy function.

Optimal hyper-parameters are set to: maximum no. of words in a question=15, CNN filter size=$\{2, 3\}$, # of shared CNN layers=1, # of shared Bi-LSTM layers=2, hidden dimension =1000, # of attention heads=4, image level, and object level feature dimension =2048, # of spatial location in image level feature =100, # of objects in object level feature=36, # of rank in bi-linear pooling=3, # of bilinear attention maps=8, # of epochs=100, initial learning rate=0.002. Optimal values of the hyperparameters are chosen based on the model performance on the development set of the VQA v1.0 dataset. Adamax optimizer [135] is used to optimize the weights during training.

### 5.4.3 Results

In order to compare the performance of our proposed model, we define the following baselines: MFB [334], MFH [335], Bottom-up-Attention [9] and Bi-linear Attention Network

---
[5]<https://github.com/libindic/indic-trans>

| Dataset | Models | Overall | Other | Number | Yes/No |
|---|---|---|---|---|---|
| English | MFB | 58.69 | 47.89 | 34.80 | 81.13 |
| English | MFH | 59.07 | 48.04 | 35.42 | 81.73 |
| English | BUTD | 63.50 | 54.66 | 38.81 | 83.60 |
| English | Bi-linear Attention | 63.85 | 54.56 | 41.08 | 81.91 |
| English | Proposed Model | **65.37** | **56.41** | **43.84** | **84.67** |
| Hindi | MFB | 57.06 | 46.00 | 33.63 | 79.70 |
| Hindi | MFH | 57.47 | 46.45 | 34.27 | 79.97 |
| Hindi | BUTD | 60.15 | 50.90 | 37.44 | 80.13 |
| Hindi | Bi-linear Attention | 62.50 | 52.99 | 40.31 | 82.66 |
| Hindi | Proposed Model | **64.51** | **55.37** | **42.09** | **84.21** |
| Code-mixed | MFB | 57.06 | 46.00 | 33.63 | 79.70 |
| Code-mixed | MFH | 57.10 | 46.09 | 33.56 | 79.71 |
| Code-mixed | BUTD | 60.51 | 51.68 | 36.47 | 80.37 |
| Code-mixed | Bi-linear Attention | 61.53 | 52.00 | 39.86 | 81.53 |
| Code-mixed | Proposed Model | **64.69** | **55.58** | **42.57** | **84.28** |

Table 5.2: Performance comparison between the state-of-the-art baselines and our proposed model on the VQA datasets. All the accuracy figures are shown in %. The improvements over the baselines are statistically significant as $p < 0.05$ for the t-test. At the time of testing, only one language input is given to the model.

[130]. These are the state-of-the-art models for VQA. We report the performance in Table 5.2.

The trained multilingual model is evaluated on the English VQA and `MCVQA` datasets as discussed in Section 5.4.1. Results of these experiments are reported in Table 5.2. Our proposed model outperforms the state-of-the-art English (with 65.37% overall accuracy) and achieves an overall accuracy of 64.51% and 64.69% on Hindi and Code-mixed VQA, respectively. Due to the shared hierarchical question encoder, our proposed model learns complementary features across questions of different languages.

### 5.4.4 Comparison to the non-English VQA

[71] created a VQA dataset for Chinese question-answer pairs and translated them into English. Their model takes the Chinese equivalent English question as input and generates an answer. A direct comparison in terms of performance is not feasible as they treat the problem as seq2seq learning [273] and their model was also trained on a monolingual (English) setup. We use their question encoding and language feature interaction compo-

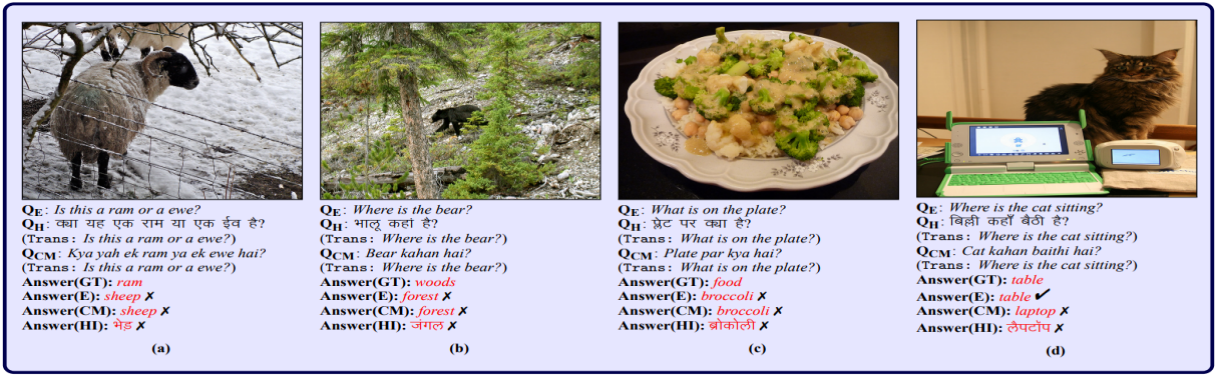

Figure 5.7: Some examples from `MCVQA` dataset where our model predicted incorrect answers. The notations are as follows:, **GT**: Ground Truth, **E**: English, **CM**: Code-mixed, **HI**: Hindi

| Dataset | Training | Overall | Other | Number | Yes/No |
|---|---|---|---|---|---|
| English | | 63.85 | 54.56 | 41.08 | 83.91 |
| Hindi | English | 24.13 | 4.41 | 0.34 | 58.46 |
| Code-mixed | | 28.04 | 11.95 | 0.49 | 58.79 |
| English | | 27.13 | 1.33 | 0.38 | 70.59 |
| Hindi | Hindi | 62.50 | 52.99 | 40.31 | 82.66 |
| Code-mixed | | 27.16 | 1.34 | 0.36 | 70.65 |
| English | | 30.92 | 9.45 | 0.39 | 69.85 |
| Hindi | Code-mixed | 21.76 | 2.02 | 0.35 | 36.22 |
| Code-mixed | | 61.53 | 52.00 | 39.86 | 81.53 |

Table 5.3: Results of cross-lingual experiments by training the [130] model on the training dataset of one language and evaluating on the rest.

nent to train a model with English questions and achieve overall accuracy of 57.89% on the English validation dataset (our model achieves 65.37%). Recently, [248] created a dataset for Japanese question-answer pairs and applied transfer learning to predict Japanese answers from the model trained on English questions. We adopt their approach, evaluate the model on English VQA and `MCVQA` dataset, and achieve 61.12%, 58.23%, 58.97% overall accuracy on English, Hindi, and Code-mixed, respectively. In comparison to these, we rather solve a more challenging problem that involves both multilingualism and code-mixing.

| Models<br>Component | English | Hindi | Code-mixed |
|---|---|---|---|
| Proposed | 65.37 | 64.51 | 64.69 |
| (-) CNN Layer | 64.92 | 64.19 | 64.38 |
| (-) Layer Attention | 64.52 | 63.72 | 63.89 |
| (-) Self Attention | 64.31 | 63.59 | 63.63 |
| (-) Image Level | 64.88 | 63.97 | 64.10 |
| (-) Object Level | 64.29 | 63.39 | 63.52 |
| (-) Counter | 64.67 | 64.03 | 63.92 |
| (-) Image (Language-only) | 40.89 | 45.70 | 45.23 |
| (-) Question (Vision-only) | 24.13 | 26.44 | 21.68 |

Table 5.4: Effect of various components of the model in terms of overall accuracy on English, Hindi and Code-mixed VQA datasets. **(-) X** shows the VQA model architecture after removal of component '**X**'

### 5.4.5 Analysis and Discussion

We perform an ablation study to analyze the contribution of various components of our proposed system. Table 5.4 shows the model performance by removing one component at a time. The self-attention on the question and object-level features seem to have the maximum effect on the model's performance. The object-level features contribute more as compared to the image-level features because the object level-features focus on encoding the objects of an image, which assists in answering the questions more accurately. Image grid-level features help the model to encode those parts of the image which could not be encoded by the object-level features.

The proposed VQA model is built on two channels: vision (image) and language (question). We perform a study (Table 5.4) to know the impact of both channels on the final prediction of the model. We turn off vision features and train the model with textual features to assess the impact of vision (image) features. Similarly, we also measure the performance of the system with image features (object and image level) only. Our study provides an answer to the following question: *"How much does a VQA model look at these channels to provide an answer?"*. The study reveals that the proposed VQA model is strongly coupled with both the vision and language channels. This confirms that the outperformance of the model is not because of the textual similarity between questions or pixel-wise similarity between the images.

We also perform experiments to evaluate the system in a cross-lingual setting. Toward this, we train the best baseline system [130] on the training dataset of one language and evaluate it on test datasets of the other two. The model performs pretty well when the languages for training and validation are the same. However, the performance of the model drops significantly when it is trained on one language and evaluated on a different language. We analyze the answers predicted by the model and make the following observations: **(1)** Our model learns the question representation from different surface forms (English, Hindi, and Hinglish) of the same word. It helps for a much better representation of multilingual questions by encoding their linguistic properties. This rich information also interacts with the image and extracts language-independent joint representation of the question and image. However, the state-of-the-art models are language-dependent. The question representation obtained from the state-of-the-art models could not learn language-independent features. Therefore, they perform poorly in cross-lingual and multilingual setups (results are reported in Table 5.3). **(2)** We observe that the model performance on the English VQA dataset is slightly better than Hindi and Code-mixed. One possible reason could be that the object-level features are extracted after training on the English Visual Genome dataset. Our VQA approach is language agnostic and can be extended to other languages as well.

**Error Analysis:** We perform a thorough analysis of the errors encountered by our proposed model on English VQA and `MCVQA` datasets. We categorize the following major sources of errors:

**(i) Semantic similarity**: This error occurs when an image can be interpreted in two ways based on its visual surroundings. In those scenarios, our model sometimes predicts the incorrect answer that is semantically closer to the ground truth answer. For example, in Figure 5.7(b), the question is *Where is the bear?*. Our model predicts the *forest* as the answer. However, the ground truth answer is *woods* which is semantically similar to *forest* and is a reasonable answer.

**(ii) Ambiguity in object recognition**: This error occurs when objects of an image have the similar object and image-level features. For example, in Figure 5.7(a) the question is *Is this a ram or a ewe?*. Our model predicts *sheep* as the answer in all three setups, but the ground truth answer is *ram*. As a *sheep*, a *ram*, and an *ewe* have similar object and image-level features and all of them resemble the same, our model could not predict the correct answer in such cases.

**(iii) Object detection at fine-grained level**: This type of error occurs when our model focuses on the fine-grained attributes of an image. In Figure 5.7(c), the question is *What is on the plate?*. The ground truth answer for this question is *food*. However, our model predicts *broccoli* as the answer. The food that is present on the plate is *broccoli*. This shows that our model is competent enough to capture the fine-grained characteristics of the image and thus predicts an incorrect answer.

**(iv) Cross-lingual training of object-level features**: Our proposed model has the capability to learn question features across multiple languages. However, the object-level features used in this work are trained on the English language dataset (Visual Genome dataset). We observe (c.f. Figure 5.7(d)) that the model sometimes fails when the question is in Hindi or Hinglish.

## 5.5 Conclusion

In this chapter, we have proposed a unified end-to-end framework for solving multilingual and code-mixed visual question answering. We also created a synthetic dataset for Hindi and code-mixed: MCVQA. Our created dataset MCVQA consists of $2,48,349$ training questions and $1,21,512$ test questions for real images in Hindi and Code-Mixed. We believe this dataset will enable the research in multilingual and code-mixed VQA. We also analyze the complexity of formulated code-mixed text in terms of CMI and CF scores. Besides, we also proposed a language-agnostic model for multilingual and code-mixed VQA.

Our unified end-to-end model is capable of predicting the answer for English, Hindi, and Code-Mixed questions. We have established our claim by performing a thorough analysis of the results obtained from the state-of-the-art baselines and proposed model. Experiments show that we achieve state-of-the-art performance on multilingual VQA tasks. We believe our work will pave the way toward the creation of multilingual and code-mixed AI assistants.

The next chapter is the beginning of part II of this dissertation. Part II focuses on advancing and improving the Question Answering system's performance by learning from the related sub-task of QA like semantic question matching, multi-hop question generation, and code-mixed text generation. Notably, in the next chapter, we will discuss the semantic question-matching framework. The model utilizes linguistic features obtained from taxonomy to improve the performance of the semantic question-matching architecture.

# Chapter 6

# Semantic Question Matching

## 6.1 Introduction

The previous chapters conclude Part I of this dissertation which focused on developing a robust MQA system to bridge the digital language gap by expanding the breadth of the current QA system in a multilingual and code-mixed setting.

In this chapter, we solve the problem of semantic question matching. Towards this, first, we created a semantic question-matching dataset. Next, we focused on building a hierarchical taxonomy for natural questions that can categorize the questions into their coarse and fine classes. Finally, we infuse the taxonomy knowledge into the proposed neural semantic question matching model to improve the performance of the system for the semantic question retrieval task.

Question answering is a well-investigated research area in NLP. Several existing QA systems answer factual questions with short answers [116, 21, 193]. However, systems that attempt to answer questions with long answers with several well-formed sentences are rare in practice. This is mainly due to some of the following challenges: **(i).** selecting appropriate text fragments from the document(s); **(ii).** generating answer texts with coherent and cohesive sentences; **(iii).** ensuring the syntactic as well as semantic well-formedness of the answer text. However, when we already have a set of answered questions, reconstructing the answers for semantically similar questions can be bypassed. For each unseen question, the most semantically similar question is identified by comparing the unseen question with the existing set of questions. The question, which is closest to the unseen question, can be retrieved as a possible semantically similar question. Thus, accurate semantic question matching can significantly improve a QA system.

In the proposed semantic question-matching framework, we use attention-based neural network models to generate question vectors. We create a hierarchical taxonomy by considering different types and sub-types so that questions with similar answers belong to the same taxonomy class. We propose and train a deep learning-based question classifier network to classify the taxonomy classes. The taxonomy information helps to decide on semantic similarity between classes. For example, the questions '*How do scientists work?*' and '*Where do scientists work?*', have very high lexical similarity, but they have different answer types. This can be easily identified using a question taxonomy. Taxonomy can provide useful information when we do not have enough data to generate useful deep learning-based representations, which is generally the case with restricted domains [184, 185]. In such scenarios, linguistic information obtained from prior knowledge helps significantly improve the system's performance.

We propose a neural network-based algorithm to classify the questions into the appropriate taxonomy class(es). Thus, the information obtained from the taxonomy is used along with the deep learning (DL) techniques to perform semantic question matching. Empirical evidence establishes that our taxonomy, when used in conjunction with DL representations, improves the system's performance on semantic question (SQ) matching task.

### 6.1.1 Problem Statement

Semantically similar questions are a set of questions that appear in different lexical forms, looking for almost the same answer. Given a natural language question $Q$ and a set of candidate questions $CQ$, the task is to rank each question $Q_{cq} \in CQ$ according to their semantic similarity to question $Q$. The examples of semantically similar questions are as follows:

**Question:** *"Which are the symptoms of the coronavirus disease?"*.
**Question:** *"How do we find out if anyone is suffering from the COVID-19?"*.
**Question:** *"What does having the coronavirus feel like?"*.

The above questions are semantically similar questions as the answer to all of them are nearly the same.

### 6.1.2 Motivation

- Identifying already answered semantically similar questions greatly increases the accuracy of a question-answering (QA) system.
- [249] discovered that about 15% of the questions on Yahoo! Answers do not receive any answer and leave an asker unsatisfied. This event is also termed as '*Question Starvation*' in the literature.
- Another motivation comes from question-answer website Quora[1] (Fig 6.1), where the question like "*What should I know before going to watch Interstellar?*" has many answers (84) but the related (semantically similar) questions may not have a sufficient answer. In this case, an efficient semantically similar question-matching system can help to redirect the questions to have the maximum answer.

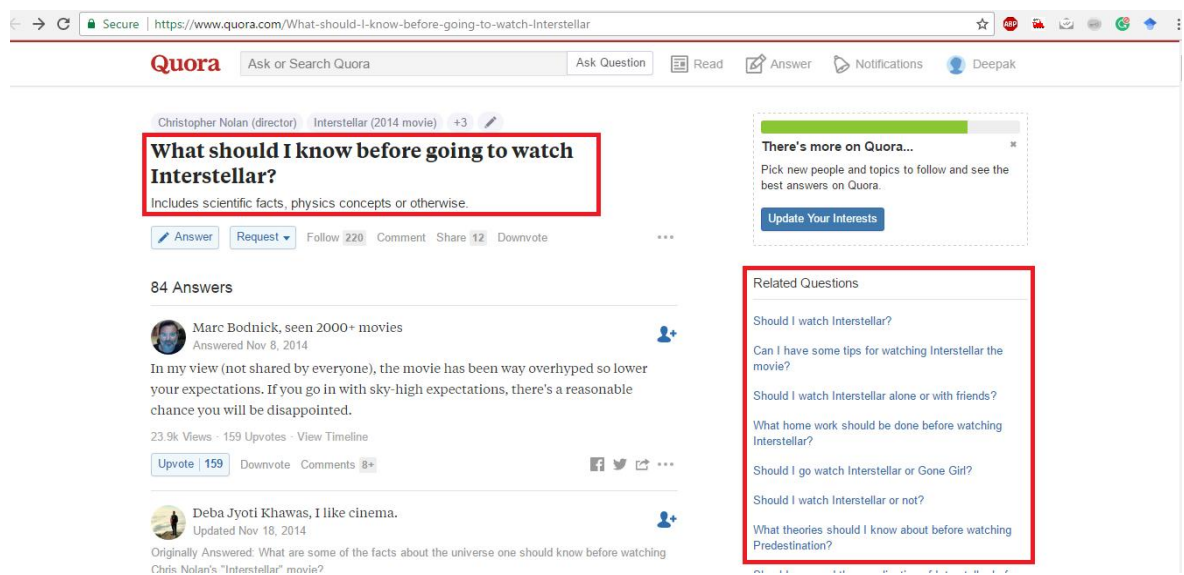

Figure 6.1: Example of question and related questions from Quora.

- Motivating examples from Quora dataset[2] are shown in Table 6.1, where the question pair share may lexical terms, however, the questions are not semantically similar. In this case, the taxonomy can help the model to detect semantic similarity because the questions differ at coarse and fine level information.

---

[1]<https://www.quora.com/>
[2]<https://data.quora.com/First-Quora-Dataset-Release-Question-Pairs>

| Question-1 | Question-2 | isSimilar |
|---|---|---|
| How do scientists work? | Where do scientists work? | No |
| How magnets are made? | What are magnets made of? | No |
| How do lymph nodes form? | What are lymph nodes? | No |

Table 6.1: Example of the question pair from Quora, where the question differ at a finer level.

### 6.1.3 Challenges

- **Closed-domain Dataset:** There is an issue of data scarcity in a closed domain, which is a significant cause to inhibit building a deep learning system for semantic question matching.
- **Taxonomy Coverage:** There are existing question-answering taxonomy [155], which has less coverage on the open-domain dataset. This challenge is manifold in the case of a close-domain dataset.
- **Taxonomy Class Identification:** The identification of the taxonomy class from a question is another challenge. The rule-based identification system lacks generalization, whereas a neural network-based system requires sufficient annotated data to train the network for correctly identifying taxonomy classes.
- **Question Focus:** Identifying the main focus [183, 319, 318, 326] of the question is key to encoding the semantics of the question. The focus of the question helps the model to identify their counterpart semantic similar questions. However, the identification of focus is not trivial and requires a sophisticated technique to detect the main focus of the question.
- **Domain Adaptability :** This is another challenge with a semantic question-matching algorithm. A system trained on a particular domain does not perform well when it is evaluated on slightly different domains. This is a well-known challenge to taxonomy creation as well.

## 6.2 Question Matching Framework

When framed as a computational problem, semantic question matching for QA becomes equivalent to ranking questions in the existing question base according to their semantic similarity to the given input question. Existing state-of-the-art systems use either deep learning models [149] or traditional text similarity methods [118, 289] to obtain the similarity scores. In contrast, our framework of SQ matching efficiently combines deep learning-based question encoding and a linguistically motivated taxonomy. Algorithm 3 describes the precise method we follow. *Similarity*(.) is the standard cosine similarity function. '*fsim*' is focus embedding similarity, which is described later in Section 6.3.3.

**Algorithm 3** Semantic Question Matching Framework
```
procedure SQ MATCHING(QSet)
    RESULTS ← {}
    for (p, q) in QSet do
        p⃗, q⃗ ← Question-Encoder(p, q)
        sim ← Similarity(p⃗, q⃗)
        T_p^c, T_q^c ← Taxonomy-Classes(p, q)
        F_p, F_q ← Focus(p, q)
        F⃗_p, F⃗_q ← Focus-Encoder(F_p, F_q)
        fsim ← Similarity(F⃗_p, F⃗_q)
        Feature-Vector=[sim, T_p^c, T_q^c, fsim]
        result ← Classifier(Feature-Vector)
        RESULTS.append(result)
    return RESULTS
```

### 6.2.1 Question Encoder Model

Our question encoder model is inspired by the state-of-the-art question encoder architecture proposed by [149]. We extend the question encoder model of [149] by introducing an attention mechanism similar to [13] and [41]. We propose the attention-based version of two question encoder models, namely Recurrent Convolutional Neural Network (RCNN) [149] and GRU.

A question encoder with attention does not need to capture the whole semantics of the question in its final representation. Instead, it is sufficient to capture a part of the hidden state vectors of another question it needs to attend to while generating the final representation. Let $\mathbf{H} \in \mathbb{R}^{d \times n}$ be a matrix consisting of hidden state vectors $[h_1, h_2 \ldots h_n]$ that the question encoder (RCNN, GRU) produced when reading the $n$ words of the question, where $d$ is a hyperparameter denoting the size of embeddings and hidden layers. The attention mechanism will produce an attention weight vector $\alpha_t \in \mathbb{R}^n$ and a weighted hidden representation $r_t \in \mathbb{R}^d$.

$$\begin{aligned} C_t &= tanh(W^H H + W^v(v_t \otimes I_n)) \\ \alpha_t &= softmax(w^T C_t) \\ r_t &= H\alpha^T \end{aligned} \quad (6.1)$$

where $W^H$, $W^v \in \mathbb{R}^{d \times d}$, are trained projection matrices. $w^T$ is the transpose of the trained vector $w \in \mathbb{R}^d$. $v_t \in \mathbb{R}^d$ shows the embedding of token $x_t$ and $I_n \in \mathbb{R}^n$ is the vector of 1. The product $W^v(v_t \otimes I_n)$ is repeating the linearly transformed $v_t$ as many times (n) as there are words in the candidate question. Similarly, we can obtain the attentive hidden state vectors $[r_1, r_2 \ldots r_n]$. We apply the averaging pooling strategy to determine the final representation of the question.

Annotated data, $\mathscr{D} = \{(q_i, p_i^+, p_i^-)\}$ is used to optimize $f(p, q, \phi)$, where $f(.)$ is a measure of similarity between the questions $p$ and $q$, and $\phi$ is a parameter to be optimized. Here $p_i^+$ and $p_i^-$ correspond to the similar and non-similar question sets, respectively for question $q_i$. The maximum margin approach is used to optimize the parameter $\phi$. For a particular training example, where $q_i$ is similar to $p_i^+$, we minimize the max-margin loss $\mathcal{L}(\phi)$ defined as:

$$\mathcal{L}(\phi) = \max_{p \in Q'(q_i)} \left\{ f(q_i, p; \phi) - f(q_i, p_i^+; \phi) + \lambda(p, p_i^+) \right\} \quad (6.2)$$

where $Q'(q_i) = p_i^+ \cup p_i^-$, $\lambda(p, p_i^+)$ is a positive constant set to 1 when $p \neq p_i^+$, 0 otherwise.

### 6.2.2 Question Taxonomy

Questions are ubiquitous in natural language. Questions essentially differ on two fronts: semantic and syntactic. Questions that differ syntactically might still be semantically equivalent. Let us consider the following two questions:

- *What is the number of new hires in 2018?*
- *How many employees were recruited in 2018?*

Although the above questions are not syntactically similar, both are semantically equivalent and have the same answer. A well-formed taxonomy and question classification scheme can provide this information which eventually helps determine the semantic similarity between the questions.

According to [80], ontologies are commonly defined as specifications of shared conceptualizations. Informally, conceptualization is the relevant informal knowledge one can extract from their experience, observation, or introspection. Specification corresponds to the encoding of this knowledge in representation language. In order to create a taxonomy for questions, we observe and analyze questions from SQuAD released by [216] and question classifier data from [111] and [155]. The SQuAD dataset consists of 100,000+ questions and their answers, along with the text extracts from which the questions were formed. The other question classifier dataset contains 5,500 questions. In the succeeding subsections, we describe the coarse classes, fine classes, and focus of a question in detail. We have included an additional hierarchical taxonomy table with one example question for each class in Table 6.2.

| Coarse Classes | Fine Classes | Example Question |
|---|---|---|
| Quantification | Temperature | What are the approximate temperatures that can be delivered by phase change materials? |
| | Time/Duration | How long did Baena worked for the Schwarzenegger/Shriver family? |
| | Mass | What is the weight in pounds of each of Schwarzenegger 's Hummers? |
| | Number | How many students are in New York City public schools? |
| | Distance | How many miles away from London is Plymouth? |
| | Money | What is the cost to build Cornell Tech? |
| | Speed | Give the average speed of 1987 solar powered car winner? |
| | Size | How large is Notre Dame in acres? |
| | Percent | What is the college graduation percentage among Manhattan residents? |
| | Age | How old was Schwarzenegger when he won Mr. Universe? |
| | Rank/Rating | What rank did iPod achieve among various computer products in 2006? |
| Entity | Person | Who served as Plymouth 's mayor in 1993? |
| | Location | In what city does Plymouth 's ferry to Spain terminate? |
| | Organization | Who did Apple partner with to monitor its labor policies? |
| | Animal | Which animal serves as a symbol throughout the book? |
| | Flora | What is one aquatic plant that remains submerged? |
| | Entertainment | What album caused a lawsuit to be filed in 2001? |
| | Food | What type of food is NYC 's leading food export? |
| | Abbreviation | What does AI stand for? |
| | Technique | What is an example of a passive solar technique? |
| | Language | What language is used in Macedonia? |
| | Monuments | Which art museum does Notre Dame administer? |
| | Activity/Process | What was the name of another activity like the Crusades occurring on the Iberian peninsula? |
| | Disease | What kind of pain did Phillips endure? |
| | Award/Title | Which prize did Frederick Buechner create? |
| | Date | When was the telephone invented? |
| | Event | What event in the novel was heavily criticized for being a plot device? |
| | Sport/Game | Twilight Princess uses the control setup first employed in which previous game? |
| | Policy | What movement in the '60s did the novel help spark? |
| | Publication | Which book was credited with sparking the US Civil War? |
| | Body | What was the Executive Council an alternate name for? |
| | Thing | What is the name of the aircraft circling the globe in 2015 via solar power? |
| | Feature/Attribute | What part of the iPod is needed to communicate with peripherals? |
| | Industry Sector | In which industry did the iPod have a major impact? |
| | Website | Which website criticized Apple's battery life claims? |
| | Other Tangible | In what body of water do the rivers Tamar and Plym converge? |
| | Other Intangible | The French words Notre Dame du Lac translate to what in English? |
| Definition | Person | Who was Abraham Lincoln? |
| | Entity | What is a solar cell? |
| Description | Reason | Why are salts good for thermal storage? |
| | Mechanism | How do the BBC's non-domestic channels generate revenue? |
| | Cause & Effect | What caused Notre Dame to become notable in the early 20th century? |
| | Compare & Contrast | What was not developing as fast as other Soviet Republics? |
| | Describe | What do greenhouses do with solar energy? |
| | Analysis | How did the critics view the movie , " The Fighting Temptations "? |
| List | Set of fine classes listed in the coarse classes *Quantification* and *Entity* | What are some examples of phase change materials? Which two national basketball teams play in NYC? |
| Selection | Alternative/Choice | Are the Ewell 's considered rich or poor? |
| | True/False | Is the Apple SDK available to third-party game publishers? |

Table 6.2: The exemplar description of proposed taxonomy classes

**Coarse Classes:** To choose the correct answer to a question, one needs to understand the question and categorize the answer into the appropriate category, which could vary from a basic implicit answer (the question itself contains the answer) to a more elaborate answer (description). The coarse class of questions provides a broader view of the expected answer type. We define the following six coarse class categories: *'Quantification'*, *'Entity'*, *'Definition'*, *'Description'*, *'List'* and *'Selection'*. *Quantification* class deals with the questions which look for a specific quantity as an answer. Similarly, *'Entity'*, *'Definition'*, *'Description'* classes give evidence that the answer type will be entity, definition, or a detailed description, respectively. *'Selection'* class defines the question that looks for an answer which needs to be selected from the given set of answers. A few examples of questions, along with their coarse class, are listed here:

- **Quantity:** *Give the average speed of 1987 solar-powered car winner?*
- **Entity:** *Which animal serves as a symbol throughout the book?*

**Fine Classes:** Coarse class defines the answer type at the broad level, such as entity, quantity, and description. However, extracting the actual answer to the question needs further classification into more specific answer types. Let us consider the following examples of two questions:

1. **Entity (Flora):** *What is one aquatic plant that remains submerged?*
2. **Entity (Animal):** *Which animal serves as a symbol throughout the book?*

Although both the questions belong to the same coarse class *'entity'*, they belong to the different fine classes, (*'Flora'* and *'Animal'*). The fine class of a question is based on the nature of the expected answer. It is useful in restricting potential candidate matches. Although questions belonging to the same fine class need not be semantically similar, questions belonging to different fine classes rarely match. We show the set of the proposed coarse class and their respective fine classes in Table 6.3.

**Focus of a Question:** According to [183], *focus* of a question is a word or a sequence of words, which defines the question and disambiguate it to find the correct answer the question is expecting to retrieve. In the following example, *"Describe the customer service model for Talent and HR BPO"*, the term 'model' serves as the *focus*. As per [27], *focus* of a question is contained within the noun phrases of a question. In the case of imperatives, the direct object (*dobj*) of the *question word* contains the *focus*. Similarly, in the case of interrogatives, there are certain dependencies that capture the relation between the question word and its focus. The *'dobj'* relation of the root verb or *'det'* relation of *question word* for interrogatives contains the *'focus'*. Question word *'how'* has *'advmod'* relations that contain *focus* of the question. Priority order of the relations used to extract *'focus'* is obtained by observation on the SQuAD data.

| Coarse Classes | Fine Classes |
|---|---|
| **Quantification** | Temprature, Time/Duration, Mass, Number, Age, Distance, Money, Speed, Size, Percent, Rank/Rating |
| **Entity** | Person, Location, Organization, Animal, Technique, Flora, Entertainment, Food, Abbreviation, Language, Disease, Award/Title, Event, Sport/Game, Policy, Date, Publication, Body, Thing, Feature/Attribute, Website, Industry Sector, Monuments, Activity/Process, Other Tangible, Other Intangible |
| **Definition** | Person, Entity |
| **Description** | Reason, Mechanism, Cause & Effect, Describe, Compare & Contrast, Analysis |
| **List** | Set of fine classes listed in the coarse classes *Quantification* and *Entity* |
| **Selection** | Alternative/Choice, True/False |

Table 6.3: Set of proposed coarse and respective fine classes

### 6.2.3 Question Classification

Question classification guides a QA system to extract the appropriate candidate answer from the document/corpus. For example, the question *"How much does international cricket player get paid?"* should be accurately classified as the coarse class *'quantification'* and fine class *'money'* to further extract the appropriate answer. In our problem, we attempt to exploit the taxonomy information to identify semantically similar questions. Therefore, the question classifier should be capable enough to accurately classify the coarse and fine classes of a reformulated question:

1. *What is the salary of an international level cricketer?*
2. *What is the estimated wage of an international cricketer?*

**Question Classification Network:** In order to identify the coarse and fine classes of a given question, we employ a deep learning-based question classifier. In our question classification network, CNN and bidirectional GRU are applied sequentially. The obtained question vector is passed through a feed-forward NN layer, and then through a softmax layer to obtain the final class of the question. We use two separate classifiers for coarse and fine class classification.

Firstly, an embedding layer maps a question $Q = [w_1, w_2 \ldots w_n]$, which is a sequence of words $w_i$, into a sequence of dense, real-valued vectors, $E = [v_1, v_2 \ldots v_n]$, $v_i \in \mathbb{R}^d$. Thereafter, a convolution operation is performed over the zero-padded sequence $E^p$. $F \in \mathbb{R}^{k \times m \times d}$, a set of $k$ filters is applied to the sequence. We obtain convoluted features $c_t$ at given time $t$ for $t = 1, 2, \ldots, n$.

$$c_t = tanh(F[v_{t-\frac{m-1}{2}} \ldots v_t \ldots v_{t+\frac{m-1}{2}}]) \tag{6.3}$$

Then, we generate the feature vectors $C' = [c'_1, c'_2 \ldots c'_n]$, by applying max pooling on $C$. This sequence of convolution feature vector $C'$ is passed through a bidirectional GRU network. We obtain the forward hidden states $\overrightarrow{h_t}$ and backward hidden states $\overleftarrow{h_t}$ at every step time $t$. The final output of recurrent layer $h$ is obtained as the concatenation of the last hidden states of forward and backward hidden states.

Finally, the fixed-dimension vector $h$ is fed into the softmax classification layer to compute the predictive probability $p(y = l|Q) = \frac{exp(w_l^T h + b_l)}{\sum_{i=1}^{L} exp(w_i^T h + b_i)}$ for all the question classes (coarse or fine). We assume there are $L$ classes where $w_x$ and $b_x$ denote the weight and bias vectors, respectively, and $x \in \{l, i\}$.

### 6.2.4 Comparison with Existing Taxonomy

[155] proposed a taxonomy to represent a natural semantic classification for a specific set of answers. This was built by analyzing the TREC questions. In contrast to [155], along with TREC questions, we also make a thorough analysis of the most recent question-answering dataset (SQuAD), which has a collection of more diversified questions. Unlike [155], we introduce the *'list'* and *'selection'* type question classes in our taxonomy. Each of these question types has its own strategy to retrieve an answer, and therefore, we put these separately in our proposed taxonomy. The usefulness of list as a different coarse class in semantic question matching can be understood considering the following questions:

1. *What are some techniques used to improve crop production?*
2. *What is the best technique used to improve crop production ?*

These two questions are not semantically similar as **(1)** and **(2)** belong to *'list'* and *'entity'* coarse classes, respectively. Moreover, [155]'s taxonomy has overlapping classes (*'Entity'*,

'Human' and 'Location'). In our taxonomy, we put all these classes in a single coarse class named *'Entity'*, which helps in identifying semantically similar questions better. We propose a set of coarse and respective fine classes with more coverage compared to [155]. [155] taxonomy does not cover many important fine classes such as *'entertainment'*, *'award/title'*, *'activity'*, *'body'* etc., under *'entity'* coarse class. We include these fine classes in our proposed taxonomy. We further redefine *'description'* type questions by introducing *"cause & effect"*, *"compare and contrast"* and *'analysis'* fine classes in addition to *'reason'*, *'mechanism'* and *'description'* classes. This finer categorization helps in choosing a more appropriate answer strategy for descriptive questions.

## 6.3 Datasets and Experimental Details

### 6.3.1 Datasets

We perform experiments on three benchmark datasets, namely Partial Ordered Question Ranking (POQR)-Simple, POQR-Complex [26], and Quora datasets. In addition to this, we also perform experiments on a new semantic question-matching dataset (Semantic SQuAD[3]) created by us. In order to evaluate the system performance, we perform experiments in two different settings. The first setting deals with semantic question ranking (SQR), and the other deals with semantic question classification (SQC) with two classes (match and no-match). We perform SQR experiments on Semantic SQuAD and POQR datasets. For SQC experiments, we use Semantic SQuAD and Quora datasets.

1. **Semantic SQuAD:** We built a semantically similar question-pair dataset based on a portion of SQuAD data. SQuAD, a crowd-sourced dataset, consists of 100,000+ answered questions along with the text from which those question-answer pairs were constructed. We randomly selected $6,000$ in question-answer pairs from the SQuAD dataset. For a given question, we asked 12 annotators[4] to formulate semantically similar questions referring to the same answers. Each annotator was asked to formulate 500 questions. We divided this dataset into training, validation, and test sets of $2,000$ pairs each. We further constructed $4,000$ *semantically dissimilar* questions automatically. We use these $8,000$ question pairs ($4,000$ semantic similar questions pair from test and validation + 4,000 semantically different pairs) to train the semantic question classifier for the SQC setting of the experiments. *Semantically dissimilar* questions are created by maintaining the constraint that questions should be from the different taxonomy classes. We perform 3-fold cross-validation on these 8,000 question pairs.

---

[3]All the datasets used in this work are publicly available at https://figshare.com/articles/Semantic_Question_Classification_Datasets/6470726

[4]The annotators are the post-graduate students having proficiency in the English language.

| Datasets | Simple | | Complex | |
|---|---|---|---|---|
| | Simple-1 | Simple-2 | Complex-1 | Complex-2 |
| $\mathcal{P}$ | 164 | 134 | 103 | 89 |
| $\mathcal{U}$ | 775 | 778 | 766 | 730 |
| $\mathcal{N}$ | 594 | 621 | 664 | 714 |
| Pairs | 11015 | 10436 | 10654 | 9979 |

Table 6.4: Brief statistics of POQR datasets

2. **POQR Dataset:** POQR dataset consists of 60 groups of questions, each having a reference question that is associated with a partially ordered set of questions. Each group has three different sets of questions named as *paraphrase* ($\mathcal{P}$), *useful* ($\mathcal{U}$) and *neutral* ($\mathcal{N}$). For each given reference question $q_r$ we have $q_p \in \mathcal{P}$, $q_u \in \mathcal{U}$, and $q_n \in \mathcal{N}$. As per [26] the following two relations hold:

   (a) $(q_p \succ q_u | q_r)$: A *paraphrase* question is *'more useful than'* useful question.
   (b) $(q_u \succ q_n | q_r)$: A *useful* question is *'more useful than'* neutral question.

   By transitivity, it was assumed by [26] that the following ternary relation holds $(q_p \succ q_n | q_r)$: "A *paraphrase* question is *'more useful than'* a neutral question". We show the statistics of these datasets for *Simple* and *Complex* question types for two annotators (1, 2) in Table 6.4.

3. **Quora Dataset:** We perform experiments on the semantic question matching dataset consisting of 404,290 pairs released by Quora[5]. The dataset consists of 149,263 matching pairs and 255,027 non-matching pairs.

### 6.3.2 Evaluation Schemes

We employ different evaluation schemes for our SQR and SQC evaluation settings. For the **Semantic SQuAD** dataset, we use the following metrics for ranking the evaluation: Recall@$k$ results for $k = 1, 3$ and $5$, MRR, and MAP. The set of all candidate questions in 2,000 pairs of the test set is ranked against each input question. As we have only a 1 correct match out of 2,000 questions for each question in the test set, recall@1 is equivalent to precision@1. Given that we only have one relevant result for each input question, MAP is equivalent to MRR. We evaluate the semantic question classification performance in terms of accuracy. To ensure fair evaluation, we keep the ratio of semantically similar and dissimilar questions to 1:1. In order to compare the performance on **POQR dataset** with the state-of-the-art results, we followed the same evaluation scheme as described in [26]. It is measured in terms of 10-fold cross-validation accuracy on the set of ordered pairs, and the performance is averaged between the two annotators (1,2) for the Simple and Complex datasets. For **Quora dataset**, we perform 3-fold cross-validation on the entire dataset evaluating based on the classification accuracy only. We did not perform the semantic question ranking (SQR) experiment on the Quora dataset as 149,263 × 149,263 ranking experiment for matching pairs takes a very long time.

---
[5]https://data.quora.com/First-Quora-Dataset-Release-Question-Pairs

### 6.3.3 Experimental Setup

**Question Encoder:** We train two different question encoders (hidden size=300) on *Semantic SQuAD* and *Quora* datasets. For the Semantic SQuAD dataset, we used 2,000 training pairs to train the question encoder, as mentioned in Section 1. For the Quora dataset, we randomly select 74,232 semantically similar question pairs to train the encoder and 10,000 question pairs for validation. The best hyper-parameters for the deep learning-based attention encoder are identified on validation data. Adam [136] is used as the optimization method. Other hyper-parameters used are learning rate (0.01), dropout probability [109]: (0.5), CNN feature width (2), batch size (50), epochs (30) and size of the hidden state vectors (300). These optimal hyperparameter values are the same for the attention-based RCNN and GRU encoder. We train two different question encoders trained on *Semantic SQuAD* and *Quora* datasets. We could not train the question encoder on the **POQR dataset** because of the unavailability of a sufficient number of similar question pairs in this dataset. Instead, we use the question encoder trained on the Quora dataset to encode the questions from the POQR dataset.

**Question Classification Network:** To train the model, we manually label (using 3 English proficient annotators with an inter-annotator agreement of 87.98%) a total of 5,162 questions[6] with their coarse and fine classes, as proposed in Section 6.2.2. We release this question classification dataset to the research community. We evaluate the performance of question classification for 5-fold cross-validation in terms of F-Score. Our evaluation shows that we achieve 94.72% and 86.19% F-Score on coarse class (6-labels) and fine class (72-labels), respectively. We use this trained model to obtain the coarse and fine classes of questions in all datasets.

---

[6]4,000 questions are a part of the training set of Semantic SQuAD. The remaining 1,162 questions

We perform the SQC experiments with the SVM classifier. We use *libsvm* implementation [32] with linear kernel and polynomial kernel of degree $\in \{2, 3, 4\}$. The best performance was obtained using a linear kernel. Due to the nature of the POQR dataset, we employ SVM$^{light}$[7] implementation of ranking SVMs, with a linear kernel keeping standard parameters intact. In our experiments, we use the pre-trained Google embeddings provided by [179]. The focus embedding is obtained through word vector composition (averaging).

## 6.4 Results and Analysis

### 6.4.1 Baselines

We compare our proposed approach to the following IR-based baselines:

**1) TF-IDF:** The candidate questions are ranked using cosine similarity value obtained from the TF-IDF based vector representation.

**2) Jaccard Similarity:** The questions are ranked using Jaccard similarity calculated for each candidate question with the input question.

**3) BM-25:** The candidate questions are ranked using BM-25 score, provided by Apache Lucene[8].

### 6.4.2 Results

We present the detailed results of the semantic question ranking experiment on the Semantic SQuAD dataset in Table 6.5. In Table 6.6 and 6.7 we report the performance results on the respective dataset using the models **GRU**, **RCNN, GRU-Attention and RCNN-Attention** (c.f. Section 6.2.1). For all these models, the results reported in the tables are based on the cosine similarity of the respective question encoder. The attention-based

---

are from the dataset used in [155]

[7] http://svmlight.joachims.org/
[8] https://lucene.apache.org/core/

model obtains the maximum gains of 2.40% and 2.60% in terms of recall and MRR for the *GRU* model. The taxonomy augmented model outperforms the respective baselines and state-of-the-art deep learning-based question encoder models. We obtain the best improvements for the *Tax+RCNN-Attention* model, 3.75%, and 4.15% in terms of Recall and MRR, respectively. Experiments show that taxonomy features assist in consistently improving the R@k and MRR/MAP across all the models.

| Models | R@1 | R@3 | R@5 | MRR/MAP |
|---|---|---|---|---|
| **IR based Baselines** | | | | |
| TF-IDF | 54.75 | 66.15 | 70.25 | 61.28 |
| Jaccard Similarity | 48.95 | 62.80 | 67.40 | 57.26 |
| BM-25 | 56.40 | 69.35 | 71.45 | 61.93 |
| **Deep Neural Network (DNN) based Techniques** | | | | |
| GRU [149] | 73.25 | 84.12 | 86.39 | 76.77 |
| RCNN [149] | 75.10 | 86.35 | 89.01 | 78.24 |
| GRU-Attention | 74.89 | 86.02 | 88.47 | 78.30 |
| RCNN-Attention | 76.41 | 88.41 | 91.78 | 80.28 |
| **DNN + Taxonomy based Features** | | | | |
| Tax + GRU | 76.19 | 87.02 | 88.47 | 78.98 |
| Tax + RCNN | 78.32 | 88.91 | 92.35 | 81.49 |
| Tax + GRU-Attention | 77.35 | 89.22 | 91.28 | 80.95 |
| Tax + RCNN-Attention | 78.88 | 90.20 | 93.25 | 83.12 |

Table 6.5: **Semantic Question Ranking (SQR)** performance of various models on **Semantic SQuAD** dataset, R@k and Tax denote the recall@k & augmentation of taxonomy features respectively.

Performance of the proposed model on the POQR dataset is shown in Table 6.6. The '*overall*' column in Table 6.6 shows the performance average on simple-1,2 and complex-1,2 datasets. We obtain improvements (maximum of 1.55% with *GRU-Attention* model on Complex-1 dataset) in each model by introducing attention mechanism on both simple and complex datasets. The augmentation of taxonomy features helps in improving the performance further (8.75% with *Tax+RCNN-Attention* model on Simple dataset).

The system performance on semantic question classification (SQC) experiment with Semantic SQuAD and Quora datasets are shown in Table 6.7. Like ranking results, we obtain significant improvement by introducing the attention mechanism and augmenting the taxonomy features on both datasets.

| Models | Simple | | | Complex | | |
|---|---|---|---|---|---|---|
| | Simple-1 | Simple-2 | Overall | Complex-1 | Complex-2 | Overall |
| GRU [149] | 74.20 | 73.68 | 73.94 | 74.67 | 75.22 | 74.94 |
| RCNN [149] | 76.19 | 75.81 | 76.00 | 75.33 | 76.44 | 75.88 |
| GRU-Attention | 75.39 | 74.83 | 75.11 | 76.22 | 76.18 | 76.20 |
| RCNN-Attention | 77.28 | 77.01 | 77.14 | 76.63 | 77.31 | 76.97 |
| DNN + Taxonomy based Features | | | | | | |
| Tax+GRU | 78.29 | 79.01 | 78.65 | 77.63 | 78.97 | 78.30 |
| Tax+RCNN | 80.92 | 81.55 | 81.23 | 80.15 | 80.83 | 80.49 |
| Tax+GRU-Attention | 81.69 | 81.03 | 81.36 | 81.22 | 81.56 | 81.39 |
| Tax+RCNN-Attention | 83.67 | 83.98 | 83.82 | 83.32 | 84.10 | 83.71 |
| State-of-the art techniques | | | | | | |
| Unsupervised *Cos* [26] | - | - | 73.70 | - | - | 72.60 |
| Supervised *SVM* [26] | - | - | 82.10 | - | - | 82.50 |

Table 6.6: Semantic question ranking performance of various models on **POQR datasets**. All the numbers reported are in terms of accuracy.

**K-means Clustering Results:** The k-means clustering is performed on the question representation obtained from the best question (RCNN-Attention) encoder of $2,000$ semantic question pairs. The clustering experiment is evaluated on the test set of the Semantic SQuAD dataset (4000 questions). The performance is evaluated using the following metric:

$$\texttt{Recall} = \frac{100 \times no.\ of\ SQ\ pairs\ in\ same\ cluster}{total\ no.\ of\ SQ\ pairs} \quad (6.4)$$

K-means Clustering results are as follows: R@1:50.12, R@3:62.44, and R@5:66.58. As the number of clusters decreases, `Recall` is expected to increase as there is a higher likelihood of matching questions falling in the same cluster. `Recall` with $2,000$ clusters for $2,000$ SQ pairs i.e. $4,000$ questions is comparable to Recall@1 as we have 2 questions per cluster on average, `Recall` with $1,000$ clusters is a proxy for Recall@3 and `Recall` with 667 clusters is comparable to Recall@5.

### 6.4.3 Qualitative Analysis

We analyze the results that we obtain by studying the following effects:

1. **Effect of Attention Mechanism:** We analyze the hidden state representation the model is attending to when deciding the semantic similarity. We depict the visualization of attention weight between two semantically similar questions from the Semantic SQuAD dataset. We observe that the improvement due to the attention

| Models | Semantic SQuAD Dataset | Quora Dataset |
|---|---|---|
| **IR based Baselines** | | |
| TF-IDF | 59.28 | 70.19 |
| Jaccard Similarity | 55.76 | 67.11 |
| BM-25 | 63.78 | 73.27 |
| **Deep Neural network (DNN) based Techniques** | | |
| GRU [149] | 74.05 | 77.53 |
| RCNN [149] | 77.54 | 79.32 |
| GRU-Attention | 75.18 | 79.22 |
| RCNN-Attention | 79.94 | 80.79 |
| **DNN + Taxonomy based Features** | | |
| Tax + GRU | 77.32 | 79.21 |
| Tax + RCNN | 79.89 | 81.15 |
| Tax + GRU-Attention | 78.11 | 80.91 |
| Tax + RCNN-Attention | 82.25 | 83.17 |

Table 6.7: **Performance of Semantic Question Classification (SQC)** for various models on **Semantic SQuAD** and **Quora** datasets.

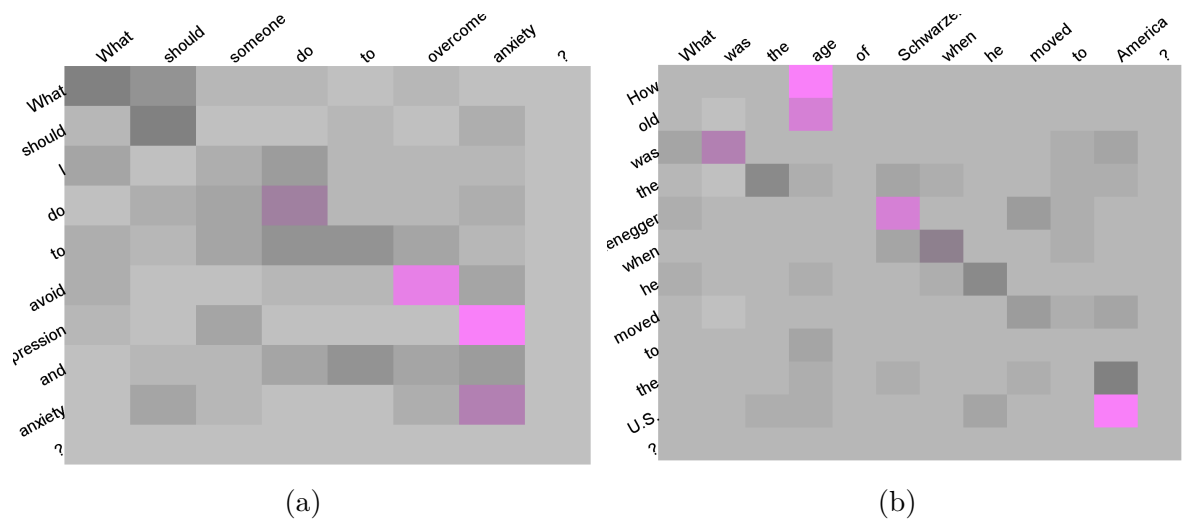

Figure 6.2: In **(a)** Attention mechanism detects semantically similar words (*avoid, overcome*). The attention mechanism is also able to align the multi-word expression '*how old*' to '*age*' as shown in **(b)**

mechanism in the Quora dataset is comparatively less than in the Semantic SQuAD dataset. The question pairs from the Quora dataset have matching words, and the problem is more focused on differences rather than similar or related words. For example, for the questions "*How magnets are made?*" and "*What are magnets made of?*", the key difference is question words '*how*' versus '*what*', while the remaining words are similar. The attention visualization is shown in Fig 6.2.

2. **Effect of Taxonomy Features:** We perform a feature ablation study on all the datasets to analyze the impact of each taxonomy feature. Table 6.8 shows the results[9] with the full features and after removing coarse class (-CC), fine class (-FC) and the focus features one by one. We observe from the Quora dataset that the starting word of the questions *(what, why, how)* is a deciding factor for semantic similarity. As the taxonomy features categorize these questions into different coarse and fine classes; therefore, it helps the system distinguish between semantically similar and dissimilar questions. It can be observed from the results that the augmentation of **CC** and **FC** features significantly improves the performance, especially on the Quora dataset. Similar trends were also observed on the other datasets.

| Sr. No. | Datasets | All | -CC | -FC | -Focus Word |
|---|---|---|---|---|---|
| 1 | Semantic SQuAD (SQR) | 83.12 | 81.66 | 81.84 | 82.20 |
| 2 | Semantic SQuAD (SQC) | 82.25 | 80.85 | 81.19 | 81.13 |
| 3 | POQR-Simple | 83.82 | 80.85 | 81.44 | 82.57 |
| 4 | POQR-Complex | 83.71 | 81.04 | 81.97 | 82.19 |
| 5 | Quora | 83.17 | 80.93 | 81.75 | 82.24 |

Table 6.8: Results of feature ablation on all datasets. **SQR** results are in **MAP**. The other results are shown in terms of **Accuracy**.

### 6.4.4 Comparison to State-of-the-Art

We compare the system performance on the POQR dataset with the state-of-the-art technique of [26]. [26] used several cosine similarities as features obtained using bag-of-words, dependency tree, focus, main verb, etc. Compared to [26], our model achieves better performance with an improvement of 2.1% and 1.46% on a simple and complex dataset, respectively. A direct comparison to SemEval-2017 Task-3[10] CQA or AskUbuntu [149] datasets could not be made due to the difference in the nature of questions. The proposed classification method is designed for well-formed English questions and could not be applied to multi-sentence / ill-formed questions. We evaluate [149]'s model (RCNN) on our datasets and report the results in Section 6.4.2. Quora has not yet released any official test set. Hence, we report the performance of 3-fold cross-validation on the entire dataset to minimize the variance. We can not directly make any comparisons with others due to the non-availability of an official gold standard test set.

---

[9]The results are statistically significant as $p < 0.002$.
[10]http://alt.qcri.org/semeval2017/task3/

### 6.4.5 Error Analysis

We observe the following as significant sources of errors in the proposed system: **(1)** Misclassification at the fine class level is often propagated to the semantic question classifier when some of the questions contain more than one sentence. E.g., "*What's the history behind human names? Do non-human species use names?*". **(2)** Semantically dissimilar questions having the same function words but different coarse and fine classes were incorrectly predicted as similar questions. It is because of the high similarity in the question vector and focus, which forces the classifier to commit mistakes. **(3)** In the semantic question ranking (SQR) task, some of the questions with higher lexical similarity to the reference question are selected prior to the actual similar question.

## 6.5 Conclusion

In this chapter, we have proposed an efficient model for semantic question matching where deep learning models are combined with pivotal features obtained from a taxonomy. We have created a two-layered taxonomy (coarse and fine) to organize the questions in interest and proposed a deep learning-based question classifier to classify the questions. We have established the usefulness of our taxonomy on two different tasks (SQR and SQC) on four different datasets. We empirically established that effective usage of semantic classification and focus of questions helps improve the performance of various benchmark datasets for semantic question matching.

In the next chapter, we will take the question generation problem and discuss the shortcoming of existing work along with the new direction to solve the problem of question generation.

# Chapter 7

# Multi-hop Question Generation

## 7.1 Introduction

In the previous chapter, we attempt to solve the problem of semantic question matching towards the improvement in the question answering system. In that direction, this chapter explores the task of multi-hop question generation. In NLP, question generation is an essential yet challenging problem. An effective question-generation system can assist in several real-world applications such as automatic tutoring systems to improve the performance of question-answering models and enable chatbots to lead a conversation. The goal of the question generation system is to generate a fluent question for a given passage and answer.

In the past, question generation has been tackled using rule-based approaches such as question templates [162] or utilizing named entity information and predictive argument structures of sentences [30]. Recently, neural-based approaches have accomplished impressive results [59, 272, 133] for the task of question generation. The availability of large-scale machine reading comprehension datasets such as SQuAD [216], NewsQA [281], MSMARCO [194] etc. has facilitated research in QA task. The SQuAD dataset has been the de-facto choice for most of the previous works in question generation. However, 90% of the questions in SQuAD can be answered from a single sentence [181]; hence former QG systems trained on SQuAD are not capable of distilling and utilizing information from multiple sentences. Recently released multi-hop datasets such as `QAngaroo` [299], `ComplexWebQuestions` [275] and `HotPotQA` [330] are more suitable for building QG systems that required to gather and utilize information across multiple documents as opposed to a single paragraph or sentence.

In multi-hop question answering, one has to reason over multiple relevant sentences from different paragraphs to answer a given question. We refer to these relevant sentences as supporting facts in the context. Hence, we frame *Multi-hop question generation* (MHQG) as the task of generating the question conditioned on the information gathered from reasoning over all the supporting facts across multiple paragraphs/documents. Since this task requires assembling and summarizing information from multiple relevant documents in contrast to a single sentence/paragraph, it is more challenging than the existing single-hop QG task. Further, the presence of irrelevant information makes it challenging to capture the supporting facts required for question generation. The explicit information about the supporting facts in the document is not often readily available, which makes the task more complicated. In this work, we provide an alternative to get the supporting facts information from the document with the help of multi-task learning.

In this chapter, we aim to tackle the Multi-hop QG problem in two stages. In the first stage, we learn supporting facts aware encoder representation to predict the supporting facts from the documents by jointly training with question generation. In the second stage, we aim to generate the question by enforcing the utilization of these supporting facts. The former is achieved by sharing the encoder weights with an *answer-aware* supporting facts prediction network, trained jointly in a multi-task learning framework. The latter objective is formulated as a *question-aware* supporting facts prediction reward, which is optimized alongside supervised sequence loss. Additionally, we observe that the multi-task framework offers substantial improvements in the performance of question generation and also avoid the inclusion of noisy sentence information in generated question, and reinforcement learning brings the complete and complex question to an otherwise maximum likelihood estimation optimized QG model.

### 7.1.1 Problem Statement

In multi-hop question generation, we consider a document list $L$ with $n_L$ documents, and an $m$-word answer $A$. Let the total number of words in all the documents $D_i \in L$ combined be $N$. Let a document list $L$ contains a total of $K$ candidate sentences $CS = \{S_1, S_2, \ldots, S_K\}$ and a set of supporting facts[1] $SF$ such that $SF \in CS$. The answer $A = \{w_{D_k^{a_1}}, w_{D_k^{a_2}}, \ldots, w_{D_k^{a_m}}\}$ is an $m$-length text span in one of the documents $D_k \in L$. Our task is to generate an $n_Q$-word question sequence $\hat{Q} = \{y_1, y_2, \ldots, y_{n_Q}\}$ whose answer is based on the supporting facts $SF$ in document list $L$. Mathematically,

$$\begin{aligned}
\hat{Q} &= \arg\max_Q Prob(Q|L, A; \phi) \\
&= \arg\max_Q Prob(Q|S_1, S_2, \ldots S_K, A; \phi)
\end{aligned} \quad (7.1)$$

---

[1] It is only used to train the network. At the time of testing, the network predicts the supporting facts to be used for question generation.

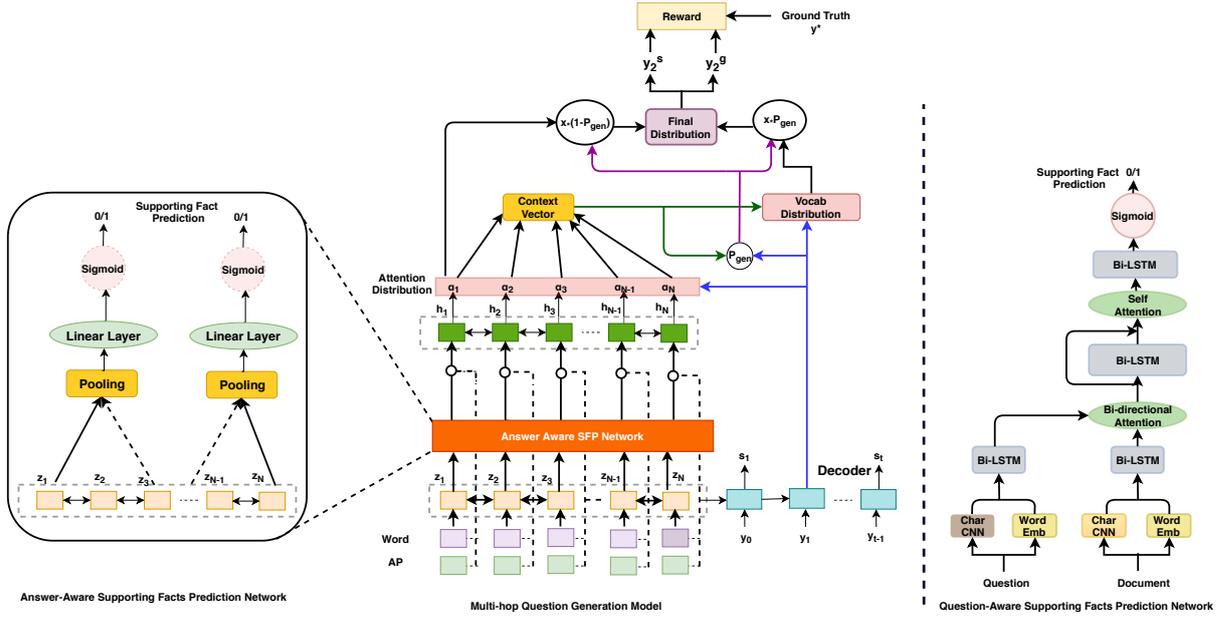

Figure 7.1: The architecture of our proposed Multi-hop QG network. The inputs to the model are the document word embeddings and answer position (**AP**) features. Question generation and answer-aware supporting facts prediction model (**left**) jointly train the shared document encoder (Bi-LSTM) layer. The image on the **right** depicts our question-aware supporting facts prediction network, which is our MultiHop-Enhanced Reward function. The inputs to this model are the generated question (output of multi-hop QG network) and a list of documents.

The example of single-hop question generation is shown in Table 7.1. Table 7.2 gives sample examples from SQuAD [216] and HotPotQA [330] dataset. It is clear from the example that the single-hop question is formed by focusing on a single sentence/document and answer, while in multi-hop questions, many supporting facts from different documents and answers are accumulated to form the question.

| Passage |
|---|
| Regional social norms are generally antagonistic to hunting, while a few sects, such as the **Bishnoi**[(1)], lay special emphasis on the conservation of particular species, such as the antelope. India's **wildlife protection act of 1972**[(2),(3)] bans the killing of all wild animals. |
| **Generated Question** |
| **Question:** Who lay special emphasis on the conservation of particular species? (1) <br> **Question:** What bans the killing of all wild animals in India? (2) <br> **Question:** What year was this protection act put into place? (3) |

Table 7.1: Example of the generated question given the answer and passage.

### 7.1.2 Motivation

- Multi-hop QG has real-world applications in several domains, such as education, chatbots, etc. The questions generated from the multi-hop approach will inspire critical thinking in students by encouraging them to reason over the relationship between multiple sentences to answer correctly. Specifically, solving these questions requires higher-order cognitive skills (e.g., applying, and analyzing). Therefore, forming challenging questions is crucial for evaluating a student's knowledge and stimulating self-learning.

- Similarly, in goal-oriented chatbots, multi-hop QG is an important skill for chatbots, e.g., in initiating conversations, and asking and providing detailed information to the user by considering multiple sources of information. In contrast to a single-hop QG where only a single source of information is considered to generate the question.

- The generated question using a multi-hop question generation technique can be used to improve the performance of the multi-hop question answering system.

- The existing research focused on single-hop question generation, and the state-of-the-art system on question generation does not account for multiple supporting facts while generating the questions.

### 7.1.3 Challenges

The challenges of dealing with multi-hop question generation are as follows:

- **Information Distillation:** The task of MHQG requires distilling and summa-

| |
|---|
| **Document:** Regional social norms are generally antagonistic to hunting, while a few sects, such as the Bishnoi, lay special emphasis on the conservation of particular species, such as the antelope. *(ii)* India's wildlife protection act of 1972 bans the killing of all wild animals.<br>**Question$_{SHQ}$:** *Who lay special emphasis on conservation of particular species ?* |
| **Document (1):** `Stig Blomqvist`<br>*(i)* Stig Lennart Blomqvist (born 29 July 1946) is a Swedish rally driver.<br>*(ii)* He made his international breakthrough in 1971. *(iii)* Driving an Audi Quattro for the Audi factory team, Blomqvist won the World Rally Championship drivers' title in 1984 and finished runner-up in 1985. *(iv)* He won his home event, the Swedish Rally, seven times.<br>**Document (2):** `Audi Quattro`<br>*(i)* The Audi Quattro is a road and rally car, produced by the German automobile manufacturer Audi, part of the Volkswagen Group. *(ii)* It was first shown at the 1980 Geneva Motor Show on 3 March. Production of the original version continued through 1991.<br>**Question$_{MHQ}$:** *Which car produced by the German automobile manufacturer, was driven by Stig Lennart Blomqvist?* |

Table 7.2: An example of Single-hop question (SHQ) from the `SQuAD` dataset and a Multi-hop Question (MHQ) from the `HotPotQA` dataset. The relevant sentences and answers required to form the question are highlighted in blue and red respectively.

rizing information from multiple relevant documents in contrast to a single sentence/paragraph, therefore, it is more challenging than the existing single-hop QG task.

- **Document Length:** In MHQG, there are many sentences in the documents. But the supporting facts are few, therefore it creates noise in the system. The system has to eliminate those noise while generating the questions.

- **Unknown Supporting Facts:** The presence of irrelevant information makes it difficult to capture the supporting facts required for question generation. The explicit information about the supporting facts in the document is not often readily available, making the task more complex.

- **Unstable Training:** Training neural generation with reinforcement learning-based approaches can face prohibitively poor sample efficiency and high variance. This makes the training unstable, there should be a mechanism that can mitigate this issue for sequence generation.

- **Metric for Multi-hoping in QG:** To assess the multi-hop capability of the ques-

tion generation model, there should be an automated metric. Human evaluation is an alternative way to measure the multi-hopping in QG. However, human evaluation is not a feasible solution to assess the thousands of generated questions. Additionally, human evaluation is subjective and causes a high variance in the result.

## 7.2 Multi-Hop Question Generation Model

In this section, we discuss the various components of our proposed Multi-Hop QG model. Our proposed model has four components *(i). Document and Answer Encoder* which encodes the list of documents and answers to further generate the question, *(ii). Multi-task Learning* to facilitate the learning of the primary task QG by introducing the auxiliary task of supporting facts prediction, *(iii). Question Decoder*, which generates questions using the pointer-generator mechanism and *(iv). MultiHop-Enhanced QG* component that forces the model to generate those questions that can maximize the supporting facts prediction-based reward. The architecture of the proposed model is depicted in Fig 7.1.

### 7.2.1 Document and Answer Encoder

The encoder of the Multi-Hop QG model encodes the answer and documents using the layered Bi-LSTM network.

- **Answer Encoding:** We introduce an answer tagging feature that encodes the relative position information of the answer in a list of documents. The answer tagging feature is an $N$ length list of vectors of dimension $d_1$, where each element has either a tag value of 0 or 1. Elements that correspond to the words in the answer text span have a tag value of 1, else the tag value is 0. We map these tags to the embedding of dimension $d_1$. We represent the answer encoding features using $\{a_1, \ldots, a_N\}$.
- **Hierarchical Document Encoding:** To encode the document list $L$, we first concatenate all the documents $D_k \in L$, resulting in a list of $N$ words. Each word in this list is then mapped to a $d_2$ dimensional word embedding $u \in \mathbb{R}^{d_2}$. We then concatenate the document word embeddings with answer encoding features and feed it to a bi-directional LSTM encoder $\{LSTM^{fwd}, LSTM^{bwd}\}$.

$$z_t = LSTM(z_{t-1}, [u_t, a_t]) \qquad (7.2)$$

We compute the forward hidden states $\vec{z}_t$ and the backward hidden states $\overleftarrow{z}_t$ and concatenate them to get the final hidden state $z_t = [\vec{z}_t \oplus \overleftarrow{z}_t]$. The answer-aware supporting facts predictions network (which will be introduced shortly) takes the encoded representation as input and predicts whether the candidate sentence is a supporting fact. We represent the predictions with $p_1, p_2, \ldots, p_K$. Similar to answer encoding, we map each prediction $p_i$ with a vector $v_i$ of dimension $d_3$.

A candidate sentence $S_i$ contains the $n_i$ number of words. In a given document list $L$, we have $K$ candidate sentences such that $\sum_{i=1}^{i=K} n_i = N$. We generate the supporting fact encoding $sf_i \in \mathbb{R}^{n_i \times d_3}$ for the candidate sentence $S_i$ as follows:

$$sf^i = e_{n_i} v_i^{\mathrm{T}} \qquad (7.3)$$

where $e_{n_i} \in \mathbb{R}^{n_i}$ is a vector of 1s. The rows of $sf_i$ denote the supporting fact encoding of the word present in the candidate sentence $S_i$. We denote the supporting facts encoding of a word $w_t$ in the document list $L$ with $s_t \in \mathbb{R}^{d_3}$. Since we also, deal with the answer-aware supporting facts predictions in a multi-task setting, therefore, to obtain supporting facts induced encoder representation, we introduce another Bi-LSTM layer.

$$h_t = LSTM(h_{t-1}, [z_t, u_t, a_t, s_t]) \qquad (7.4)$$

Similar to the first encoding layer, we concatenate the forward and backward hidden states to obtain the final hidden state representation.

### 7.2.2 Multi-Task Learning for Supporting Facts Prediction

We introduce the task of *answer-aware supporting facts prediction* to condition the QG model's encoder with the supporting facts information. Multi-task learning facilitates the QG model to automatically select the supporting facts conditioned on the given answer. This is achieved by using a multi-task learning framework where the *answer-aware* supporting facts prediction network and Multi-hop QG share a common document encoder (Section 7.2.1). The network takes the encoded representation of each candidate sentence $S_i \in CS$ as input and sentence-wise predictions for the supporting facts. More specifically, we concatenate the first and last hidden state representation of each candidate sentence from the encoder outputs and pass it through a fully-connected layer that outputs a Sigmoid probability for the sentence to be a supporting fact. The architecture of this network is illustrated in Figure 7.1 (**left**). This network is then trained with a binary cross-entropy loss and the ground-truth supporting facts labels:

$$\mathcal{L}_{sp} = -\frac{1}{N} \sum_{j=1}^{N} \sum_{i=1}^{n_j} \delta_{y_i^j = 1} \log(p_i^j) + (1 - \delta_{y_i^j \neq 1}) \log(1 - p_i^j) \tag{7.5}$$

where $N$ is the number of document lists, $S$ the number of candidate sentences in a particular training example, $\delta_i^j$ and $p_i^j$ represent the ground truth supporting facts label and the output Sigmoid probability, respectively.

### 7.2.3 Question Decoder

We use an LSTM network with global attention mechanism [170] to generate the question $\hat{Q} = \{y_1, y_2, \ldots, y_m\}$ one word at a time. We use copy mechanism [235, 81] to deal with rare or unknown words. At each timestep $t$,

$$s_t = LSTM(s_{t-1}, y_{t-1}) \tag{7.6}$$

The attention distribution $\alpha_t$ and context vector $c_t$ are obtained using the following equations:

$$\begin{aligned}
e_{t,i} &= s_t^T * h_i \\
\alpha_{t,i} &= \frac{\exp(e_{t,i})}{\sum_{j=1}^{N} \exp(e_{t,j})} \\
c_t &= \sum_{i=1}^{N} \alpha_{t,i} h_i
\end{aligned} \tag{7.7}$$

The probability distribution over the question vocabulary is then computed as,

$$P_{vocab} = \text{softmax}(\tanh(\mathbf{W_q} * [c_t \oplus s_t])) \tag{7.8}$$

where $\mathbf{W_q}$ is a weight matrix. The probability of picking a word (generating) from the fixed vocabulary words, or the probability of not copying a word from the document list

$L$ at a given timestep $t$ is computed by the following equation:

$$P_{gen} = 1 - \sigma(\mathbf{W_a}c_t + \mathbf{W_b}s_t) \tag{7.9}$$

where, $\mathbf{W_a}$ and $\mathbf{W_b}$ are the weight matrices and $\sigma$ represents the sigmoid function. The probability distribution over the words in the document is computed by summing over all the attention scores of the corresponding words:

$$P_{copy}(w) = \sum_{i=1}^{N} \alpha_{t,i} * \mathbf{1}\{w == w_i\} \tag{7.10}$$

where $\mathbf{1}\{w == w_i\}$ denotes the vector of length $N$ having the value 1 where $w == w_i$, otherwise 0. The final probability distribution over the dynamic vocabulary (document and question vocabulary) is calculated by the following:

$$P(w) = P_{gen} * P_{vocab}(w) + (1 - P_{gen}) * P_{copy}(w) \tag{7.11}$$

## 7.3 MultiHop-Enhanced Reward (MER):

We introduce a reinforcement learning-based reward function and sequence training algorithm to train the RL network. The proposed reward function forces the model to generate those questions which can maximize the reward. We model our reward function using a neural framework which we named as *"Question-Aware Supporting Fact Prediction"* network. We train our neural network-based reward function for the supporting fact prediction task on `HotPotQA` dataset. This network takes as inputs the list of documents $L$, the generated question $\hat{Q}$, and predicts the supporting fact probability for each candidate sentence. This model subsumes the latest technical advances in question answering, including character-level models, self-attention [296], and bi-attention [240]. The network architecture of the supporting facts prediction model is similar to [330], as shown in Figure 7.1 (**right**). For each candidate sentence in the document list, we concatenate the output of the self-attention layer at the first and last positions and use a binary linear classifier to predict whether the current sentence is a supporting fact. This network is pre-trained on the HotPotQA dataset using binary cross-entropy loss.

For each generated question, we compute the F1 score (as a reward) between the ground truth supporting facts and the predicted supporting facts. This reward is sup-posed to be carefully used because the QG model can cheat by greedily copying words from the supporting facts to the generated question. In this case, even though a high MER is achieved, the model loses the question generation ability. To handle this situation, we regularize this reward function with an additional Rouge-L reward, which avoids the process of greedily copying words from the supporting facts by ensuring the content matching between the ground truth and generated question. We also experiment with BLEU as an additional reward, but Rouge-L, as a reward, has been shown to outperform the BLEU reward function.

### 7.3.1 Adaptive Self-critical Sequence Training

We use the REINFORCE [300] algorithm to learn the policy defined by question generation model parameters, which can maximize our expected rewards. To avoid the high variance problem in the REINFORCE estimator, the self-critical sequence training (SCST) [223] framework is used for sequence training that uses a greedy decoding score as a baseline. In SCST, during training, two output sequences are produced: $y^s$, obtained by sampling from the probability distribution $P(y_t^s|y_1^s,\ldots,y_{t-1}^s,\mathcal{D})$, and $y^g$, the greedy-decoding output sequence. We define $r(y, y^*)$ as the reward obtained for an output sequence $y$ when the ground truth sequence is $y^*$. The SCST loss can be written as,

$$\mathcal{L}_{rl}^{scst} = -(r(y^s, y^*) - r(y^g, y^*)) * R \qquad (7.12)$$

where, $R = \sum_{t=1}^{n'} \log P(y_t^s|y_1^s,\ldots,y_{t-1}^s,\mathcal{D})$. However, the greedy decoding method only considers the single-word probability, while the sampling considers the probabilities of all words in the vocabulary. Because of this, the greedy reward $r(y^g, y^*)$ has a higher variance than the Monte-Carlo sampling reward $r(y^s, y^*)$, and their gap is also very unstable. We experiment with the SCST loss and observe that the greedy strategy causes SCST to be unstable in the training progress. Towards this, we introduce a weight history factor similar to [350]. The history factor is the ratio of the mean sampling reward and means greedy strategy reward in previous $k$ iterations. We update the SCST loss function in the following way:

$$\mathcal{L}_{rl} = -\big(r(y^s, y^*) - \alpha \frac{\sum_{i=t-h+1}^{i=t} r_i(y^s, y^*)}{\sum_{i=t-h+1}^{i=t} r_i(y^g, y^*)} r(y^g, y^*)\big) \times \sum_{t=1}^{n'} \log P(y_t^s|y_1^s,\ldots,y_{t-1}^s,\mathcal{D}) \qquad (7.13)$$

where $\alpha$ is a hyper-parameter, $t$ is the current iteration, $h$ is the history determines, and the number of previous rewards is used to estimate. The denominator of the history factor is used to normalize the current greedy reward $r(y^g, y^*)$ with the mean greedy reward of previous $h$ iterations. The numerator of the history factor ensures the greedy reward has a similar magnitude to the mean sample reward of previous $h$ iterations.

## 7.4 Datasets and Experimental Details

### 7.4.1 Network training

We train our multi-task learning-based question generator model (*MTL-QG*) using the "teacher forcing" algorithm [301] to minimize the negative log-likelihood of the model on the training data. With $y^* = \{y_1^*, y_2^*, \ldots, y_m^*\}$ as the ground-truth output sequence for a given input sequence $D$, the maximum-likelihood training objective can be written as,

$$\mathcal{L}_{ml} = -\sum_{t=1}^{m} \log P(y_t^* | y_1^*, \ldots, y_{t-1}^*, \mathcal{D}) \tag{7.14}$$

Since we are jointly training the supporting-fact prediction network with the question generation network then the total loss for the *MTL-QG* network is $\mathcal{L}_{ml} + \mathcal{L}_{sp}$. We use a mixed-objective learning function [204] to train the final network:

$$\mathcal{L}_{mixed} = \gamma_1 \mathcal{L}_{rl} + \gamma_2 \mathcal{L}_{ml} + \gamma_3 \mathcal{L}_{sp}, \tag{7.15}$$

where $\gamma_1$, $\gamma_2$, and $\gamma_3$ correspond to the weights of $\mathcal{L}_{rl}$, $\mathcal{L}_{ml}$, and $\mathcal{L}_{sp}$, respectively.

### 7.4.2 Dataset

We use the `HotPotQA` [330] dataset to evaluate our methods. This dataset consists of over 113k Wikipedia-based question-answer pairs, with each question requiring multi-step reasoning across multiple supporting documents to infer the answer. While there exist other multi-hop datasets [299, 275], only `HotPotQA` dataset provides the sentence-level labels to locate the supporting facts in the list of documents. The ground-truth information of the supporting facts facilitates stronger supervision for tracing the multi-step reasoning chains across the documents used to link the question with the answer.

Our approach utilizes the ground-truth supporting facts information (only at the time of training) to form a better input representation and reinforcing desired behavior through multi-task learning and adaptive self-critical RL framework, respectively.

Each question in `HotPotQA` is associated with 10 documents and the span information of the answer and supporting facts in these documents. However, only two of these documents effectively contain all the supporting facts and the ground-truth answer. The average number of supporting facts associated with a question is 2.38 (these are all present in the 2, out of 10 documents that contain the answer and all the supporting facts). The average length of a question and a document in `HotPotQA` are 21.82 and 198.3, respectively. In the pre-processing stage, we remove all the documents that do not contain an answer or supporting facts. Further, we remove all the comparison-based question-answer pairs that have a '*yes*' or '*no*' answer. We then combine the training set $(90,564)$ and development set $(7,405)$ and randomly split the resulting data, with 80% for training, 10% for development, and 10% for testing. We call this split as training, development, and test dataset of `HotPotQA`. We ensure that the proportion of difficulty level (easy, medium, hard) of questions was nearly uniform in the train, dev, and test sets.

### 7.4.3 Network Hyper-parameters

In our experiments, we use the same vocabulary for both the encoder and decoder. Our vocabulary consists of the top 50,000 frequent words from the training data. We use the development dataset for hyperparameter tuning.

Pre-trained GloVe embeddings [205] of dimension 300 are used in the document encoding step. The hidden dimension of all the LSTM cells is set to 512. Answer tagging features and supporting facts position features are embedded in 3-dimensional vectors. The dropout [268] probability $p$ is set to 0.3. The beam size is set to 4 for beam search. We initialize the model parameters randomly using a Gaussian distribution with Xavier scheme [76].

We first pre-train the network by minimizing only the maximum likelihood (ML) loss. Next, we initialize our model with the pre-trained ML weights and train the network with the mixed-objective learning function. The following values of hyperparameters were used to generate the results in Tables 2 and 3: (i) $\gamma_1 = 0.99$, $\gamma_2 = 0.01$, $\gamma_3 = 0.1$, (ii) $d_1 = 300$, $d_2 = d_3 = 3$, (iii) $\alpha = 0.9, \beta = 10, h = 5000$. Adam [135] optimizer is used to train the model with (i) $\beta_1 = 0.9$, (ii) $\beta_2 = 0.999$, and (iii) $\epsilon = 10^{-8}$.

For MTL-QG training, the initial learning rate is set to 0.01. For our proposed model training, the learning rate is set to 0.00001. We also apply gradient clipping [202] with range $[-5, 5]$. Based on grid-search, a mini-batch size of 16 results in chosen for quick training and fast convergence. To train the supporting facts prediction model, we use the same dataset split as discussed in Section 7.4.2 and follow the hyper-parameter setting, as discussed in [330]. The optimal beam size =4 is obtained from the model performance on the development dataset.

For the baseline implementation, we follow the hyper-parameters setting discussed in the respective work. For the semantic-reinforced model, we follow the experimental setup discussed in [339] to train the question paraphrase and question answering model. These pre-trained models are used to compute the question paraphrase probability (QPP) and question answering probability (QAP) rewards.

## 7.5 Results and Analysis

### 7.5.1 Baselines

The following sequence-to-sequence baselines are used to draw comparisons with our proposed approach:

1. **s2s** [59]: A basic neural encoder-decoder model with an attention mechanism to generate the question with only a document list as the input.
2. **s2s+copy** : We extend the *s2s* model with the copy mechanism [81] to handle rare and unknown words.
3. **s2s+answer**: This model is the extension of *s2s* with answer tagging feature. This allows the encoder to learn an answer-aware input representation.
4. **NQG** [347]: This is a variant of the QG model proposed by [347]. It is an encoder-decoder model with attention, answer tagging features, and copy mechanism.
5. **ASs2s** [133]: We extend our experiment with the answer-seprated *s2s* based model proposed in [133].
6. **Max-out Pointer** [343]: We experiment with the max-out pointer and gated-self attention-based question generation model proposed by [343].
7. **Semantic-Reinforced** [339] : We compare the performance of our proposed model

with ELMO [206] encoder based Semantic-Reinforced QG model[2] proposed by [339]. This model also uses the RL-based framework (with question paraphrase probability and question answering probability as reward functions) to improve the performance of question generation.

8. **SharedEncoder-QG**: This is an extension of the `NQG` model [347] with shared encoder for QG and answer-aware supporting fact predictions tasks. This model is a variant of our proposed model, where we encode the document list using a two-layer Bi-LSTM, which is shared between the tasks. The input to the shared Bi-LSTM is a word and answer encoding as shown in Eq. 7.2. The decoder is a single-layer LSTM that generates the multi-hop question.

9. **MTL-QG**: This variant is similar to the `SharedEncoder-QG`, here we introduce another Bi-LSTM layer which takes the question, answer, and supporting fact embedding as shown in Eq. 7.4.

### 7.5.2 Results

We report the evaluation scores on our proposed method, baselines, and state-of-the-art single-hop question generation model on the `HotPotQA` test set in Table 7.3. The performance improvements with our proposed model over the baselines and state-of-the-arts are statistically significant[3] as ($p < 0.005$). For the question-aware supporting fact prediction model, we obtain the F1 and EM scores of 84.49 and 44.20, respectively, on the `HotPotQA` development dataset. We cannot directly compare the result (21.17 BLEU-4) on the HotPotQA dataset reported in [192] as their dataset split is different, and they only use the ground-truth supporting facts to generate the questions.

We also measure the multi-hopping in terms of SF coverage and reported the results in Table 7.3 and Table 7.4. We achieve skyline performance of 80.41 F1 value on the ground-truth questions of the test dataset of HotPotQA.

### 7.5.3 Quantitative Analysis

Our results in Table 7.3 are in agreement with [272, 343, 347], which establish the fact that providing the answer tagging features as input leads to considerable improvement in the QG system's performance. Our *SharedEncoder-QG* model, which is a variant of our proposed MultiHop-QG model, outperforms all the baselines state-of-the-art models except *Semantic-Reinforced* method. The proposed *MultiHop-QG* model achieves the absolute improvement of 4.02 and 3.18 points compared to *NQG* and *Max-out Pointer* model, respectively, in terms of BLEU-4 metric.

---
[2]https://github.com/ZhangShiyue/QGforQA/
[3]We follow the bootstrap test [61] using the setup provided by [58].

| Model | BLEU-1 | BLEU-2 | BLEU-3 | BLEU-4 | METEOR | ROUGE-L | SF Coverage |
|---|---|---|---|---|---|---|---|
| *s2s* [59] | 34.98 | 22.55 | 16.79 | 13.25 | 17.58 | 33.75 | 59.61 |
| *s2s+copy* | 31.86 | 22.47 | 17.70 | 14.63 | 19.47 | 30.45 | 60.48 |
| *s2s+answer* | 39.63 | 27.35 | 21.00 | 16.83 | 17.58 | 33.75 | 61.26 |
| *NQG* [347] | 39.82 | 29.24 | 23.45 | 19.55 | 21.39 | 36.63 | 61.55 |
| *ASs2s* [133] | 39.08 | 29.06 | 23.45 | 19.66 | 22.84 | 36.98 | 64.22 |
| *Max-out Pointer* [343] | 42.58 | 30.91 | 24.61 | 20.39 | 20.36 | 35.31 | 63.93 |
| *Semantic-Reinforced* [339] | 44.07 | 32.72 | 26.18 | 21.69 | **23.61** | 39.40 | 68.74 |
| *SharedEncoder-QG* (Ours) | 41.72 | 30.75 | 24.72 | 20.64 | 22.01 | 37.18 | 65.46 |
| *MTL-QG* (Ours) | 44.17 | 32.34 | 25.74 | 21.28 | 21.21 | 37.55 | 70.11 |
| Proposed Model | **46.80** | **34.94** | **28.21** | **23.57** | 22.88 | **39.68** | **74.37** |

Table 7.3: Performance comparison between proposed approach and state-of-the-art QG models on the test set of `HotPotQA`. Here ***s2s***: sequence-to-sequence, ***s2s+copy***: s2s with copy mechanism [235], ***s2s+answer***: s2s with answer encoding.

| Model | BLEU-1 | BLEU-2 | BLEU-3 | BLEU-4 | ROUGE-L | METEOR | SF Coverage |
|---|---|---|---|---|---|---|---|
| *NQG* [347] | 39.82 | 29.24 | 23.45 | 19.55 | 21.39 | 36.63 | 61.55 |
| *SharedEncoder-QG* (NQG + Shared Encoder) | 41.72 | 30.75 | 24.72 | 20.64 | 22.01 | 37.18 | 65.46 |
| *MTL-QG (SharedEncoder-QG + SF)* | 44.17 | 32.34 | 25.74 | 21.28 | 21.21 | 37.55 | 70.11 |
| *MTL-QG + Rouge-L* | 46.58 | 34.52 | 27.59 | 22.83 | 22.64 | 39.41 | 71.27 |
| Proposed Model (MTL-QG + SF + Rouge-L + MER) | **46.80** | **34.94** | **28.21** | **23.57** | **22.88** | **39.68** | **74.37** |

Table 7.4: A relative performance (on the test dataset of `HotPotQA`) of different variants of the proposed method, by adding one model component.

To analyze the contribution of each component of the proposed model, we perform an ablation study reported in Table 7.4. Our results suggest that multitask learning with a shared encoder helps the model improve the QG performance from 19.55 to 20.64 BLEU-4. Introducing the supporting facts information obtained from the answer-aware supporting fact prediction task further improves the QG performance from 20.64 to 21.28 BLEU-4. Joint training of QG with the supporting facts prediction provides stronger supervision for identifying and utilizing the supporting facts information. In other words, by sharing the document encoder between both tasks, the network encodes a better representation (supporting facts aware) of the input document. Such a presentation can efficiently filter out irrelevant information when processing multiple documents and perform multi-hop reasoning for question generation. Further, the MultiHop-Enhanced Reward (MER) with

| Model | BLEU-1 | BLEU-2 | BLEU-3 | BLEU-4 | METEOR | ROUGE-L | SF Coverage |
|---|---|---|---|---|---|---|---|
| *s2s* [59] | 35.00 | 22.57 | 16.80 | 13.29 | 14.90 | 29.12 | 60.31 |
| *s2s+copy* | 33.52 | 23.47 | 18.46 | 15.27 | 18.81 | 30.66 | 60.45 |
| *s2s+answer* | 39.72 | 27.25 | 20.82 | 16.62 | 17.58 | 33.94 | 61.14 |
| *NQG* [347] | 42.67 | 30.53 | 23.86 | 19.42 | 20.69 | 36.79 | 61.68 |
| *ASs2s* [133] | 42.18 | 29.97 | 23.30 | 18.87 | 21.27 | 37.34 | 63.98 |
| *Max-out Pointer* [343] | 42.05 | 30.48 | 24.29 | 20.17 | 20.17 | 34.93 | 64.19 |
| *Semantic-Reinforced* [339] | 43.84 | 32.84. | 26.02 | 21.43 | 23.47 | 39.17 | 68.25 |
| *SharedEncoder-QG* (Ours) | 44.51 | 31.95 | 25.04 | 20.40 | 21.66 | 37.83 | 65.22 |
| *MTL-QG* (Ours) | 44.10 | 32.39 | 25.94 | 21.58 | 21.76 | 37.77 | 69.24 |
| Proposed Model | **46.95** | **38.85** | **27.83** | **23.25** | **22.93** | **39.74** | **73.82** |

Table 7.5: Automatic evaluation scores of the proposed approach and the baselines on the development set of `HotPotQA`.

Rouge reward provides a considerable advancement on automatic evaluation metrics. We also show the additional results on the development dataset in Table 7.5.

### 7.5.4 Qualitative Analysis

We have shown the examples in Table 7.6, where our proposed reward assists the model to maximize the uses of all the supporting facts to generate better human alike questions. In the first example, Rouge-L reward-based model ignores the information '*second czech composer*' from the first supporting fact. In contrast, our MER reward-based proposed model considers that to generate the question. Similarly, in the last example, our model considers the information '*disused station located*' from the supporting fact where the former model ignores it while generating the question.

**Human Evaluation:** For human evaluation, we directly compare the performance of the proposed approach with the NQG model. We randomly sample 100 document-question-answer triplets from the test set and ask four professional English speakers to evaluate them. We consider three modalities: *naturalness*, which indicates the grammar and fluency; *difficulty*, which measures the document-question syntactic divergence and the reasoning needed to answer the question, and *SF coverage* similar to the metric discussed in Section 7.4 except we replace the supporting facts prediction network with a human evaluator. We measure the relative supporting facts coverage compared to the ground-truth supporting facts. measure the relative coverage of supporting facts in the questions with respect to the ground-truth supporting facts. *SF coverage* provides a measure of the extent of supporting facts used for question generation. For the first two modalities, evaluators are asked to rate the question generator's performance on a 1- 5 scale (5 for the best). To estimate the *SF coverage* metric, the evaluators are asked to highlight the supporting facts from the documents based on the generated question.

| |
|---|
| **Document (1): (a)** after bedřich smetana, he was the second czech composer to achieve worldwide recognition . ... **(b)** following smetana's nationalist example , dvořák frequently employed aspects , specifically rhythms , of the folk music of moravia and his native bohemia . <br> **Document (2): (a)** concert at the end of summer ( czech : koncert na konci léta ) is 1980 czechoslovak historical film. ... <br> **Target Answer:** bedřich smetana <br> **Reference:** *which czech composer achieved worldwide recognition before the subject of "concert at the end of summer" ?* <br> **NQG:** who was the composer of concert at the end of summer? <br> **with only Rouge-L reward:** who was the composer of the composer of concert at the end of summer? <br> **with Rouge-L and MER:** what was the **second czech composer** to achieve worldwide recognition for the composer of the concert at the end of summer ? |
| **Document (1): (a)** buddy Stephens is an american football coach who is currently the head coach at east mississippi community college , where he has won three njcaa national championships and coached players such as chad kelly and john franklin iii . **(b)** with an overall record of 87–12 , stephens has a higher winning percentage ( .879 ) than the njcaa all - time leader ( butler cc 's troy morrell at 154–22 for .875 ) , but has not yet coached the required 100 games to appear on the list . <br> **Document (2): (a)** john franklin iii is an american football wide receiver for the florida atlantic owls football . <br> **(b)** he formerly played for florida state university , east mississippi community college and auburn university . <br> **Target Answer:** john franklin iii <br> **Reference:** *what player did buddy stephens coach who went on to play for florida state university , east mississippi community college and auburn university ?* <br> **NQG:** buddy stephens is an american football coach who is currently the head coach at east mississippi community college , where he has won three njcaa national championships and coached players such as chad kelly and which american football wide receiver for the florida atlantic owls football ? <br> **with only Rouge-L reward:** what american football wide receiver formerly played for the florida atlantic owls football wide receiver for the florida atlantic owls football , coached buddy stephens ? <br> **with Rouge-L reward and MER:** what is the name of the chad kelly american football coach who formerly formerly played for florida state university , east mississippi community college and auburn university ? |
| **Document (1): (a)** seedley railway station is a disused station located in the seedley area of pendleton , salford , on the liverpool and manchester railway **(b)** it was opened on 1 may 1882 and closed on 2 january 1956 . **(c)** parts of the station wall can still be seen but part of the trackbed has been covered over following the construction of the m602 motorway . <br> **Document (2): (a)** pendleton is an inner city area of salford in greater manchester , england. <br> **(b)** it is about 2 mi from manchester city centre . **(c)** the a6 dual carriageway skirts the east of the district . <br> **Target Answer:** england <br> **Reference:** *seedley railway station is a disused station located in the seedley area of pendleton , is an inner city area of salford in greater manchester , in which country ?* <br> **NQG:** what country does [UNK] railway station and pendleton, greater manchester have in common? <br> **with only Rouge-L reward**: seedley railway station is located in a city area of salford in what country? <br> **with Rouge-L and MER:** seedley railway station is a **disused station located** in a city area of salford in greater manchester , in which country ? |

Table 7.6: Sample questions, where our proposed MER-based reward model generates better questions than only Rouge-L reward and NQG model.

| Model | Naturalness | Difficulty | SF Coverage |
|---|---|---|---|
| *NQG* | 3.20 | 2.42 | 73.12 |
| Proposed | **3.47** | **3.21** | **83.04** |

Table 7.7: Human evaluation results for our proposed approach and the NQG model. Naturalness and difficulty are rated on a 1–5 scale.

**Document (1): (a)** the m6 motorway runs from junction 19 of the m1 at the catthorpe interchange , near rugby via birmingham then heads north , passing stoke - on - trent , liverpool , manchester , preston , lancaster , carlisle and terminating at the gretna junction ( j45 ) . **(b)** here , just short of the scottish border it becomes the a74(m ) which continues to glasgow as the m74 .
**Document (2): (a)** shap is a linear village and civil parish located among fells and isolated dales in eden district , cumbria , england . **(b)** the village lies along the a6 road and the west coast main line , and is near to the m6 motorway . **(c)** it is situated 10 mi from penrith and about 15 mi from kendal , in the historic county of westmorland .
**Target Answer:** m6 motorway
**Reference:** *what motoway runs from junction 19 of the m1 and is near the linear village shap ?*
**NQG:** what motorway runs from junction 19 of the m1 at the catthorpe interchange ?
**Proposed:** what motorway runs from junction 19 of the m1 at the catthorpe interchange and is near shap ?

**Document (1): (a)** the east mamprusi district is one of the twenty ( 20 ) districts in the northern region of north ghana . **(b)** the capital is gambaga .
**Document (2): (a)** shienga ( shinga ) is a village in east mamprusi district , of the northern region of ghana . **(b)** it lies at an elevation of 349 meters near the right ( southern ) bank of the white volta .
**Target Answer:** gambaga
**Reference:** *what is the capitol of the district that also includes the village of shienga ?*
**NQG:** what is the capital of the east mamprusi district ?
**Proposed:** what is the capital of the district in which shienga is located ?

**Document (1): (a)** robert clinton smith ( born march 30 , 1941 ) is an american politician who served as a member of the united states house of representatives for new hampshire 's 1st congressional district from 1985 to 1990 and the state of new hampshire in the united states senate from 1990 to 2003 .
**Document (2): (a)** new hampshire 's 1st congressional district covers the southeastern part of new hampshire . **(b)** the district consists of three general areas : greater manchester , the seacoast and the lakes region .
**Target Answer:** three
**Reference:** *bob smith served as a member of the united states house of representatives for a district that consists of how many general areas ?*
**NQG:** bob smith served as a member of how many general areas ?
**Proposed:** bob smith served as a member of the united states house of representatives for a district that consists of how many general areas ?

**Document (1): (a)** leanne rowe ( born 1982 ) is an english actress and singer , known for portraying nancy in " oliver twist " , may moss in " lilies " and baby in " dirty dancing : the classic story on stage " .
**Document (2): (a)** oliver twist is a 2005 drama film directed by roman polanski **(b)** the screenplay by ronald harwood is based on the 1838 novel of the same name by charles dickens .
**Target Answer:** roman polanski
**Reference:** *who directed the 2005 film in which leanne rowe portrayed nancy ?*
**NQG:** who directed the film in which leanne rowe played nancy ?
**Proposed:** who directed the 2005 drama film in which leanne rowe played nancy ?

**Document (1): (a)** not without laughter is the debut novel by langston hughes published in 1930 .
**Document (2): (a)** james mercer langston hughes ( february 1 , 1902 – may 22 , 1967 ) was an american poet . **(b)** he was a social activist , novelist , playwright , and columnist from joplin , missouri .
**Target Answer:** american
**Reference:** *what was the nationality of the author of " not without laughter " ?*
**NQG:** not without laughter is a novel by a man of what nationality ?
**Proposed:** not without laughter is the debut novel by the poet of what nationality ?

Table 7.8: Samples generated by multi-hop question generation approach. In each document, the supporting facts are shown in blue and the target answer is in red.

We report the average scores of all the human evaluators for each criterion in Table 7.7. The proposed approach is able to generate better questions in terms of *Difficulty*, *Naturalness*, and *SF Coverage* when compared to the *NQG* model. We also compare the questions generated from the *NQG* and our proposed method with the ground-truth questions. These questions with additional generated questions are given in Table 7.8.

## 7.6 Conclusion

This chapter has introduced the multi-hop question generation task, which extends the natural language question generation paradigm to multiple document QA. We employ multitask learning with the auxiliary task of answer-aware supporting fact prediction to guide the question generator. After that, we present a novel reward formulation to improve the multi-hop question generation using the reinforcement learning framework. Our proposed method performs considerably better than the state-of-the-art question generation systems on `HotPotQA` dataset. We also introduce SF Coverage, an evaluation metric to compare the performance of question-generation systems based on their capacity to accumulate information from various documents. Overall, we propose a new direction for question-generation research with several practical applications.

The next and final contributing chapter will focus on code-mixed text generation, which can facilitate building a transformer-based model for large-scale code-mixed question answering and several other NLP tasks.

# Chapter 8

# Code-Mixed Text Generation

## 8.1 Introduction

In the previous chapter, we introduce the task of multi-hop question generation and reinforcement-based encoder-decoder architecture to generate complex questions by considering multiple documents. The generated questions can be used as an additional dataset to improve a question-answering system's performance.

In this chapter, we investigate the task of code-mixed text generation in multiple Indian and European languages. The task has the potential to overcome the scarcity of code-mixed text. To circumvent the data scarcity issue, we propose an effective deep learning approach for automatically generating the code-mixed text from English to multiple languages without any parallel data. In order to train the neural network, we create synthetic code-mixed texts from the available parallel corpus by modeling various linguistic properties of code-mixing.

Multilingual content is prominent on social media handles, especially in multilingual communities. Code-mixing is a common expression of multilingualism in informal text and speech, where there is a switch between the two languages, frequently with one in the character set of the other language. This has been a mean of communication in a multi-cultural and multilingual society, and varies according to the culture, beliefs, and moral values of the respective communities. Linguists have studied the phenomenon of code-mixing and have put forward many linguistic hypotheses [15, 207, 208], and formulated various constraints [233, 53, 122] to define a general rule for code-mixing. However, for all the scenarios of code-mixing, particularly for the syntactically divergent languages [19], these limitations cannot be postulated as a universal rule. In literature, there are mainly three dominant theories to explain the phenomenon of code-mixing formation. These are **(1)** Equivalence Constraint (EC) theory [209, 232], **(2)** Functional Head Constraint (FHC) theory [15, 53] and **(3)** Matrix Language Frame (MLF) theory [188, 122]. We adopt MLF to generate synthetic code-mixed sentences as it is less restrictive and can easily be applied to many language pairs

In this chapter, we model the code-mixed phenomenon using the feature-rich and pre-trained language model-assisted encoder-decoder paradigm. The feature-rich encoder helps the model capture the linguistic phenomenon of code-mixing, especially to decide when to switch between the two languages. Similarly, the pre-trained language model provides the task-agnostic feature, which helps the model encode the generic feature. We adopt the gating mechanism to fuse the features of the pre-trained language model and the encoder. In addition to this, we also perform transfer learning to learn the prior distribution from the pre-trained NMT. The pre-trained NMT weights are used to initialize the code-mixed generation network. Transfer learning guides the code-mixed generator to generate a syntactically correct and fluent sentence. We show the effectiveness of the proposed neural code-mixed generator on eight different language pairs, *viz.* English-Hindi (en-hi), English-Bengali (en-bn), English-Malayalam (en-ml), English-Tamil (en-ta), English-Telugu (en-te), English-French (en-fr), English-German (en-de) and English-Spanish (en-es).

### 8.1.1 Problem Statement

There is an abundant source of structured and unstructured information available on the web, literature, and news article. The majority of this information is available in the English language. For most of the NLP tasks, several benchmark datasets are created primarily in the English language. The code-mixed benchmark datasets are very few and limited to only a set of tasks like sentiment analysis, language identification, entity extraction, etc. Motivated by this fact, we set up our problem statement as follows:

Given an English sentence $E$ having $m$ words $e_1, e_2, \ldots, e_m$. The task is to generate the optimal code-mixed sentence $\hat{C}$ having a sequence of $n$ words $\hat{C} = \{y_1, y_2, \ldots, y_n\}$.

|    | **Language (en)**                                               |       | **Code-Mixed (en-xx[1])**                                   |
|----|-----------------------------------------------------------------|-------|-------------------------------------------------------------|
| en | India's agriculture is their main strength.                     | en-hi | India's कृषि इसकी main strength है।                         |
| en | Especially valuable people like Connor Rooney.                  | en-bn | বিশেষ Connor Rooney মতো valuable ব্যক্তি।                   |
| en | Glasses and cups, whatever they are, can be turned upside down. | en-ta | Glasses மற்றும் cups அவை எதுவாக இருந்தாலும், தலைகீழாக மாற்றலாம் |
| en | Democracy and development go hand in hand.                      | en-de | Democracy und Development gehen Hand in Hand.               |
| en | We abolish national embassies.                                  | en-fr | Nous abolissons les embassies national.                     |

Table 8.1: Examle of English and corresponding code-mixed (en-xx) sentences.

Mathematically,

$$\begin{aligned}
\hat{C} &= \arg\max_C prob(C|E;\phi) \\
&= \arg\max_C prob(C|e_1, e_2, \ldots e_m; \phi)
\end{aligned} \quad (8.1)$$

The generated words can be either from the English sentence or from the fixed set of vocabulary words. The example of English sentences and corresponding code-mixed sentences are shown in Table 8.1.

### 8.1.2 Motivation

- In recent times, the pre-trained language model-based architectures [52, 213] have become the state-of-the-art models for language understanding and generation. The underlying data to train such models comes from the huge amount of corpus available in the form of Wikipedia, book corpus, etc. Although these are readily available in various languages, there is a scarcity of such code-mixed data, which could be used to train state-of-the-art transformer [284] based language model, such as BERT [52], RoBERTa [164], XLM [146]. The existing benchmark dataset on various NLP tasks can also be transformed into the code-mixed environmental setup, which can be used to assess the multilingual capability of the model. Therefore, it is necessary to have an automated code-mixed generation system capable of modeling intra-sentential language phenomena and generating synthetic code-mixed text.
- Creating large-scale code-mixed datasets for various NLP tasks is expensive and time-consuming as it requires particular language expertise and human efforts. Therefore, it is necessary to have a system that can directly assist the human with code-mixed text generation or automatically generate the code-mixed text.

- The large-scale dataset created using the automated approach can be used as a benchmark setup for code-mixed question answering and generation problems. The generated dataset can also be used to build multiple tools such as virtual assistants, automated help desks, etc. to assist the bilingual community.

### 8.1.3 Challenges

- **Code-Mixed Dataset:** There is a scarcity of the code-mixed corpora to train the neural network model (e.g., encoder-decoder [273] and transformer) based code-mixed text generator. Moreover, we need parallel corpora of language (L1) and Mixed Language (L1-L2) (e.g., en ↔ en-hi) to train such a system, which is very limited on large scale.
- **Performance of NLP Tools:** In order to generate the synthetic code-mixed dataset, the underlying NLP tools have to be efficient enough to perform the task of translation, parts-of-speech tagging, transliteration, word alignment, etc. The errors in these downstream NLP tools can lead to degradation in the final system performance.
- **Modeling Code-Mixed Phenomenon**: Linguists have studied the phenomenon of code-mixing and have put forward many linguistic hypotheses [15, 207, 208], and formulated various constraints [233, 53, 122] to define a general rule for code-mixing. However, for all the scenarios of code-mixing, particularly for the syntactically divergent languages [19], these limitations cannot be postulated as a universal rule.
- **Fluency and Adequacy:** This is one of the common challenges, where the model generates incorrect words/phrases. The generated sentence may not be syntactically correct or convey the full and proper meaning. The missing or wrong words can cause fluency problems in the generated code-mixed sentence.

## 8.2 Synthetic Code-Mixed Text Generation

In this section, we describe the approach to generate a synthetic code-mixed dataset. We follow the MLF theory to generate the code-mixed text. According to MLF, a code-mixed text will have a dominant language (matrix language) and inserted language (embedded language). The insertions could be words or larger constituents, and they will comply with the grammatical frame of the matrix language. However, random word insertions could lead to an unnatural code-mixed sentence, which is very rare in practice. Linguistically informed strategies to insert the words/constituents can improve the quality of a code-mixed text. It is also shown in our previous study (Section 4.2) that such a strategy benefits the quality of the generated code-mixed text. In our work, we utilize parallel corpora to learn English words' alignment in other languages. Given a pair of parallel sentences, we identify the words from English (shown in Algo 4) and substitute their aligned counterparts with the identified English words to synthesize the English-embedded code-mixed sentences. Our synthetic code-mixed generation algorithm takes a parallel sentence as input. We use the Indic-nlp-library[2] to tokenize the sentences of the Indic languages. Moses-based tokenizer[3] is used to translate European and English language texts. Thereafter, we learn the alignment matrix, which guides us to select the words or phrases to be mixed into the language. We follow the fast alignment technique proposed in [60] to align the sentences. The fast alignment algorithm is a simple log-linear re-parameterization of IBM Model 2 [25] that overcomes the problems arising from Model 1's strong assumptions and Model 2's over-parameterization. We use the official implementation[4] of the fast-align algorithm to obtain the alignment matrix. The alignment matrix is used to construct the aligned phrases between the parallel sentence. We extract the PoS, named entity (NE), and noun phrase (NP) from the English sentences, and mix these linguistic features in the proper places of the sentences of the counterpart language. We use the Stanford library[5] Stanza [212] to extract the PoS, NE, and noun phrases from the sentence. We can extract multiple aligned phrases from the alignment matrix. However, in our proposed algorithm, we are interested in aligned words/phrases which are the NEs of types, 'PERSON', 'LOCATION', 'Organization', noun phrases, and adjective words. Let us see an example of an En-Hi parallel sentence:

- **En:** When was Mahatma Gandhi born?
- **Hi:** महात्मा गांधी का जन्म कब हुआ था?
- **Code-Mixed (En-Hi):** *Mahatma Gandhi* का जन्म कब हुआ था?

**Algorithm 4** Code-Mixed Text Generation
1: **Input**: a parallel sentence (*en-sentence, x-sentence*)
2: **Output**: an equivalent code-mixed sentence (*en-x-sentence*)
3: **procedure** GETCODEMIXEDTEXT(*en-sentence, x-sentence*)
4:    *en-tokens* ← tokenize(*en-sentence*)    ▷ Tokenize the English sentence
5:    *x-tokens* ← tokenize(*x-sentence*)    ▷ Tokenize the language-x sentence
6:    *alignment* ← getAlignment(*en-sentence, x-sentence*)
7:                                                  ▷ Learn the alignment matrix
8:    *phrases* ← extractPhrase(*en-tokens, x-tokens, alignment*)    ▷ Phrase Extraction
9:    *en-x-tokens* ← *x-tokens*    ▷ Initialize the code-mixed sentence
10:    *pos* ← getPartsOfSpeechTags(*en-tokens*)
11:                                               ▷ Parts-of-speech tagging of English sentence
12:    *ner* ← getNERTags(*en-tokens*)    ▷ NER tagging of English sentence
13:    *noun-phrases* ← getNounPhrase(*en-tokens*)    ▷ Extraction of noun phrases
14:    **for** *(entity, entity-type)* **in** *ner* **do**
15:                                      ▷ Looping for each entity in English sentence
16:       **if** *entity-type* **in** [`PER', `LOC',`ORG'] and *entity* **in** *phrases* **then**
17:          *aligned-phrase* = getAlignedPhrase(phrases, entity)
18:          *en-x-tokens* ← *en-x-tokens.replace(aligned-phrase, entity)*
19:    **for** *nphrase* **in** *noun-phrase* **do**
20:                                    ▷ Looping for each noun phrase in English sentence
21:       *aligned-phrase* = getAlignedPhrase(phrases, nphrase)
22:       *en-x-tokens* ← *en-x-tokens.replace(aligned-phrase, nphrase)*
23:    **for** *(token, pos-type)* **in** *pos* **do**    ▷ Looping for each token of English sentence
24:       **if** *pos-type* == `ADJ' and *token* **in** *phrases* **then**
25:          *aligned-phrase* = getAlignedPhrase(phrases, token)
26:          *en-x-tokens* ← *en-x-tokens.replace(aligned-phrase, token)*
27:    *en-x-sentence* ← ' '.*join(en-x-tokens)*
28:                                               ▷ Join each token to form the code-mixed sentence
29:    **return** *en-x-sentence*

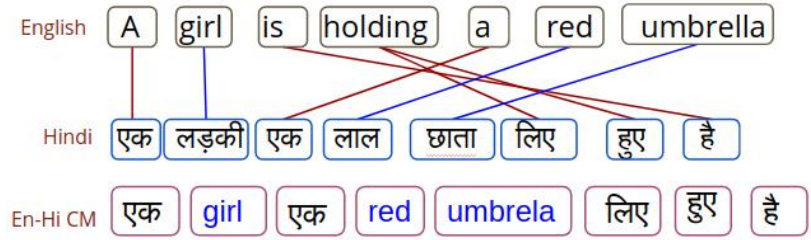

Figure 8.1: An example of the alignment between a pair of parallel sentences. The aligned words which are mixed in En-Hi code-mixed, are shown in blue.

The NE *Mahatma Gandhi* of type 'PERSON' is mixed in the En-Hi code-mixed sentence.

The need to replace the aligned noun phrases can be understood with the example of parallel sentences shown in Fig 8.1. In the given example, '*girl*' and '*red umbrella*' are the noun phrases[6] in the English sentence. To get the corresponding code-mixed sentence their aligned phrases 'लड़की' and 'लाल छाता' need to be replaced with English counterparts '*girl*' and '*red umbrella*', respectively. Similarly, we can visualize the requirement of choosing the adjective words to be mixed in a code-mixed sentence by the following example:

- **En:** The situation in Mumbai has not yet come to <u>normal</u>.
- **Hi:** मुंबई में स्थिति अभी तक <u>सामान्य</u> नहीं हुई है
- **Code-Mixed (En-Hi):** Mumbai में situation अभी <u>normal</u> नहीं हुई है ।

---

[2]https://github.com/anoopkunchukuttan/indic_nlp_library
[3]https://github.com/moses-smt/mosesdecoder
[4]https://github.com/clab/fast_align
[5]https://github.com/stanfordnlp/stanza

In the given example the adjective '*normal*' is present in the English sentence. For generating its corresponding code-mixed sentence, the adjective word has to be inserted in the code-mixed sentence. In this case, counterpart word `सामान्य' needs to be replaced with the word '*normal*' in the En-Hi code-mixed sentence.

### 8.2.1 Dataset Statistics

We create the synthetic datasets for eight different language pairs: en-hi, en-bn, en-ml, en-ta, en-te, en-fr, en-de, and en-es. We used the Europarl parallel corpus [140] v7[7] for the European languages, namely French, German and Spanish. For Indic languages, namely Hindi, Bengali, Malayalam, Tamil, and Telugu, we obtain the parallel corpus from the multilingual parallel corpus directory[8] based on the open parallel corpus[9]. We show the detailed statistics of the generated code-mixed corpus in Table 8.2.

### 8.2.2 Code-mixed Complexity

We measure the complexity of the generated code-mixed text in terms of the following metrics:

**Switch-Point Fraction (SPF):** Switch-point is the point in a sentence where the language of each side of the words is different. Following [210, 304], we compute the SPF as the number of switch-points in a sentence divided by the total number of word boundaries. A sentence having more switch points is more complex since it contains many interleaving words in different languages.

---

[6]We remove the determiner from the noun phrases as the insertion of determiner in the code-mixed makes the sentence unnatural and incorrect.
[7]https://www.statmt.org/europarl/
[8]http://lotus.kuee.kyoto-u.ac.jp/WAT/indic-multilingual/index.html
[9]http://opus.nlpl.eu/

| Language Pairs | # Parallel Sentences | # Code-Mixed Sentences | Train/Dev/Test | SPF | CMI |
|---|---|---|---|---|---|
| en-es | 1,965,734 | 200,725 | 196,725/2,000/2,000 | 68.59 | 28.80 |
| en-de | 1,920,209 | 192,131 | 188,131/2,000/2,000 | 68.41 | 28.26 |
| en-fr | 2,007,723 | 197,922 | 193,922/2,000/2,000 | 68.12 | 28.40 |
| en-hi | 1,561,840 | 252,330 | 248,330/2,000/2,000 | 62.92 | 23.49 |
| en-bn | 337,428 | 167,893 | 163,893/2,000/2,000 | 67.61 | 25.41 |
| en-ml | 359,423 | 182,453 | 178,453,371/2,000/2,000 | 81.84 | 28.13 |
| en-ta | 26,217 | 12,380 | 11,380/500/500 | 78.74 | 28.16 |
| en-te | 22,165 | 10,105 | 9,105/500/500 | 76.19 | 28.69 |

Table 8.2: Statistics of parallel corpus and generated synthetic code-mixed sentences along with the training, development and test set distributions. We also show the complexity of the generated code-mixed sentence in terms of **SPF** and **CMI**.

| | Language (L1) | | Language (L2) | | Code-Mixed (L1-L2) |
|---|---|---|---|---|---|
| en | India's agriculture is their main strength. | hi | भारत की कृषि उनकी मुख्य ताकत है। | | India's कृषि इसकी main strength है। |
| en | Especially valuable people like Connor Rooney. | bn | বিশেষত কনর রুনির মতো মূল্যবান ব্যক্তি। | | বিশেষ Connor Rooney মতো valuable ব্যক্তি। |
| en | Glasses and cups, whatever they are, can be turned upside down. | ta | கண்ணாடிகள் மற்றும் கோப்பைகள், அவை எதுவாக இருந்தாலும், தலைகீழாக மாற்றலாம் . | | Glasses மற்றும் cups அவை எதுவாக இருந்தாலும், தலைகீழாக மாற்றலாம் |
| en | Democracy and development go hand in hand. | de | Demokratie und Entwicklung gehen Hand in Hand. | | Democracy und Development gehen Hand in Hand. |
| en | We abolish national embassies. | fr | Nous abolissons les ambassades nationales. | | Nous abolissons les embassies national. |

Table 8.3: Samples code-mixed **(L1-L2)** generated sentences from the parallel sentence of the language **L1** and **L2**.

**Code-mixing Index (CMI):** It is used to measure the amount of code-mixing in a corpus by accounting for the language distribution. The sentence-level CMI score can be computed with the following formula:

$$C_u(x) = \frac{N(x) - max(\ell_i \in \ell\{w_{\ell_i}(x)\})}{N(x)}, \tag{8.2}$$

where $N(x)$ is the number of tokens of utterance $x$, $w_{\ell_i}$ is the word in language $\ell_i$. We compute this metric at the corpus level by averaging the values for all sentences. We have reported the SPF and CMI values for all the language pairs in Table 8.2. We achieved the highest SPF and CMI values for *en-ml* and *en-es* pairs respectively.

## 8.3 Methodology

This section will be focused on the methodology to generate the code-mixed text using our proposed neural network-based code-mixed generator. We depict the architecture of our proposed model in Figure 8.2.

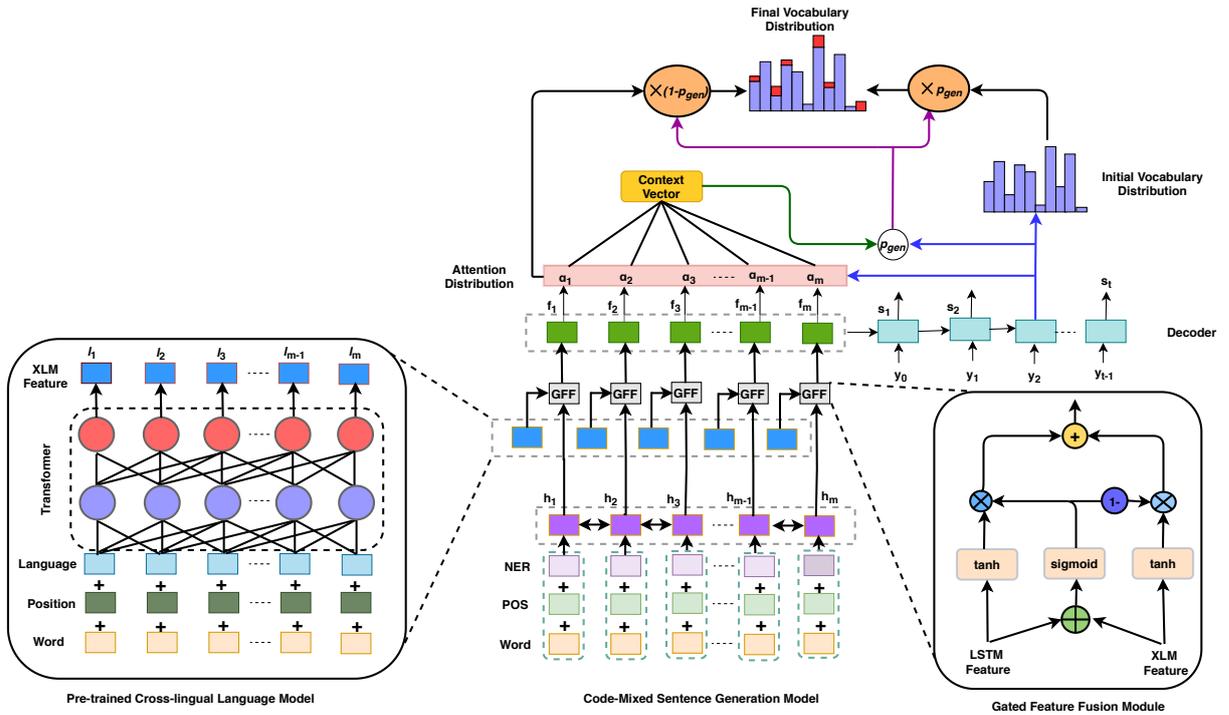

Figure 8.2: The architecture of the proposed code-mixed sentence generation model. The **left** part of the image shows the architecture of the cross-lingual language model (XLM). We use the pre-trained XLM model to extract the language model feature, which, along with the linguistic feature obtained from the Bi-LSTM encoder, is passed to the Gated Feature Fusion (GFF) module. The **right** part of the image demonstrates the working of the GFF module. It is to be noted that the transfer learning is enabled by initializing the parameters of the proposed model from the pre-trained neural machine translation model.

### 8.3.1 Sub-word Vocabulary

The task of generation using neural networks requires fixed vocabulary sizes. This generally poses the problem for out-of-vocabulary (OOV) words. To address the issue of OOV words, [238] proposed a solution, that consists of using Byte-pair encoding (BPE) to segment words into sub-words. The sub-word based tokenization schemes inspired by BPE have become the norm in most advanced models, including the very popular family of contextual language models like XLM [146], GPT-2 [213], etc. In this work, we process the language pairs with the vocabulary created using the BPE. The BPE-based sub-word is used as the vocabulary to handle the rare and unknown words.

### 8.3.2 Feature-rich and Pre-trained Assisted Encoder

We introduce a particular encoder, which is equipped with linguistic features and pre-trained language model features. First, we will discuss the linguistically-driven feature encoding technique. Later, we describe the language model-based feature encoder. We also discuss the efficient mechanism to fuse linguistically-driven features and language model features.

In order to encode the input English sentence, we use the two-layered LSTM networks. First, we tokenize the English sentence to the sub-word token using BPE. Each sub-word is mapped to a real-valued vector through an embedding layer. We also introduce the linguistic features by NER and PoS taggings[10]. The motivation to use these linguistic features comes from the synthetic code-mixed generation (c.f. section 8.2) itself, where these features guide the generation process by selecting the words to either replace with their aligned English words or to keep the same word in the code-mixed sentence. In the neural-based generation, explicit linguistic features help the decoder to decide whether to copy the word from the English word or generate it from the vocabulary.

The network takes the concatenation of word embedding $u_t$, NER encoding $n_t$ and $p_t$ (will be discussed shortly) at each time step $t$ and generate the hidden state as follows:

$$h_t = LSTM(h_{t-1}, [u_t, n_t, p_t]) \tag{8.3}$$

We compute the forward hidden states $(\vec{h}_1, \vec{h}_2, \ldots, \vec{h}_m)$ and the backward hidden states $(\overleftarrow{h}_1, \overleftarrow{h}_2, \ldots, \overleftarrow{h}_m)$. Lastly, the document encoder is the concatenation of the two hidden states, $h_i = [\vec{h}_i \oplus \overleftarrow{h}_i]$.

**Feature Encoding:** The NE and PoS features are encoded to the real-valued vectors. We initialize the NE and PoS feature representations $n_t$ and $p_t$ at time $t$ using the random vectors of size 20. The features, NE and PoS are represented by the $\{n_1, n_2, \ldots, n_m\}$ and $\{p_1, p_2, \ldots, p_m\}$, respectively.

---

[10]https://github.com/stanfordnlp/stanza

**Pre-trained language model feature:** Recent studies have shown the effectiveness of pre-trained language models for natural language understanding [52, 164] and generation [213, 55, 263]. We utilize the pre-trained feature from the XLM [146]. The XLM model is trained with three objective functions: Masked Language Modeling (MLM), Causal Language Modeling (CLM), and Translation Language Modeling (TLM). In the CLM objective, the task is to model the probability of a word given the previous words. The MLM objective was introduced in the [52], where the task is to predict the masked word from the sentence given the remaining words. The TLM objective is an extension of MLM for the parallel sentence. In TLM, the input sentence is the concatenation of the source and target sentence. A random word is masked from the concatenated sentence, and the rest of the words are used to predict the masked word.

The XLM model trained with multiple objective functions on different languages has shown the effectiveness of cross-lingual classification and machine translation. By dealing with multiple languages and setting the state-of-the-art language generation task, the pre-trained XLM model is adopted to extract the language model feature for a code-mixed generation as it is a reminiscence of the cross-lingual and generation paradigm. For the given input sentence $E : \{e_1, e_2, \ldots, e_m\}$, we extracted the language model feature $L : \{l_1, l_2, \ldots, l_m\}$.

The extracted language model features are fused to the linguistic feature as follows:

$$\begin{aligned}
h_t^* &= tanh(W_h h_t + b_h) \\
l_t^* &= tanh(W_l l_t + b_l) \\
g &= \sigma(W_g.[h_t \oplus l_t]) \\
f_t &= g \odot h_t^* + (1-g) \odot l_t^*
\end{aligned} \quad (8.4)$$

where, $\oplus$ and $\odot$ are the concatenation and element-wise multiplication operator. First, we project both the features $h_t$ and $l_t$ into the same vector space $h_t^*$ and $l_t^*$ via feed-forward network. Thereafter, we learn the gated value $g$, which controls the flow of each feature. The gated value $g$ learned how much of each feature should be part of the final encoder representation $f_t$.

### 8.3.3 Decoding with Pointer Generator

We use the one-layer LSTM network with the attention mechanism [13] to generate the code-mixed sentence $y_1, y_2, \ldots, y_n$ one word at a time. In order to deal with the rare or unknown words, the decoder has the flexibility to copy the words from documents via the pointing mechanism [235, 81]. The LSTM decoder reads the word embedding $u_{t-1}$ and the hidden state $s_{t-1}$ to generate the hidden state $s_t$ at time step $t$. Concretely,

$$s_t = LSTM(s_{t-1}, u_{t-1}) \tag{8.5}$$

We computed the context vector $c_t$ following [235]. The probability distribution over the code-mixed sentence vocabulary is computed as,

$$P_{vocab} = softmax(\tanh(\mathbf{W_c} * [c_t \oplus s_t])) \tag{8.6}$$

where, $\mathbf{W_c}$ is a weight matrix. The generation probability (generating the word from the vocabulary) is computed as follows:

$$p_{gen} = \sigma(\mathbf{W_a} c_t + \mathbf{W_b} s_t + \mathbf{W_u} u_t) \tag{8.7}$$

where $\mathbf{W_a}$, $\mathbf{W_b}$ and $\mathbf{W_s}$ are the weight matrices and $\sigma$ is the Sigmoid function. We also consider copying the word from the English sentence. The probability to copy a word from an English sentence at a given time $t$ is computed by the following equation:

$$P_{copy}(w) = \sum_{i=1}^{m} \alpha_{t,i} * \mathbf{1}\{w == w_i\} \tag{8.8}$$

where $\mathbf{1}\{w == w_i\}$ denotes the vector of length $m$ having the value 1 where $w == w_i$ otherwise 0. The final probability distribution over the dynamic vocabulary (English and code-mixed sentence vocabulary) is calculated by the following:

$$P(w) = p_{gen} P_{vocab}(w) + (1 - p_{gen}) P_{copy}(w) \tag{8.9}$$

### 8.3.4 Transfer Learning for Code-mixing

Transfer learning deals with the performance improvement of a task by using the learned knowledge from a near similar task. It has shown promising results in solving various NLP problems [274, 199] by significantly reducing the number of training instances. In our case, we formulate the problem of code-mixed text generation with respect to the neural machine translation (NMT) framework. The closer observation of the code-mixed sentence reveals that the output of the target (XX[11]) machine translation and code-mixed (En-XX) shares many words. For example:

- **Source (En):** The situation in Mumbai has not yet come to normal.
- **Target (Hi):** मुंबई में स्थिति अभी तक सामान्य नहीं हुई है ।
- **Code-Mixed (En-Hi):** Mumbai में situation अभी normal नहीं हुई है।

In the above sentences, Target (Hi) and Code-Mixed (En-Hi) share many words (underlined words). Because of this inherent similarity between machine translation and code-mixed sentence generation, we adapted the transfer learning method used in machine translation [354, 139] for code-mixing text generation.

We first train an NMT model on a large corpus of parallel data discussed in Section 8.2.1. Next, we initialize the code-mixed model with the already-trained NMT model. This code-mixed model is then trained on the synthetic code-mixed dataset. Rather than initializing the code-mixed model from the random parameters, we initialize it with the weights from the NMT model. By doing this, we achieve strong prior distribution from the NMT model to the code-mixed generation model. When we train the code-mixed generation model initialized with the NMT model, the generation model has prior knowledge of translating the English sentence into the target language XX. It has to fine-tune the model to adapt to the code-mixing phenomenon.

## 8.4 Datasets and Experimental Details

### 8.4.1 Experimental Setup

In our experiments, we use the same vocabulary for both the encoder and decoder. For the language pairs: en-hi, en-es, en-de, en-fr, we use the learned BPE codes[12] on 15 languages to segment the sentences into sub-words and use this vocabulary[13] to index the sub-words. For the language pairs: en-bn, en-ml, en-ta, en-te, we use the learned BPE codes[14] on 100 languages from the XLM model to segment the sentences into sub-words and use the correspondent vocabulary to index the sub-words. The same set of vocabulary is used to extract the pre-trained language model feature and the corresponding NMT model for transfer learning. We use the aligned multilingual word embedding[15] of dimension 300 for the language pairs: en-es, en-de, en-fr, en-hi, and en-bn from [22, 124]. For the rest of the language pairs, we obtain the monolingual embedding[16] from [22] and use the MUSE library[17] to align the vector in the same vector space. The embeddings of NE and PoS information are randomly initialized with the dimension of 20.

---

[11]XX may belong to *'es', 'de', 'fr', 'hi', 'bn', 'ml', 'ta', 'te'*
[12]https://dl.fbaipublicfiles.com/XLM/codes_xnli_15

The hidden dimension of all the LSTM cells is set to 512. We use the pre-trained XLM model[18] to extract the language model feature of dimension 1024 for en-hi, en-es, en-de, en-fr language pairs. For the rest of the language pairs, the pre-trained model[19] trained on MLM objective function is used to extract the language model feature. We use beam search of beam size 4 to generate the code-mixed sentence. Adam [136] optimizer is used to train the model with (i) $\beta_1 = 0.9$, (ii) $\beta_2 = 0.999$, and (iii) $\epsilon = 10^{-8}$ and initial learning rate of 0.0001. The maximum length of English and code-mixed tokens are set to 60 and 30, respectively. We set 5 as the minimum decoding steps in each code-mixed language pair. We use the *en-hi* development dataset to tune the network hyper-parameters. Similar to the other generation tasks such as machine translation and question generation, we evaluate the generated text using the metrics, BLEU [200], ROUGE [160] and METEOR [14]. We compute the metrics score between the reference code-mixed sentence and the generated code-mixed sentence.

---

[13]https://dl.fbaipublicfiles.com/XLM/vocab_xnli_15
[14]https://dl.fbaipublicfiles.com/XLM/codes_xnli_100
[15]https://fasttext.cc/docs/en/aligned-vectors.html
[16]https://fasttext.cc/docs/en/pretrained-vectors.html
[17]https://github.com/facebookresearch/MUSE
[18]https://dl.fbaipublicfiles.com/XLM/mlm_tlm_xnli15_1024.pth
[19]https://dl.fbaipublicfiles.com/XLM/mlm_100_1280.pth

| Model | en-es | | | en-de | | | en-fr | | | en-hi | | |
|---|---|---|---|---|---|---|---|---|---|---|---|---|
| | B | R | M | B | R | M | B | R | M | B | R | M |
| Seq2Seq | 16.42 | 36.03 | 24.23 | 19.19 | 36.19 | 24.87 | 19.28 | 38.54 | 26.41 | 15.49 | 35.29 | 23.72 |
| Attentive-Seq2Seq | 17.21 | 36.83 | 25.41 | 20.12 | 37.14 | 25.64 | 20.12 | 39.30 | 27.54 | 16.55 | 36.25 | 24.97 |
| Pointer Generator | 18.98 | 37.81 | 26.13 | 21.45 | 38.22 | 26.14 | 21.41 | 40.42 | 28.76 | 17.62 | 37.32 | 25.61 |
| Proposed Model | **22.47** | **41.24** | **29.45** | **24.15** | **42.76** | **30.47** | **24.89** | **43.54** | **31.26** | **21.55** | **40.21** | **28.37** |
| (-) BPE | 21.72 | 40.67 | 28.65 | 23.31 | 41.89 | 29.76 | 24.27 | 43.02 | 30.84 | 20.89 | 39.54 | 27.43 |
| (-) POS Feature | 22.21 | 40.92 | 29.12 | 23.76 | 42.12 | 29.88 | 24.21 | 42.95 | 30.86 | 21.02 | 39.84 | 27.91 |
| (-) NER Feature | 21.52 | 40.32 | 28.41 | 22.19 | 41.64 | 29.39 | 23.92 | 42.52 | 30.37 | 20.42 | 39.20 | 27.46 |
| (-) LM Feature | 21.56 | 40.36 | 28.42 | 23.21 | 41.85 | 29.56 | 23.82 | 42.48 | 30.29 | 20.47 | 39.17 | 27.24 |
| (-) Transfer Learning | 20.69 | 39.39 | 27.53 | 22.39 | 40.98 | 28.87 | 22.64 | 41.57 | 29.34 | 19.48 | 38.34 | 26.41 |

Table 8.4: Performance comparison of the proposed model for the code-mixed generation with the baseline models. The impact of each component (by removing one at a time) on the performance of the model. Here, **B**: BLEU, **R**: Rouge-L and **M**: METEOR

## 8.5 Results and Analysis

### 8.5.1 Baselines

We compare the performance of our proposed code-mixed generation model with the following baseline models:

- **Seq2Seq** [273]: This is the first baseline focused on generating a code-mixed text given an English sentence. The FastText embedding discussed in Section 8.4.1 is used with the basic tokenization mechanism. We use the one-layer LSTM of the hidden dimension 512 for encoder and decoder both. We evaluate the performance of the Seq2Seq model on each code-mixed language pair. The results are reported in Table 8.4 and 8.5.
- **Attentive-Seq2Seq** [13]: This baseline is the extension to *Seq2Seq* baseline with the attention mechanism.
- **Pointer Generator** [235]: We extend the *Attentive-Seq2Seq* model with the copy mechanism to handle the rare and unknown words. Similar to the other baselines, we evaluated the performance of the pointer generator model on each language pair for the code-mixed generation task.

### 8.5.2 Quantitative Analysis

We have reported the results of our proposed model in Table 8.4 and Table 8.5. The performance of the proposed model is compared over the three baselines and reported in Table 8.4 and Table 8.5. It can be visualized from the results that the Pointer Generator baseline is superior amongst all the baselines and achieves the maximum BLEU score of 21.45 for the *en-de* code-mixed language pair. Our proposed model achieves the maximum BLEU score of 24.89 for the *en-fr* code-mixed language pair. The minimum BLEU score that we achieve is 14.81 for the en-te language pair. We achieve lower BLEU scores for the language pairs en-ta, and en-te compare to the other language pairs. It is because the number of training samples for *en-ta* and *en-te* are very low (11, 380 and 9, 105) as compared to the other language pairs. For the European languages, *en-fr* pair achieves the highest performance while *en-hi* in Indian languages reports the comparable BLEU score with the *en-bn* language pair.

| Model | en-bn | | | en-ml | | | en-ta | | | en-te | | |
|---|---|---|---|---|---|---|---|---|---|---|---|---|
| | B | R | M | B | R | M | B | R | M | B | R | M |
| Seq2Seq | 16.32 | 33.02 | 21.82 | 15.92 | 34.97 | 23.12 | 11.82 | 25.14 | 20.21 | 10.87 | 24.92 | 19.05 |
| Attentive-Seq2Seq | 17.29 | 34.12 | 23.08 | 17.21 | 35.91 | 23.94 | 13.09 | 26.57 | 21.41 | 12.14 | 26.17 | 20.11 |
| Pointer Generator | 18.24 | 35.86 | 24.36 | 18.49 | 37.16 | 25.12 | 14.03 | 27.84 | 22.53 | 13.21 | 27.37 | 21.17 |
| Proposed Model | **21.49** | **39.11** | **27.32** | **21.61** | **40.23** | **28.01** | **15.69** | **29.56** | **23.88** | **14.81** | **29.23** | **22.56** |
| (-) BPE | 20.81 | 38.64 | 26.65 | 20.89 | 39.73 | 27.49 | 15.12 | 28.92 | 23.19 | 14.15 | 28.75 | 21.82 |
| (-) POS Feature | 21.04 | 38.77 | 26.94 | 21.11 | 39.91 | 27.55 | 15.23 | 28.11 | 22.34 | 14.23 | 28.67 | 21.86 |
| (-) NER Feature | 20.49 | 38.11 | 26.32 | 20.63 | 39.23 | 27.01 | 15.19 | 29.06 | 23.48 | 14.51 | 28.63 | 22.26 |
| (-) LM Feature | 20.13 | 37.73 | 25.95 | 20.54 | 38.69 | 26.44 | 14.73 | 28.64 | 22.89 | 13.97 | 28.07 | 21.79 |
| (-) Transfer Learning | 19.67 | 37.49 | 25.87 | 20.12 | 38.74 | 26.54 | 14.48 | 28.34 | 22.72 | 13.79 | 28.12 | 21.53 |

Table 8.5: Performance comparison of the proposed model for the code-mixed generation with the baseline models. The impact of each component (by removing one at a time) on the performance of the model.

We also perform the ablation study to assess the model's component on the system performance. We remove each component at a time from the proposed model and report the results for each language pair in Table 8.4 and Table 8.5. The removal of BPE brings down the BLEU score from 0.57 (*en-ta*) to 0.84 (*en-de*). The BPE encoding helps the model mitigate the OOV words issue by providing the sub-word level information. Similarly, the removal of the PoS feature lowers the BLEU score by 0.26 (*en-es*) to 0.58 (*en-te*). The NE feature helps most to the *en-bn* code-mixed language pair as we observe the decrease of 1.0 BLEU points while the NE feature is removed.

The LM feature is obtained from the pre-trained language model, and it helps the model to obtain a better-encoded representation. The ablation study reveals that the removal of the LM feature decreases the BLEU score by 1.36 points. We observe the near similar impact of the LM feature on each language pair. Finally, transfer learning is also proven to be an integral component of the proposed model as it contributes to the maximum of 2.25 BLEU score for *en-fr* and a minimum of 1.02 BLEU score of *en-te* code-mixed language pair. The difference between the maximum and minimum contribution may be attributed to the fact that we have sufficient parallel corpus (197, 922) to train

| | | |
|---|---|---|
| en-de | Input | The real problem is statesponsored lawlessness. |
| | Reference | Das real problem ist die vom statesponsored lawlessness. |
| | PG | Das echtes problem ist die vom statesponsored Gesetz. |
| | Proposed | Das real problem ist vom statesponsored lawlessness. |
| | (-) TL | Das problem ist die statesponsored Gesetzlosigkeit. |
| en-es | Input | However we have proposed some minor changes. |
| | Reference | Con todo hemos propuestos algunas minor changes. |
| | PG | Sin embargo, hemo propuestos minero changes. |
| | Proposed | Sin embargo hemos propuestos algunas minor changes. |
| | (-) TL | Con todo hemos propuestos algunas minor cambios. |
| en-hi | Input | India's agriculture is their main strength. |
| | Reference | India का agriculture इसकी main strength है। |
| | PG | India's agriculture इसकी शक्ति है। |
| | Proposed | India का agriculture इसकी main strength है। |
| | (-) TL | India कृषि इसका main strength. |
| en-fr | Input | Read the statements by Giscard dEstaing. |
| | Reference | Lisez les statements de Giscard dEstaing. |
| | PG | Lisez déclarations de Giscard dEstaing. |
| | Proposed | Lisez les statement de Giscard dEstaing. |
| | (-) TL | Lisez de déclarations Giscard dEstaing. |

Table 8.6: Sample code-mixed sentence generated using the pointer generator (**PG**), proposed model, and proposed model variant without transfer learning (**-TL**).

| Approach | Human | B | R | M |
|---|---|---|---|---|
| Synthetic | 4.19 | 67.51 | 73.56 | 71.21 |
| Pointer Generator | 2.34 | 19.47 | 39.48 | 27.39 |
| Proposed Model | 3.26 | 24.65 | 43.55 | 29.11 |

Table 8.7: Comparison of different code-mixed text generation approaches on human and automatic evaluation metrics. Here, **B**: BLEU, **R**: Rouge-L and **M**: METEOR

the *en-fr* NMT model as compared to the *en-te* parallel corpus (10, 105).

### 8.5.3 Qualitative Analysis

We assess the quality of the generated code-mixed text and compare the different machine-generated code-mixed texts. The samples from different language pairs are shown in Table 8.6. We observe that the code-mixed sentence generated using the PG model can copy the entity from the given English sentence. However, the generated code-mixed sentences are incomplete and not fluent compared to the reference sentences. For example, in *en-hi* pair the PG-based code-mixed sentence missed the '*main*' word and it copies "*India's*" rather than generating "*India का*" which seems more natural and human-like code-mixed sentence.

Our analysis also reveals that the quality of the generated code-mixed sentence without transfer learning lacks the correct syntax and fluency. The example can be seen in the (-) TL-generated code-mixed sentence (in Table 8.6) in the *en-hi* and *en-fr*. In contrast, the generated output using the proposed model takes the benefits of both the pointer generator and transfer learning to generate an adequate, fluent, and complete human-like code-mixed sentence. We observe that the proposed model learns when to switch between languages and when to either copy the entity/phrase from an English sentence or generate it from the vocabulary. The example can be seen in *en-hi* language pair, where the model copies the word '*main strength*' from the English sentence, and it also switches between the languages at the appropriate time step by generating the correct word from the vocabulary.

We perform a human evaluation to judge the code-mixed generation quality. For the human evaluation, we randomly sample 100 examples (English sentence) from *en-hi* code-mixed dataset and asked three English and Hindi speakers to manually formulate the code-mixed sentence to evaluate the quality of the generated code-mixed sentence. We asked the speakers to score (from 1 to 5) the machine-generated code-mixed sentence with respect to the human-generated sentence. The rate will define how natural and human-like the code-mixed sentence sounds as compared to the human one. The scores are associated with the quality of the generated code-mixed sentence, where the score of 1 shows that there is a *strong disagreement* between the machine-generated and human-formulated code-mixed sentences. Similarly, 2, 3, 4 and 5 are the categorical scores for *Disagreement*, *Not Sure*, *Agreement* and *Strongly Agreement*, respectively.

In addition to the human evaluation between machine-generated and human-formulated code-mixed sentences, we also compute the automatic evaluation metrics, BLEU, Rouge-L, and Meteor. The comparison between the different approaches to human and automatic evaluation is reported in Table 8.7. The reported human evaluation score corresponds to the average of all three speakers. The proposed model achieves the human evaluation score of 3.26 compared to the synthetic generation of 4.19. It is to be noted that to generate the code-mixed sentence using the synthetic generation approach, we need a parallel corpus; however, the proposed approach only take English sentence as input. The human evaluation achieves (3.26) better score than the strongest pointer generator (2.34) baseline model.

### 8.5.4 Error Analysis:

We also perform a thorough analysis of the errors produced by the system (*en-hi*) and the way to mitigate those errors. The errors are categorized into the following types:

1. **Reference Inaccuracy:** The error in word alignment propagates and leads to the inaccurate reference code-mixed sentence. Since we use synthetic reference code-mixed sentences to train our code-mixed generator, it causes errors in the generated code-mixed sentence too. This issue can be minimized by advancing the underlying alignment algorithm.

2. **Missing/Incorrect Words:** This is one of the common error types, where the model generated the incorrect words or phrases. The missing or incorrect words cause a fluency problem in the generated code-mixed sentence. We also observe that the majority of the missing words are *function words*, while incorrectly generated words belong to the *content words* category. E.g.

   **Generated:** इस book समस्त Copyright हमारे पास हैं ।

   **Gold:** इस book के समस्त Copyright हमारे पास हैं ।

   (**Translation**: *Copyright of this book is owned by us.*)

   Here the function word के *(of)* is missing from the generated code-mixed text. This error can be reduced by employing the advanced language model (like GPT [213], MASS [263]) as a decoder in the network.

3. **Factual Inaccuracy:** The model sometimes generates the factually incorrect named entities. We also observe that this type of error occurs in often longer sentences, where the model is confused with copying/generating the relevant entity in the given context. E.g.

   **Generated:** Bluetooth stack के प्रयोग से BlueZ management |

   **Gold:** BlueZ stack के प्रयोग से Bluetooth management |

   (**Translation**: *Bluetooth management using the BlueZ stack.*)

   Here the entities *'Bluetooth'* and *'BleuZ'* are misplaced in the generated code-mixed text. The factual inaccuracy can be tackled with the inclusion of a knowledge graph [349], which will help the model generate the factually correct entity at the decoding step.

4. **Code-Mixed Inaccuracy:** We observe the inaccuracy in the generated sentence, where the model sometimes produces a sentence, that either violates the code-mixed theory or is unnatural (not human-like). E.g.

   **Generated:** एक Bill और एक Act के बीच क्या अंतर है |

   **Gold:** एक Bill और एक Act के बीच क्या difference है |

   (**Translation**: *What is the difference between a bill and an act?*)

   Here the noun word '*difference*' could not be generated by the model; instead, it generates the word 'अंतर'. However, according to code-mixed theory, the noun word '*difference*' should be mixed to generate the code-mixed sentence.

5. **Rare Language Pairs:** We notice that the system makes more errors on the *en-ta* and *en-te* language pairs. It can be understood by the fact that we had a comparatively lesser number of samples of these language pairs to train the system. This error can be reduced by training the system with a sufficient number of training samples.

6. **Others:** We categorize the remaining errors into other categories. The other type of errors includes repeated word, inadequate sentence generation, and new word generation. We also observe that majority of the error occurred when the input sentence was relatively longer than 12 words. Those errors can be further reduced with sentence simplification [56] or text splitting of the longer input sentence.

## 8.6 Conclusion

In this chapter, we have proposed a neural-based transfer learning approach. Our proposed approach is based on encoder-decoder architecture, where we enrich the encoder with the linguistic and pre-trained language model features. To train and evaluate the proposed approach, we introduce a linguistic-driven approach for code-mixed sentence generation using parallel sentences from the particular language pair. Our experimental results and in-depth analysis shows that feature representation and transfer learning effectively improves the model performance and the quality of the generated code-mixed sentence. We have shown the effectiveness of the proposed approach in eight different language pairs.

In the next and final chapter of this dissertation, we will draw the dissertation's conclusion and show the future research direction.

# Chapter 9

# Conclusion and Future Scope

In this dissertation, we investigate two complementary research directions in QA, first aiming to bridge the digital language gap by expanding the breadth of the current QA system to a multilingual and code-mixed setting. The second research direction focuses on advancing and improving the QA system performance by learning from the related sub-task of QA, i.e. semantic question matching, multi-hop question generation, and code-mixed text generation.

Precisely, in the first part of this dissertation, we studied the challenge of multilingual question answering and developed linguistically motivated unsupervised approaches to generate high-quality synthetic QA/VQA datasets in English, Hindi, and Code-Mixed English-Hindi languages to overcome the data scarcity in a multilingual setting. Later, we build a QA system capable of understanding, comprehending, and retrieving the answers for multilingual/code-mixed questions. To deal with code-mixed questions, we propose an answer-type focused neural framework with bi-linear attention that enables the model to predict the multi-word answer span. We bring the multilinguality into the model by learning the lower-level representation of the words with multilingual embedding. For handling multilingual questions, we introduce a shared-question encoding mechanism, which projects and learns a joint representation of the input questions from different language origins. We learn to generate question-focused document representation by aligning the words from the question and document using the attention mechanism.

QA system capable of answering from the textual source can further benefit from the other data modalities like images, enhancing the model capacity for better semantic understanding. Towards this, we develop the VQA system, which can take code-mixed and multilingual questions and perform the reasoning over the image to predict the corresponding answer.

As the improvement in the question-answering system is a long-standing research area, the second part of this dissertation focuses on enhancing the QA systems' performance by solving multiple sub-tasks of the QA pipeline. Towards this, we propose a taxonomy-infused neural architecture for semantic question retrieval, whose ultimate goal is to improve the performance of the existing question-answering system by using external knowledge. In line with this, we have investigated the generation paradigm of question answering and introduced the novel task of multi-hop question generation. Here, we develop a novel reward function to improve the multi-hop question generation using multi-task and reinforcement learning frameworks. Our method considerably performs better over the state-of-the-art question generation systems on HotPotQA dataset. We also exploited the code-mixed text generation problem and proposed a neural transfer-learning method coupled with the linguistic and pre-trained feature representation to generate the code-mixed sentence. Our experimental results and in-depth analysis shows that feature representation and transfer learning effectively improves the model performance and the quality of the generated code-mixed sentence. We have shown the effectiveness of the proposed approach on eight different Indian and European language pairs from different origins.

As with any cutting-edge research, there are still several areas where additional work can be carried out. Some future research directions are highlighted in this section.

- In Chapter 3, we have created the resource for English-Hindi question answering [87] and proposed IR and neural network-based architecture [85] focused on learning the shared representation of the questions for multilingual question answering.

  Recently, some large-scale MQA datasets [150, 44] have been released, however, they are not appropriate for solving multi-hop question-answering task. In the near future, we will create an MQA dataset that can be used to assess the performance of a multilingual multi-hop question-answering system. We will also extend our work on developing a QA system to handle the descriptive and multi-hop reasoning questions in the multilingual environment by considering the other languages.

- We have developed an answer-type focused neural framework [89] with bi-linear attention for English-Hindi code-mixed question answering (Chapter 4). In the future, we expect to incorporate the pre-trained multilingual representations [52, 146] to improve the language understanding capability and the end-to-end performance of code-mixed question answering task. There is also a scope to extend the framework for multiple language pairs in code-mixing.

- In Chapter 5, we have proposed a VQA framework [90] based on soft-sharing of the question encoder to handle English, Hindi, or code-mixed English-Hindi questions and provide the answer given the image. In the future, the VQA work can be extended from answer prediction to answer generation in a multilingual setup. The limitation of the answer prediction network is that the model can not provide the answer if it is not present in the filtered list of answers created from the training dataset. However, answer generation does not suffer from this problem; instead, it can generate a diverse, novel, and complete answer.

- In Chapter 6, we developed a two-layered question taxonomy [91, 173] and proposed a semantic question-matching framework with the integration of taxonomy information and deep question encoder. In the future, we would like to scale the framework for semantic question matching using the enhanced question encoder such as Google Universal Sentence encoder [29], BERT, and T5 [214]. Further, this research can be extended for community question answering. Building a semantic question matching in community question answering (CQA) is challenging because the questions' nature is entirely different compared to the open-domain or closed-domain question answering. In CQA, it is more important to encode the primary latent theme of the questions as questions consist of more than one sentence and additional noise. We want to take up this challenge and work on building a semantic question-matching framework for CQA. The performance of CQA sites such Quora[1], Stack Overflow[2], Yahoo Answers[3] etc. can be further improved with a robust question matching framework.

- In Chapter 7, we have proposed the semantic-reinforced model [82] for multi-hop question generation where the model is learning to predict the supporting-facts *via* a multi-task learning framework. However, there is a significant gap between the skyline performance and multi-task learning performance. In the future, this work can be extended to improve the performance of multi-hop question generation without any weak supporting facts supervision and transformer-based sentence encoder. We would also like to assess the performance improvement in the multi-hop question-answering system by adding the questions generated using the proposed multi-hop question generation approach.

---

[1] https://www.quora.com/
[2] https://stackoverflow.com/
[3] https://in.answers.yahoo.com/

- In Chapter 8, we have studied the semi-supervised approach [86] to generate code-mixed text using the underlying encoder-decoder architecture. The unsupervised approach for language generation has recently gained visibility and shown the effectiveness of various generation tasks such as machine translation, summarization, and question-answering. Future work would be to explore the unsupervised approach for the generation of code-mixed text. Our error analysis reveals that there is a scope for improvement in the generated code-mixed sentence in terms of fluency, adequacy, and following the code-mixed theory. The incorporation of these errors can be considered a future work of code-mixed text generation.